\pdfoutput=1

\documentclass[9.5pt,journal,compsoc,cspaper,final]{IEEEtran}

\usepackage{epsfig}
\usepackage{graphicx}
\usepackage{amsmath}
\usepackage{amssymb}
\usepackage{subcaption}
\usepackage{multirow}
\usepackage{makecell}
\usepackage{booktabs}
\usepackage{enumitem}
\usepackage{color}
\usepackage{ragged2e}
\usepackage[pagebackref=true,breaklinks=true,colorlinks,bookmarks=false]{hyperref}

\newcommand{\ie}{\textit{i.e.}}
\newcommand{\eg}{\textit{e.g.}}

\newcommand{\colored}{black}
\newcommand{\seccolored}{black}

\ifCLASSOPTIONcompsoc
  \usepackage{cite}
\else
  \usepackage{cite}
\fi
\ifCLASSINFOpdf
\else
\fi
\usepackage{url}

\usepackage{tikz}
\usetikzlibrary{spy}

\begin{document}

\bstctlcite{IEEEexample:BSTcontrol}
%
\title{Learning to See Through with Events}
%
%
%
%

\author{
	Lei~Yu,
	\and Xiang~Zhang,
	\and Wei~Liao,
	\and Wen~Yang, 
	\and Gui-Song~Xia 
\IEEEcompsocitemizethanks{
\IEEEcompsocthanksitem L. Yu, X. Zhang, W. Liao, and W. Yang are with the School of Electronic Information, Wuhan University, Wuhan 430072, China. E-mail: \{ly.wd, xiangz, wei.liao, yangwen\}@whu.edu.cn.
\IEEEcompsocthanksitem G. S. Xia is with the School of Computer Science, Wuhan University, Wuhan 430072, China. E-mail: guisong.xia@whu.edu.cn.
\IEEEcompsocthanksitem The research was partially supported by the National Natural Science Foundation of China under Grants 62271354, 61871297, 61922065, 41820104006, 61871299, and the Natural Science Foundation of Hubei Province, China under Grant 2021CFB467.
\IEEEcompsocthanksitem Corresponding authors: W. Yang and G. S. Xia.
}

}

\IEEEtitleabstractindextext{
\begin{abstract}
\justifying
Although synthetic aperture imaging (SAI) can achieve the seeing-through effect by blurring out off-focus foreground occlusions while recovering in-focus occluded scenes from multi-view images, its performance is often deteriorated by dense occlusions and extreme lighting conditions.
To address the problem, this paper presents an Event-based SAI (E-SAI) method by relying on the asynchronous events with extremely low latency and high dynamic range acquired by an event camera. 
Specifically, the collected events are first refocused by a {\em Refocus-Net} module to align in-focus events while scattering out off-focus ones. 
Following that, a {\em hybrid network} composed of spiking neural networks (SNNs) and convolutional neural networks (CNNs) is proposed to encode the spatio-temporal information from the refocused events and reconstruct a visual image of the occluded targets.
Extensive experiments demonstrate that our proposed E-SAI method can achieve remarkable performance in dealing with very dense occlusions and extreme lighting conditions and produce high-quality images from pure events. Codes and datasets are available at \url{https://dvs-whu.cn/projects/esai/}.
\end{abstract}

}

\maketitle

\IEEEdisplaynontitleabstractindextext

%
\IEEEpeerreviewmaketitle

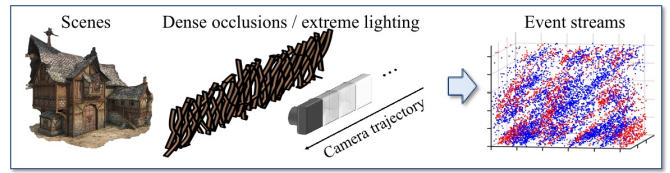
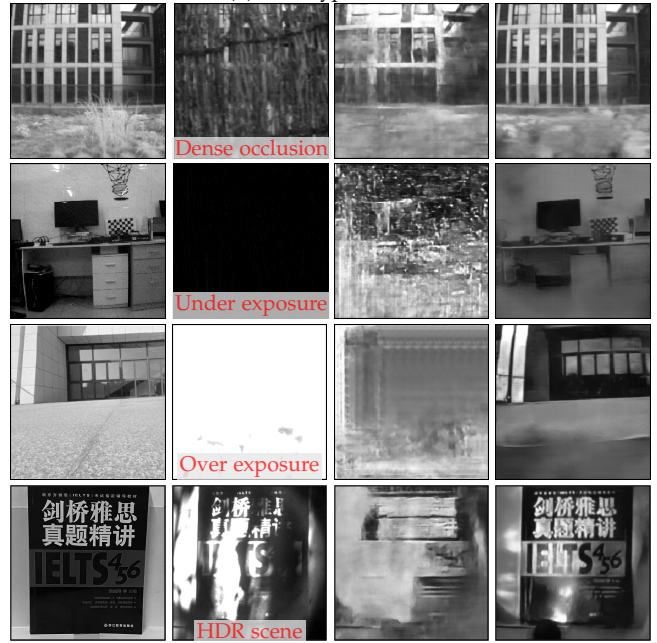
\begin{figure}[!t] 
\centering
	\begin{subfigure}[t]{0.98\linewidth}
		\includegraphics[width=\linewidth, trim={0cm 0 0cm 0}]{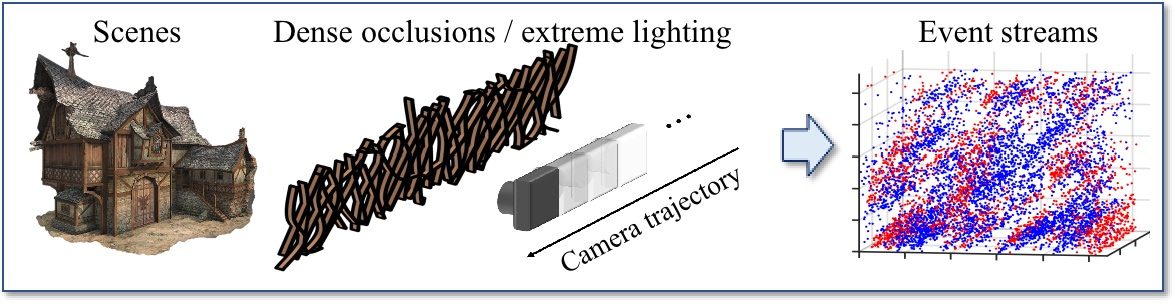}
		\vspace{-1.5em}
		\subcaption{\footnotesize Prototype of E-SAI}
		\label{fig1:a}
	\end{subfigure}
	\begin{subfigure}[t]{0.23\linewidth}
		\includegraphics[width=\linewidth, trim={0cm 0 0cm 0}]{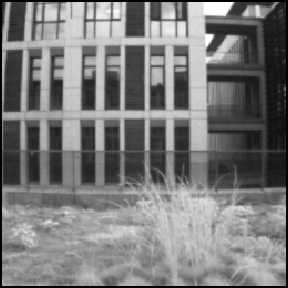}
		\vspace{-1em}\\
		\includegraphics[width=\linewidth, trim={0cm 0 0cm 0}]{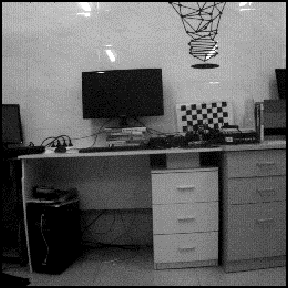}
		\vspace{-1em}\\
		\includegraphics[width=\linewidth, trim={0cm 0 0cm 0}]{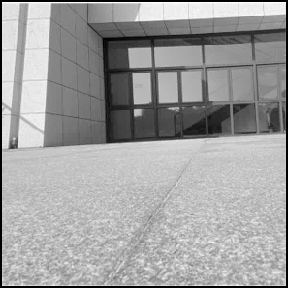}
		\vspace{-1em}\\
		\includegraphics[width=\linewidth, trim={0cm 0 0cm 0}]{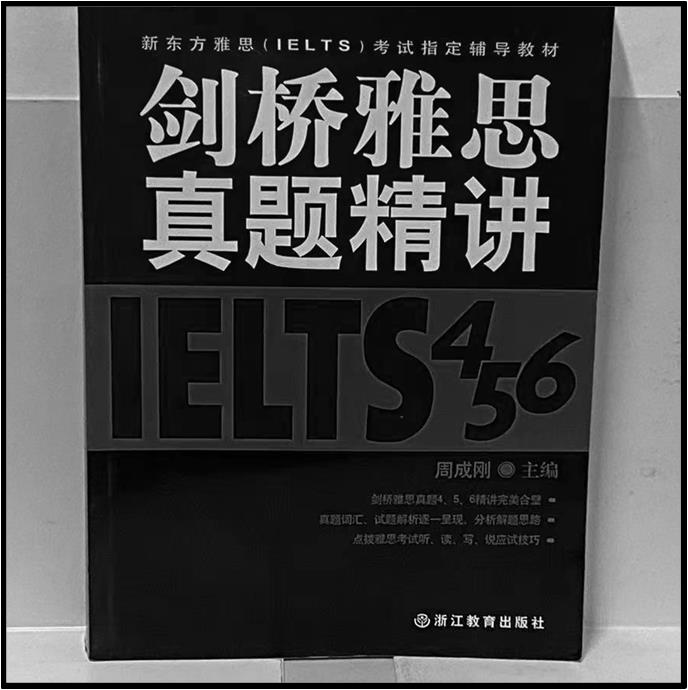}
		\vspace{-1.5em}
		\subcaption{\footnotesize Reference}
		\label{fig1:b}
	\end{subfigure}
	\begin{subfigure}[t]{0.23\linewidth}
		\begin{tikzpicture}[inner sep=0]
			\node {\includegraphics[width=\linewidth, trim={0cm 0 0cm 0}]{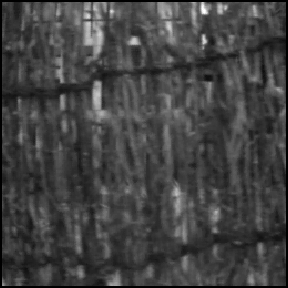}};
			\node [anchor=south,fill=black!10!white,opacity=0.75,inner sep=1] at (0,-1.03) {\textcolor{red}{\footnotesize Dense occlusion}};
		\end{tikzpicture}
		\vspace{-1em}\\
		\begin{tikzpicture}[inner sep=0]
			\node {\includegraphics[width=\linewidth, trim={0cm 0 0cm 0}]{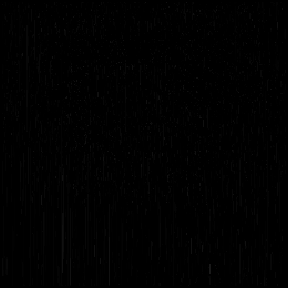}};
			\node [anchor=south,fill=black!10!white,opacity=0.75,inner sep=1] at (0,-1.03){\textcolor{red}{\footnotesize Under exposure}};
		\end{tikzpicture}
		\vspace{-1em}\\
		\begin{tikzpicture}[inner sep=0]
			\node {\includegraphics[width=\linewidth, trim={0cm 0 0cm 0}]{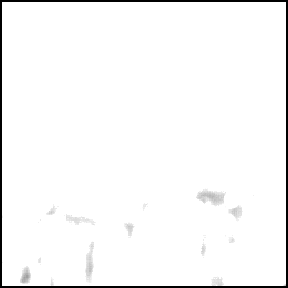}};
			\node [anchor=south,fill=black!10!white,opacity=0.75,inner sep=1] at (0,-1.03) {\textcolor{red}{\footnotesize Over exposure}};
		\end{tikzpicture}
		\vspace{-1em}\\
		\begin{tikzpicture}[inner sep=0]
			\node {\includegraphics[width=\linewidth, trim={0cm 0 0cm 0}]{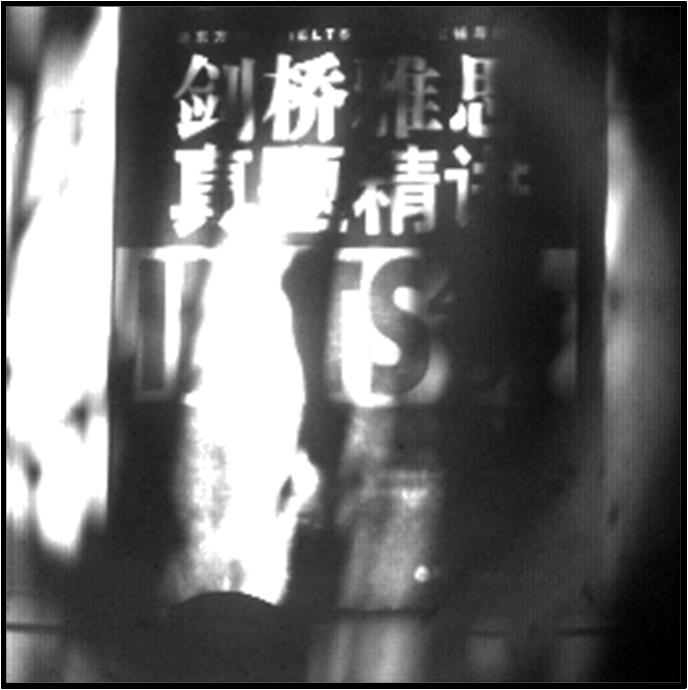}};
			\node [anchor=south,fill=black!10!white,opacity=0.75,inner sep=1] at (0,-1.03) {\textcolor{red}{\footnotesize HDR scene}};
		\end{tikzpicture}
		\vspace{-1.5em}
		\subcaption{\footnotesize View}
		\label{fig1:c}
	\end{subfigure}
	\begin{subfigure}[t]{0.23\linewidth}
		\includegraphics[width=\linewidth, trim={0cm 0 0cm 0}]{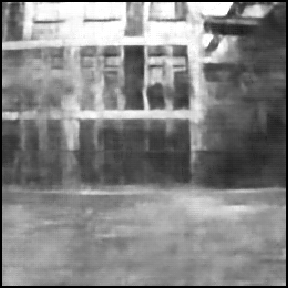}
		\vspace{-1em}\\
		\includegraphics[width=\linewidth, trim={0cm 0 0cm 0}]{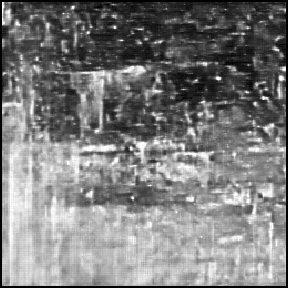}
		\vspace{-1em}\\
		\includegraphics[width=\linewidth, trim={0cm 0 0cm 0}]{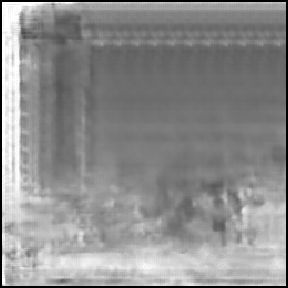}
		\vspace{-1em}\\
		\includegraphics[width=\linewidth, trim={0cm 0 0cm 0}]{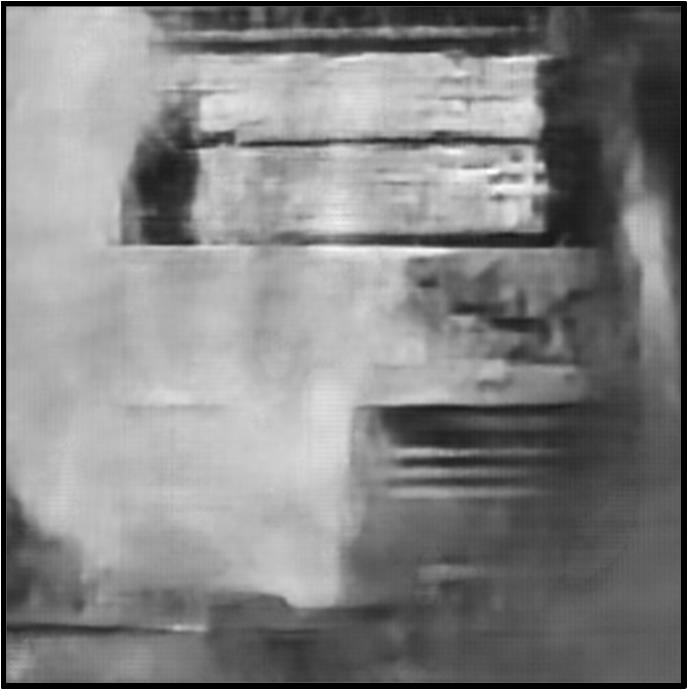}
		\vspace{-1.5em}
		\subcaption{\footnotesize F-SAI}
		\label{fig1:d}
	\end{subfigure}
	\begin{subfigure}[t]{0.23\linewidth}
		\includegraphics[width=\linewidth, trim={0cm 0 0cm 0}]{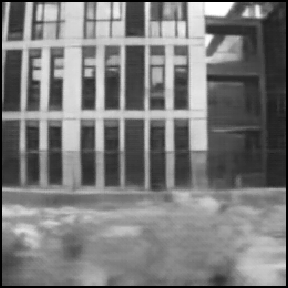}
		\vspace{-1em}\\
		\includegraphics[width=\linewidth, trim={0cm 0 0cm 0}]{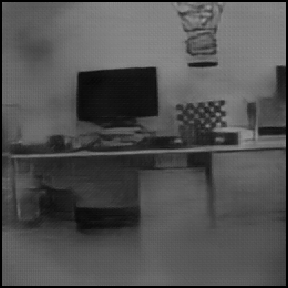}
		\vspace{-1em}\\
		\includegraphics[width=\linewidth, trim={0cm 0 0cm 0}]{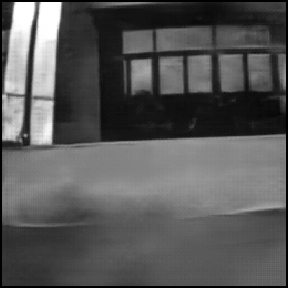}
		\vspace{-1em}\\
		\includegraphics[width=\linewidth, trim={0cm 0 0cm 0}]{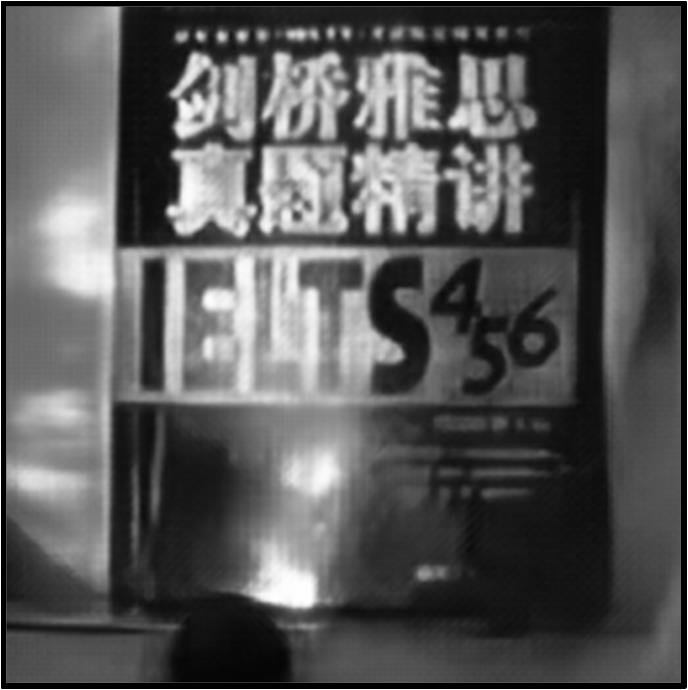}
		\vspace{-1.5em}
		\subcaption{\footnotesize {\bf E-SAI}}
		\label{fig1:e}
	\end{subfigure}
		\vspace{-0.3em}
	\caption{Prototype of the \textbf{event-based synthetic aperture imaging (E-SAI)} system (a), the illustrative indoor and outdoor scenes (b) viewing under harsh environments (c), and the corresponding seeing-through results via the state-of-the-art F-SAI \cite{wangDeOccNetLearningSee2019} (d) and the E-SAI (e). Under either very dense occlusions or extreme lighting scenes (c), the proposed E-SAI method can successfully generate high quality visual images for the occluded scenes. }
	\label{First}
\vspace{-3mm}
\end{figure}

\IEEEraisesectionheading{\section{Introduction}\label{sec:introduction}}
\IEEEPARstart{H}{arsh} environment, \eg, dense occlusions or extreme lighting conditions, often makes it difficult to efficiently acquire images of real scenes, as the collected light information is usually very limited and severely disturbed. 
Among the methods attempting to achieve {\em seeing-through} effect, synthetic aperture imaging (SAI) tackles the problem via multi-view exposures~\cite{vaishUsingPlaneParallax2004,vaishReconstructingOccludedSurfaces2006}, forming the light field of the target scene under occlusions. The basic idea of SAI is to extract the light information of the occluded scenes while filtering out foreground occlusions \cite{peiSyntheticApertureImaging2013,xiaoSeeingForegroundOcclusion2017}.
However, very dense occlusions and extreme lighting scenes may bring severe disturbances, leading to serious degradation of the imaging quality or even failure reconstructions, \eg, Fig.~\ref{First}. 
\begin{itemize}
	\item \textbf{Very dense occlusions:}  With conventional frame-based cameras, the light cues are captured via brightness intensities. Very dense occlusions will greatly decrease the ``signal'', \ie, the light from target scenes, while increase the ``noise'', \ie, disturbances from foreground occlusions, leading to a considerable reduction of the Light-SNR (ratio of ``signal'' to ``noise'').
	\item \textbf{Extreme lighting scenes:} Due to the low dynamic range (\eg, $\approx 60$ dB), images from conventional frame-based cameras usually suffer from over/under exposure problems under extreme lighting conditions, severely degrading the imaging quality and the confidence of the light information from target scenes. 
\end{itemize}

\par
\noindent Consequently, conventional frame-based SAI (F-SAI) often fails in these cases, and it is in great demand to develop new SAI methods to handle such harsh environments.
\par

In this paper, we present a novel SAI method with event cameras to address the aforementioned problems. 
Event cameras measure the pixel-wise brightness changes of scenes asynchronously, leading to many promising properties, including extremely low latency (in the order of $\mu$s), high dynamic range ($>120$ dB), and low power consumption \cite{lichtsteiner128Times1282008,9138762}.
Instead of using frame-based intensity images, as shown in Fig.~\ref{First}, event-based SAI (E-SAI) collects the light information from occluded targets via event streams, representing the brightness difference between the foreground occlusions and the occluded targets. 
{\color{\seccolored}
This mechanism means that the foreground occlusion will trigger more events for the occluded targets, \ie, more light information of targets can be recorded. 
}
With low latency, event cameras can capture adequate information of the occluded object from almost continuous viewpoints. Due to the high dynamic range of event cameras, E-SAI can collect confident light information from occluded targets even under extreme lighting conditions, leading to successful reconstructions.
\par 
{\color{\colored}

Some preliminary results have shown the feasibility of seeing-through with events~\cite{cohen2019}, but the E-SAI framework is still open, including the working mechanism and the reconstruction methodology. It is crucial to answer the following question: \textit{how to effectively process the event stream and reconstruct the high-quality visual images of occluded targets?}}
The working mechanism of the event camera differs radically from that of the frame-based one. Conventional computer vision methods, \eg, convolutional neural networks (CNNs), cannot be directly applied to such asynchronous event streams where the temporal and spatial information of events should be simultaneously considered \cite{9138762}. 
The spiking neural network (SNN) \cite{maass1998pulsed,maassNetworksSpikingNeurons1997} serves as an optimal model for integrating spatio-temporal information. Unlike other artificial neural networks, spiking neurons
do not respond to stimuli synchronously. Instead, the membrane potential of spiking neurons updates over time, and a spike will be generated whenever the membrane potential exceeds a specific spiking threshold. Thus the spatio-temporal information is naturally encoded in the spike position and timing. Exploiting this, the influence of noise events can be further mitigated from the temporal dimension, leading to the improvement of Light-SNR. 

{\color{\colored} However, adopting pure SNNs for image reconstruction tasks often suffers from performance degradation}. On the one hand, SNNs transmit information with sparse and binary spikes, which are not sufficient for high-quality image restoration tasks where high numerical precision is required to recover the accurate pixel intensities \cite{lee2020spike}.
On the other hand, recent researches have observed the vanishing spike phenomenon \cite{pandaScalableEfficientAccurate2020} in deep spiking layers.
To tackle both problems, we propose a hybrid neural network that contains an SNN encoder and a CNN decoder. With initial spiking layers, the spatio-temporal information of events can be efficiently integrated and encoded. Then, the CNN can decode the rich output of SNN and effectively reconstruct the visual image of occluded targets.
Therefore, this architecture utilizes sufficient information of events and guarantees the overall performance of reconstruction. 

\par 
The contributions of this paper are three-fold:
\begin{itemize}
	\item \textcolor{\colored}{
	We present a comprehensive analysis of the event-based SAI which can overcome the challenges of very dense occlusions and extreme lighting conditions.}
	\item We propose a novel E-SAI algorithm to reconstruct visual images of occluded target scenes through refocusing-then-reconstructing implemented in a data-driven manner, where the Refocus-Net and the hybrid SNN-CNN network are proposed respectively for refocusing and reconstruction. 
	\item \textcolor{\colored}{We build a new SAI dataset containing both image frames and event streams to facilitate event-based SAI research. Extensive experiments on the SAI dataset demonstrate the superiority of E-SAI over F-SAI under very dense occlusions and extreme lighting scenes.
	}
\end{itemize}
\par
\begin{figure}[t!]
	\centering
	\includegraphics[width=.9\linewidth]{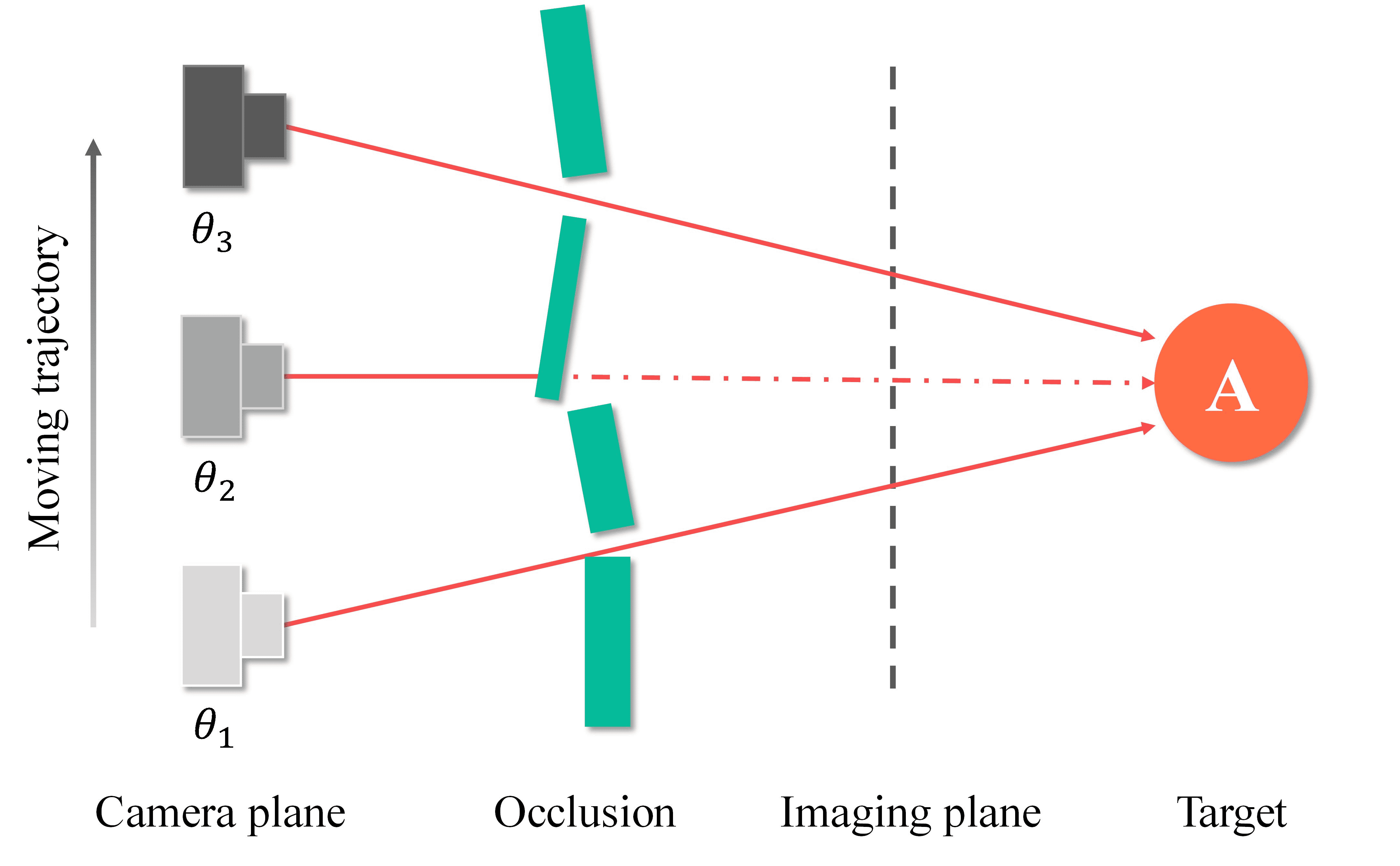}
	\vspace{-1em}
	\caption{
	{\color{\colored}
	Diagram of fronto-parallel uniform camera motion in our E-SAI system, where the event camera moves straightly with uniform velocity and fronto-parallel to occlusion and target planes.
	As the event camera moves, events triggered at camera viewpoint $\theta_1$ are induced by the brightness difference between occlusions (green)
    and target $A$ (orange), 
	while events triggered at $\theta_2$ and  $\theta_3$ are induced by high contrast textures of occlusions and targets, respectively.
	 }}
	\label{refocus}
	\vspace{-1.5em}
\end{figure}

 
A preliminary version of this work was appeared in~\cite{zhang2021event}. 
\textcolor{\colored}{
In contrast with~\cite{zhang2021event},  
this paper provides more analysis of the E-SAI framework, including more details on the components of triggered events and the corresponding epipolar geometry. 
We also design a spatial transformer network, \ie, Refocus-Net, to automatically refocus the events collected by a moving event camera with fronto-parallel uniform motion as depicted in Fig.~\ref{refocus}, relaxing the dependence on prior information such as camera velocity and target depth, and present a novel training strategy for the proposed Refocus-Net.
}
Furthermore, we enlarge the SAI dataset from 300 groups to 588 groups, including 488 groups of {\it indoor} scenes and 100 groups of {\it outdoor} scenes. Based on the enlarged dataset, we finally provide an in-depth analysis of our proposed E-SAI method.



\section{Related Work}\label{chapter-2}

\subsection{\color{\colored}Event Camera}

{\color{\colored}
As a bio-inspired vision sensor, the event camera poses a paradigm shift in visual information acquisition \cite{9138762}.
Instead of capturing full-intensity images at a fixed frame rate, event cameras only respond to the brightness change and emit asynchronous events composed of pixel position, time stamp, and polarity \cite{lichtsteiner128Times1282008}. 
This mechanism offers many outstanding properties, \eg, low latency and high dynamic range \cite{9138762}, and previous researches have revealed the potential of events in a wide variety of computer vision and robotic applications such as feature tracking \cite{gehrig2020eklt}, optical flow estimation \cite{hagenaars2021self}, and simultaneous localization and mapping (SLAM) \cite{vidal2018ultimate}. 

For imaging tasks, event data also exhibits promising benefits, especially under harsh conditions such as high dynamic range (HDR) \cite{mostafavi2021learning,rebecqEventsToVideoBringingModern2019,rebecq2019high} and high-speed motion \cite{wang2020event,9252186}.
Assuming the brightness constancy, one can reconstruct static scenes from events and simultaneously estimate the camera motion and the optical flow~\cite{cook2011,kim2014,kim2016,bardow2016}, addressing the challenges raised by fast moving cameras. Furthermore, one can reconstruct the intensity images for dynamic scenes by directly integrating the triggered events that naturally encode the temporal differentiation of the brightness in the logarithmic domain~\cite{scheerlinckContinuousTimeIntensityEstimation2019,munda2018}. However, the collected events in real-world scenarios often contain a large amount of noise induced by background activity noise, false negatives, and temporal statistics~\cite{lichtsteiner128Times1282008,9138762}, which degrades the performance of event-based imaging methods. Recently, event-based imaging gains considerable progress by leveraging deep neural networks and learning to reconstruct high frame-rate and HDR videos of the target scenes from events with improved noise robustness~\cite{rebecq2019high,wang2019,zou2021,mostafavi2021learning}.

\par 
However, existing event-based imaging methods are devoted to occlusion-free imaging, and few of them can be directly applied to the task with dense occlusions. 
}

\subsection{Synthetic Aperture Imaging} How to see through the foreground occlusion has attracted considerable interest for decades \cite{vaishUsingPlaneParallax2004,4409032,yangContinuouslyTrackingSeethrough2011,peiSyntheticApertureImaging2013,xiaoSeeingForegroundOcclusion2017,7583662,zhang2017synthetic,s19030607,wangDeOccNetLearningSee2019}. The traditional F-SAI reconstructs the occluded target via multi-view images captured by a moving camera \cite{zhang2017synthetic} or a camera array system \cite{vaishUsingPlaneParallax2004,10.1145/1186822.1073259,wangDeOccNetLearningSee2019}.
Then, by projecting all images to the plane where targets are located, the light information of the occluded target is aligned while the occlusion becomes out of focus. Afterward, reconstruction can be performed to achieve the seeing-through effect.
A plane + parallax framework has been proposed to solve the de-occlusion problem by calibrating the images captured by camera arrays \cite{vaishUsingPlaneParallax2004}.
Since the output of camera arrays can be regarded as a virtual camera imaging with a large-aperture lens, the foreground occlusion can be effectively blurred out when the background target is refocused on. But this method often results in blurry images because the information from both occlusions and targets are indiscriminately used for reconstruction. 
One can further improve the de-occlusion effect by filtering out the disturbance of occlusions using the depth-based approach \cite{yangContinuouslyTrackingSeethrough2011}, energy minimization \cite{peiSyntheticApertureImaging2013}, or k-means clustering \cite{xiaoSeeingForegroundOcclusion2017}. 
Moreover, an all-in-focus SAI method based on image matting techniques \cite{7583662} is developed to reconstruct target scenes with different depths. And a mobile camera-IMU system \cite{zhang2017synthetic} is then designed for practical usage of SAI in real-world scenarios.
Recently, deep learning based SAI methods have been proposed and achieved state-of-the-art performance \cite{wangDeOccNetLearningSee2019}. With the seeing-through ability, SAI has been exploited in a wide range of computer vision tasks, \eg, multi-object detection \cite{PEI20121637}, continuously tracking~\cite{4409032,yangContinuouslyTrackingSeethrough2011}, and 3D reconstruction of occluded objects~\cite{s19030607}.
\par 
However, the captured images of occluded targets are often severely contaminated when encountering very dense occlusions or extreme lighting conditions, making conventional F-SAI fail to achieve the seeing-through effect. In contrast, event cameras pose significant advantages in dealing with seeing-through tasks. On the one hand, sufficient light information of occluded targets can be acquired by event cameras even under the disturbance of dense occlusions due to the low latency property. On the other hand, the high dynamic range of event cameras enables the acquisition of light information even under extreme lighting conditions. Thus it motivates us to adopt event cameras to tackle the problem of SAI under very dense occlusions and extreme lighting conditions \cite{zhang2021event}. 
{\color{\colored}
Previous work \cite{cohen2019} has introduced a wide range of potential applications that can benefit from event-based vision, including low earth orbit satellite tracking, star mapping, and seeing through cloud gaps and bushes. Compared to \cite{cohen2019}, this paper not only presents a comprehensive analysis of the E-SAI task, but also proposes learning-based approaches respectively for refocusing and reconstructing which validate the superiority of E-SAI under very dense occlusions and extreme lighting conditions. 
}




\begin{figure*}[t!]
	\centering
	\includegraphics[width=.9\textwidth]{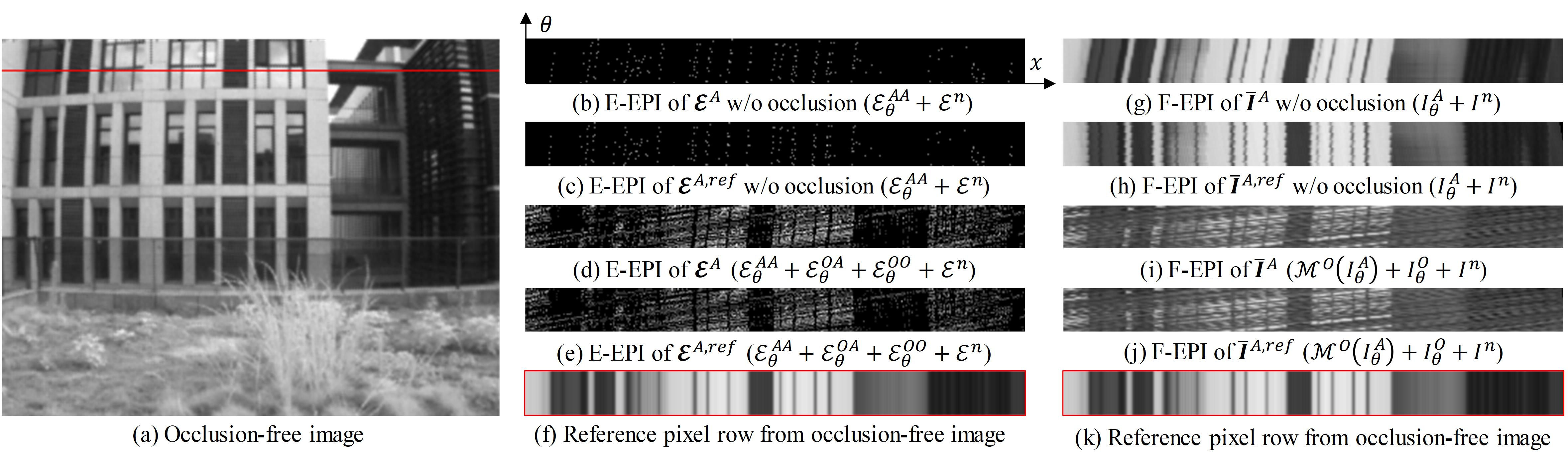}
	\vspace{-1em}
    \caption{{
    \color{\colored}
    Comparisons between frame-based EPIs (F-EPI) and event-based EPIs (E-EPIs) with and without (w/o) occlusions, where the red pixel row in the occlusion-free image (a) is selected for visualization. We compare the F-EPIs and E-EPIs under both (b, c, g, h) occlusion-free and (d, e, i, j) densely occluded scenes. E-EPIs (b, d) and F-EPIs (g, i) are generated with the collected event fields $\boldsymbol{\mathcal{E}}^A$ and light fields $\boldsymbol{\bar I}^A$, while E-EPIs (c, e) and F-EPIs (h, j) are generated with the refocused event fields $\boldsymbol{\mathcal{E}}^{A,ref}$ and light fields $\boldsymbol{\bar I}^{A,ref}$.
    Reference pixel row (f, k) from occlusion-free image is stretched for better comparison.}\vspace{-1.5em}
    }
    \label{fig:DiffEPI-common-5EPI}
\end{figure*}
\section{Problem Formulation and Analysis}
Suppose for a static unknown scene $A$ with $I_\theta^A$ representing the projected brightness intensity captured at the camera viewpoint $\theta$. Then $\boldsymbol{I}^A\triangleq \{I_\theta^A\}_{\theta\in \mathcal{P}}$ forms a tensor of  light field of $A$ with $\mathcal{P}$ denoting the set of camera viewpoints. Analogically, the light field of occlusions $O$ can be represented as  $\boldsymbol{I}^O\triangleq \{I_\theta^O\}_{\theta\in \mathcal{P}}$ with $I_\theta^O$ denoting the brightness intensity captured at the camera viewpoint $\theta$. 

\subsection{Frame-based SAI (F-SAI)} 
The task of F-SAI is to achieve the seeing-through imaging from the light fields with a limited number of occluded observations, \ie, $\boldsymbol{\bar I}^A =  \{\bar I^A_\theta\}_{\theta\in \mathcal{P}}$ with $|\mathcal{P}|<\infty$ and 
\begin{equation}\label{FSAI}
    \bar I^A_\theta = \mathcal{M}^O(I_\theta^A) + I_\theta^O + I^n,
\end{equation}
where $I^n$ denotes the measurement noise and $\mathcal{M}^O$ represents the masking operator for $\mathcal{M}^O(\cdot)=0$ only when it is occluded by $O$. 
Very dense occlusions will bring severe disturbances to $\bar I^A_\theta$ and the extreme lighting conditions often make the observations saturated, leading to failure reconstruction of visual images for $A$.
{\color{\colored}
\subsection{Event-based SAI (E-SAI)}
Instead of using frame-based cameras, we propose an event-based SAI system where the event camera is employed to collect the light field. 
Specifically, the $i$-th event $e_i = (p_i,\textbf{x}_i,t_i)$ is triggered at pixel position $\textbf{x}_i$ and time $t_i$ whenever the log-scale brightness change exceeds the event threshold $\eta$, \ie,
{\color{\colored}
\begin{equation}\label{logchange}
	\tilde{I}(\textbf{x}_i,t_i) - \tilde{I}(\textbf{x}_i,t_i-\Delta t_i) = p_i\cdot \eta,
\end{equation}
}
where $\tilde{I}=\log(I)$ denotes the log-scale pixel intensity, $\Delta t_i$ indicates the time since the last event at position $\textbf{x}_i$, $p_i\in\{+1,-1\}$ is the polarity representing the direction of brightness change, \ie, increase ($+1$) or decrease ($-1$) \cite{lichtsteiner128Times1282008}. 
\par 
As illustrated in Fig.~\ref{refocus}, events are induced by the brightness change as moving the event camera, then we can denote the collected events at camera viewpoint $\theta$ as a set of stream $\mathcal{E}_\theta^A\triangleq \{e_i\}_{i=1}^{M}=\{(p_i,\mathbf{x}_i,t_i)\}_{i=1}^{M}$ with $M=|\mathcal{E}_\theta^A|$. According to the source of brightness change, we divide the collected events into four categories, \ie,
    \begin{equation}\label{ESAI}
    \mathcal{E}^A_\theta  = \mathcal{E}^{AA}_\theta +\mathcal{E}^{OA}_\theta + {\mathcal{E}^{OO}_\theta + \mathcal{E}^{n}},
    \end{equation}
    and respectively,
    \begin{itemize}
        \item [$\bullet$] $\mathcal{E}^{AA}_{\theta}$ and $\mathcal{E}^{OO}_{\theta}$ respectively denote the set of events triggered by \textit{edges of targets $A$ and occlusions $O$ with high contrasts}. For $e_i= (p_i,\textbf{x}_i,t_i)\in \mathcal{E}^{AA}_{\theta}$ or $\mathcal{E}^{OO}_{\theta}$, the left side of Eq.~\eqref{logchange} can be written as $ \tilde{I}(\textbf{x}_i,t_i) - \tilde{I}(\textbf{x}_i,t_i-\Delta t_i) = \Delta \tilde{I}_{\theta}(\textbf{x}_i)$, \ie,
        $$\Delta \tilde{I}_{\theta}(\textbf{x}_i) = p_i\cdot \eta.$$
        By applying Taylor expansion and optical flow constraint on the brightness change on the left side, we have $
            \Delta \tilde{I}_{\theta}(\textbf{x}_i) \approx - \nabla\tilde{I}_{\theta}(\textbf{x}_i) \cdot \Delta \textbf{x}_i
        $ with $\Delta \textbf{x}_i$ denoting the pixel displacement during $\Delta t_i$  \cite{9138762}.
        Then, 
        \[
            -\nabla\tilde{I}_{\theta}(\textbf{x}_i) \cdot \Delta \textbf{x}_i \approx p_i\cdot \eta,
        \]
        which indicates that events $\mathcal{E}^{AA}_{\theta}$ or $\mathcal{E}^{OO}_{\theta}$ are mainly induced by regions of $A$ or $O$ with large gradients $\nabla\tilde{I}_{\theta}(\textbf{x}_i)$, \ie, high contrast edges. Thus, the number of events emitted for $\mathcal{E}^{AA}_{\theta}$ or $\mathcal{E}^{OO}_{\theta}$ is related to the edges of $A$ or $O$, \ie,
        {\color{\seccolored}
         \begin{equation}\label{eq:AA-OO}
            |\mathcal{E}^{AA}_\theta| \propto \left\|\nabla\tilde{I}_\theta^A(\textbf{x})\cdot \Delta \textbf{x} \right\|, \  \textnormal{and} \  |\mathcal{E}^{OO}_\theta| \propto \left\|\nabla\tilde{I}_\theta^O(\textbf{x})\cdot \Delta \textbf{x}\right\|,
        \end{equation}
        }
        with $|\cdot |$ and $\|\cdot \|$ respectively denoting the cardinality of the set and the norm / absolute value.
        \item [$\bullet$] $\mathcal{E}^{OA}_{\theta}$ denotes the set of events triggered by \textit{brightness difference between occlusions $O$ and targets $A$}. Specifically, for $e_i= (p_i,\textbf{x}_i,t_i)\in \mathcal{E}^{OA}_{\theta}$ collected at the viewpoint $\theta$, 
        the left side of Eq.~\eqref{logchange} denotes the brightness difference between $A$ and $O$
        with $\tilde{I}(\textbf{x}_i,t_i)=\tilde{I}_\theta^A(\textbf{x}_i)$ and $\tilde{I}(\textbf{x}_i,t_i-\Delta t_i)=\tilde{I}_\theta^O(\textbf{x}_i)$, \ie,
        \[
            \tilde{I}_\theta^A(\textbf{x}_i) - \tilde{I}_\theta^O(\textbf{x}_i) = p_i\cdot \eta.
        \]
        Thus the number of events emitted for $\mathcal{E}^{OA}_\theta$ is related to the brightness difference between $A$ and $O$, \ie,
        \begin{equation}\label{eq:OA}
    	|\mathcal{E}^{OA}_\theta|  \propto  \left\| \tilde{I}^A_\theta - \tilde{I}^O_\theta  \right\|.
        \end{equation}
        \item [$\bullet$] $\mathcal{E}^{n}$ denotes noise events due to \textit{physical imperfections of intrinsic circuits and ambient lights}.
    \end{itemize}
\par 
We can conclude that for a given viewpoint $\theta$, events $\mathcal{E}^{AA}_{\theta}$ and $\mathcal{E}^{OO}_{\theta}$ respectively contain edge information of targets $A$ and occlusions $O$, while events $\mathcal{E}^{OA}_{\theta}$ provide the texture information of targets $A$ relative to occlusions $O$. 
Thanks to the low latency property, E-SAI is able to collect events $\mathcal{E}^A_\theta$ from almost continuous viewpoints $\theta$ and form the event field $\boldsymbol{\mathcal{E}}^A =  \{\mathcal{E}^A_\theta\}_{\theta\in \mathcal{P}}$ with $|\mathcal{P}|\to\infty$. 
{\color{\seccolored}
\par
\noindent\textbf{Epipolar Analysis.}
}
In our setup of fronto-parallel horizontal camera motion (Fig.~\ref{refocus}), the collected event field $\boldsymbol{\mathcal{E}}^A$ can be parameterized by $({\bf x},\theta)$, with ${\bf x}\triangleq (x,y)$ representing the event coordinates and $\theta$ denoting the horizontal viewpoint. Similar to the epipolar plane images (EPIs) in light field \cite{wu2017light}, we fix $y$ coordinate and generate the event-based EPI (E-EPI) of event field $\boldsymbol{\mathcal{E}}^A$ in the $x$-$\theta$ plane as shown in Fig.~\ref{fig:DiffEPI-common-5EPI}b and~\ref{fig:DiffEPI-common-5EPI}d for scenarios without and with occlusions. 
{\color{\seccolored}
Following the frame refocusing procedure in F-SAI \cite{vaishUsingPlaneParallax2004}, an event refocusing process can be performed for 
event alignment from the collected event field $\boldsymbol{\mathcal{E}}^{A}$ to the refocused event field at the reference image plane $\boldsymbol{\mathcal{E}}^{A,ref} \triangleq \{\mathcal{E}^{A,ref}_\theta\}_{\theta\in \mathcal{P}} $ with $\mathcal{E}^{A,ref}_\theta \triangleq \{e_i^{ref}\}_{i=1}^{M}=\{(p_i,\mathbf{x}^{ref}_i,t_i)\}_{i=1}^{M}$. According to the multiple view geometry \cite{hartley2003multiple} and the pinhole imaging model \cite{PEI20121637}, the event refocusing can be formulated as:
\begin{equation}\label{trans-final}
 \mathbf{\tilde{x}}_i^{ref} = KR_{i}K^{-1}\mathbf{\tilde{x}}_{i}+\frac{KT_{i}}{d},
\end{equation}
where $\mathbf{\tilde{x}}_i$, $\mathbf{\tilde{x}}_i^{ref}$ correspond to the homogeneous coordinates of $\mathbf{x}_i$, $\mathbf{x}_i^{ref}$;
$K$ is the intrinsic matrix of camera; $R_{i}, T_{i}$ are the rotation and translation matrices between camera viewpoint $\theta_i$ and the reference one $\theta^{ref}$; target depth $d$ is the distance between target $A$ and the camera plane.
Similarly, we depict the E-EPI of the refocused event field $\boldsymbol{\mathcal{E}}^{A,ref}$ without and with occlusions in Fig.~\ref{fig:DiffEPI-common-5EPI}c and~\ref{fig:DiffEPI-common-5EPI}e.}
Compared to the reference pixel row of the occlusion-free image in Fig.~\ref{fig:DiffEPI-common-5EPI}f, the E-EPIs in Figs.~\ref{fig:DiffEPI-common-5EPI}c and \ref{fig:DiffEPI-common-5EPI}e respectively correspond to edges and textures of the target $A$, consistent with Eqs.~\eqref{eq:AA-OO} and \eqref{eq:OA}. 

\begin{figure*}[t!]
	\centering
	\includegraphics[width=.9\linewidth]{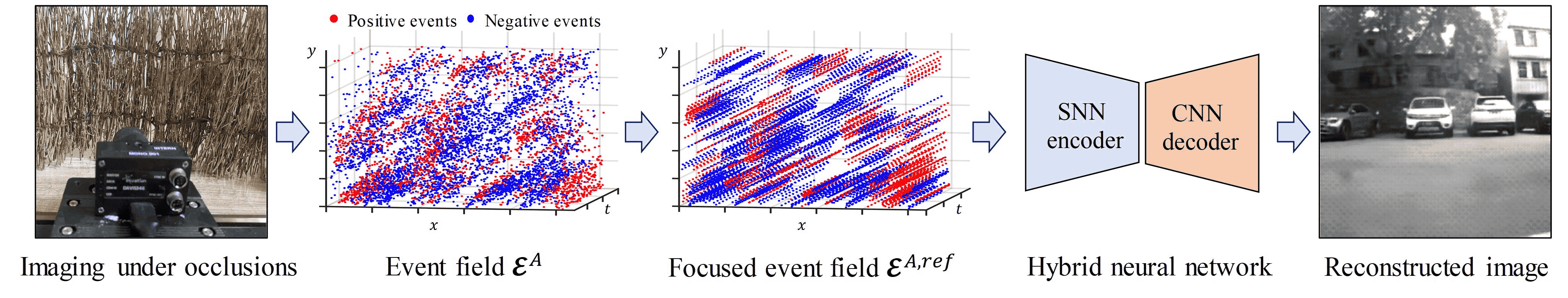}
    \vspace{-1em}
	\caption{Overall pipeline of the proposed E-SAI. As moving the event camera, E-SAI collects event streams $\mathcal{E}_\theta^A$ with almost continuous viewpoints $\theta$ and forms the \emph{event field} $\boldsymbol{\mathcal{E}}^A$. To reconstruct high quality images from $\boldsymbol{\mathcal{E}}^A$, we propose to employ the hybrid SNN-CNN network after the event refocusing process. }
	\label{pipeline}
	\vspace{-1.5em}
\end{figure*}
\par 
Therefore, when refocusing on the target plane, events $\mathcal{E}^{AA}_\theta$ and $\mathcal{E}^{OA}_\theta$ are aligned and regarded as \textbf{signal} in the E-SAI system, where $\mathcal{E}^{AA}_\theta, \mathcal{E}^{OA}_\theta$ respectively contain the target information of high contrast edges and the scene texture, and the misaligned events $\mathcal{E}^{OO}_\theta$ and $\mathcal{E}^{n}$ are treated as \textbf{noise} to be filtered out. 
{\color{\seccolored} Although the signal events $\mathcal{E}^{AA}_\theta$ will reduce when the target scenes are densely occluded, the signal events $\mathcal{E}^{OA}_\theta$ triggered by occlusions can compensate for the lost edge information in $\mathcal{E}^{AA}_\theta$ and provide more scene texture for reconstruction.
}
We will present detailed discussion of the refocusing and reconstructing methods in Sec.~\ref{sec3-1}.

\subsection{\textbf{E-SAI v.s. F-SAI}}
{\color{\seccolored}
E-SAI is better behaved than F-SAI  when encountering very dense occlusions and extreme lighting scenes. 
}
\begin{itemize}
    \item [$\bullet$] {\bf Very dense occlusions:} we present the frame-based EPIs (F-EPIs) for light fields $\boldsymbol{\bar I}^A$ and refocused light fields $\boldsymbol{\bar I}^{A,ref}$ of F-SAI with or without occlusions in Fig.~\ref{fig:DiffEPI-common-5EPI}g-j.
    {\color{\seccolored}
    It is shown that \textit{foreground occlusions bring severe disturbances to the captured frames but trigger additional signal events $\mathcal{E}^{OA}_\theta$ for reconstruction.
    }
    }
    Comparing Figs.~\ref{fig:DiffEPI-common-5EPI}g and \ref{fig:DiffEPI-common-5EPI}i, the F-EPI in occlusion-free scene highly matches the reference pixel row, but the one under occlusions is heavily contaminated by the light from foreground occlusions, \ie, $I^{O}_\theta $. By contrast, E-EPI in occlusion-free scenes mainly contains events $\mathcal{E}_{\theta}^{AA}$ that respond to high contrast regions, \eg, edges as shown in Fig.~\ref{fig:DiffEPI-common-5EPI}b. By exploiting the brightness contrast between occlusions and targets, E-EPI in Fig.~\ref{fig:DiffEPI-common-5EPI}d contains abundant events $\mathcal{E}_{\theta}^{OA}$ that provide texture information, enabling reconstruction of densely occluded scenes. 
    \item [$\bullet$] {\bf Extreme lighting scenes:} the light fields captured in F-SAI system will be severely degraded due to the over/under exposure problems when encountering extreme lighting conditions. By contrast, this issue can be largely alleviated thanks to the high dynamic range property of event cameras in E-SAI system.
\end{itemize}
}

\section{Methodology}\label{sec3-1}
The goal of E-SAI is to reconstruct the occlusion-free image from the collected event field. Fig.~\ref{pipeline} illustrates the overall pipeline of the proposed E-SAI algorithm, which consists of two main steps: \emph{refocusing} and \emph{reconstruction}. 
The purpose of refocusing is to align the signal events and scatter out the noise events, and we will address it in Sec.~\ref{sec:4-1}.
For the reconstruction, a hybrid SNN-CNN network is proposed to mitigate the disturbance of noise mentioned in Eq.~(\ref{ESAI}) and reconstruct the occluded scenes from the refocused event streams. The detailed description of the reconstruction network can be found in Sec.~\ref{sec3-2}.


\begin{figure}[t!]
	\centering
	\begin{subfigure}{1\linewidth}
	\centering
	    \begin{subfigure}[b]{0.42\linewidth}
			\includegraphics[width=\linewidth, trim={0cm 0 0cm 0}]{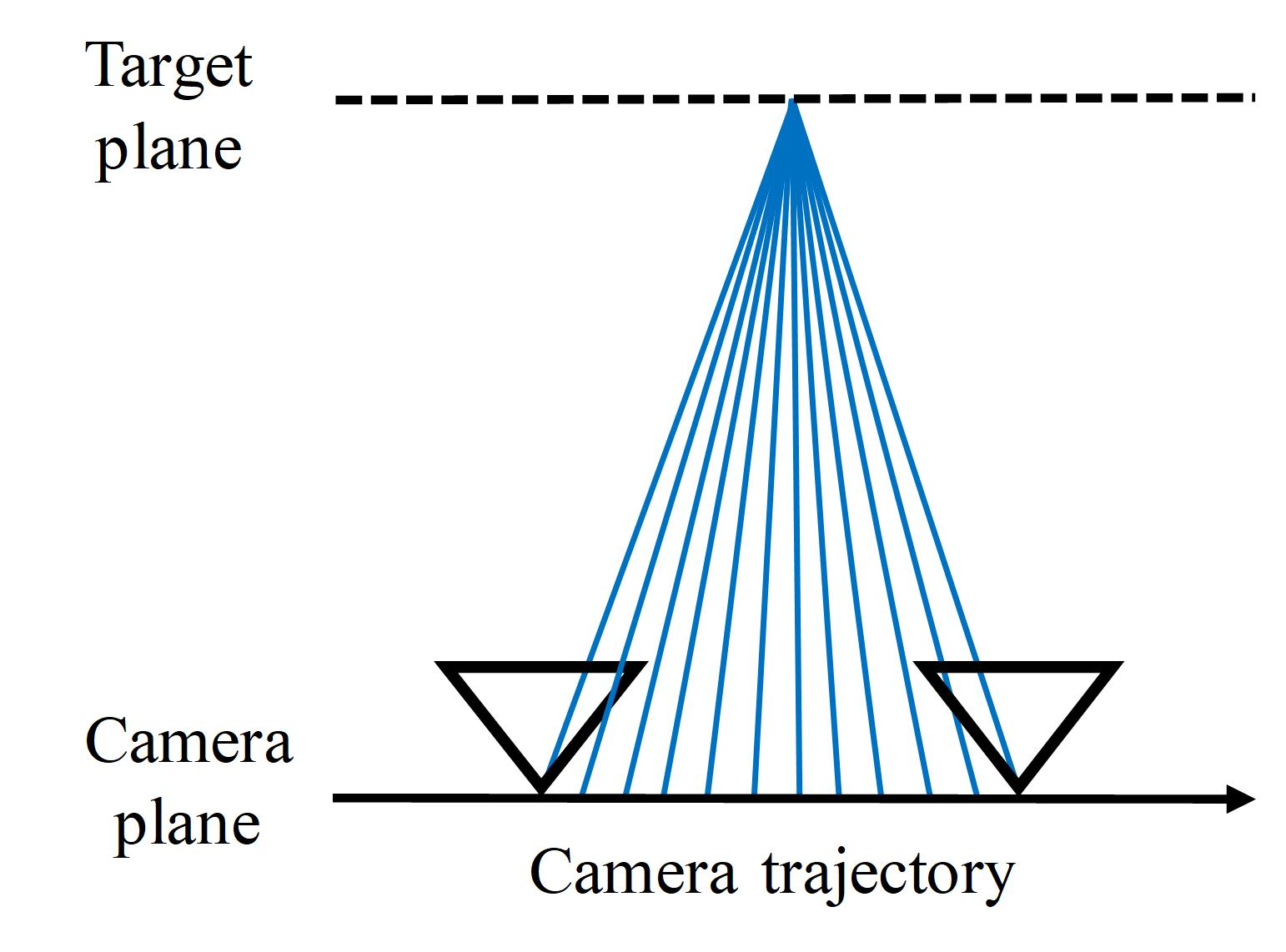}
			\vspace{-1.5em}
            \caption{\rmfamily \fontsize{8pt}{0}  Occlusion-free}
            \label{fig:EMVS-free}
		\end{subfigure}
		\begin{subfigure}[b]{0.42\linewidth}
			\includegraphics[width=\linewidth, trim={0cm 0 0cm 0}]{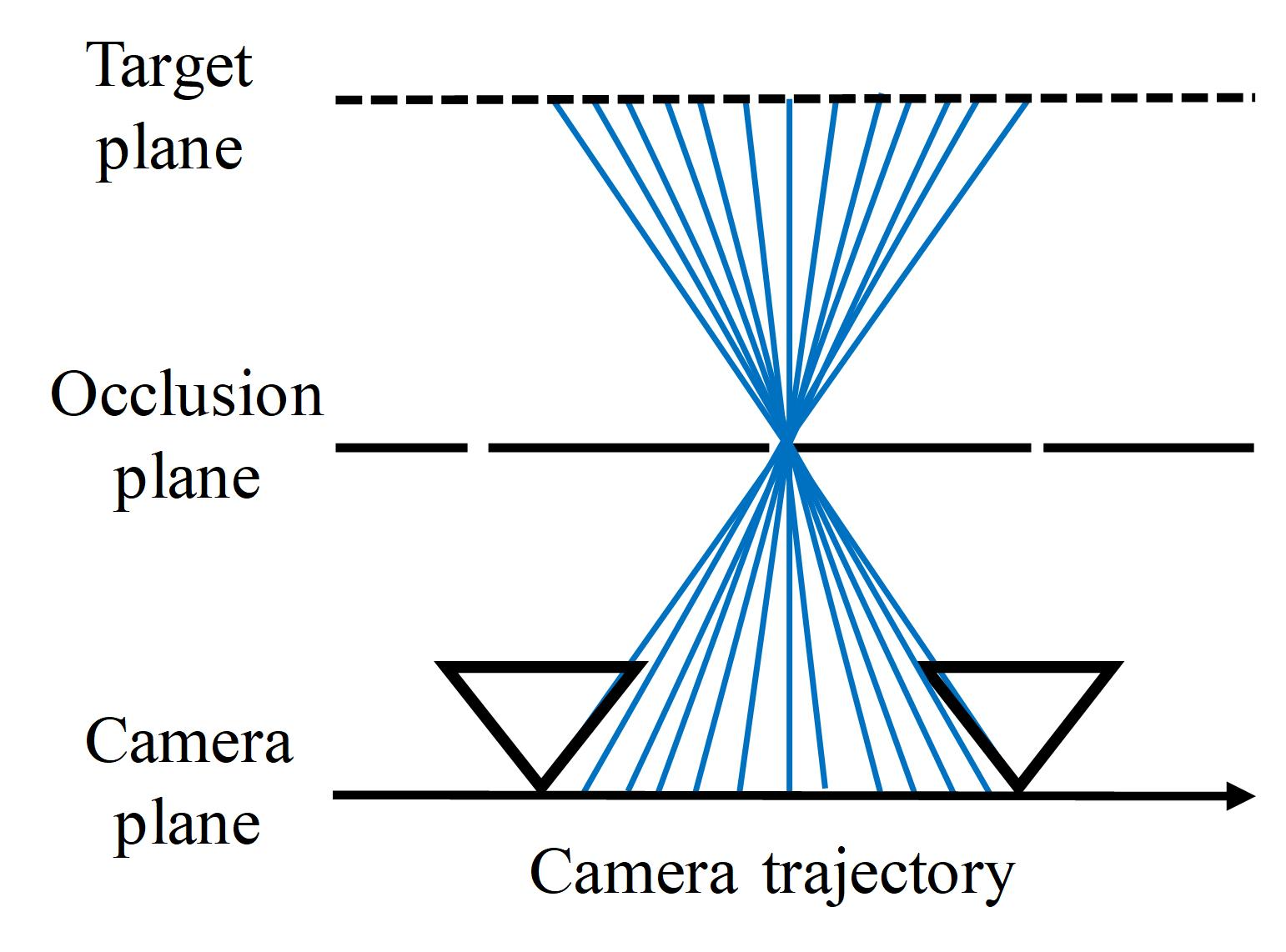}
			\vspace{-1.5em}
			\caption{\rmfamily \fontsize{8pt}{0}  Dense occlusion}
			\label{fig:EMVS-occ}
		\end{subfigure}
	\end{subfigure}	
	\begin{subfigure}{1\linewidth}
	\centering
		\begin{subfigure}[b]{0.24\linewidth}
			\includegraphics[width=\linewidth, trim={0cm 0 0cm 0}]{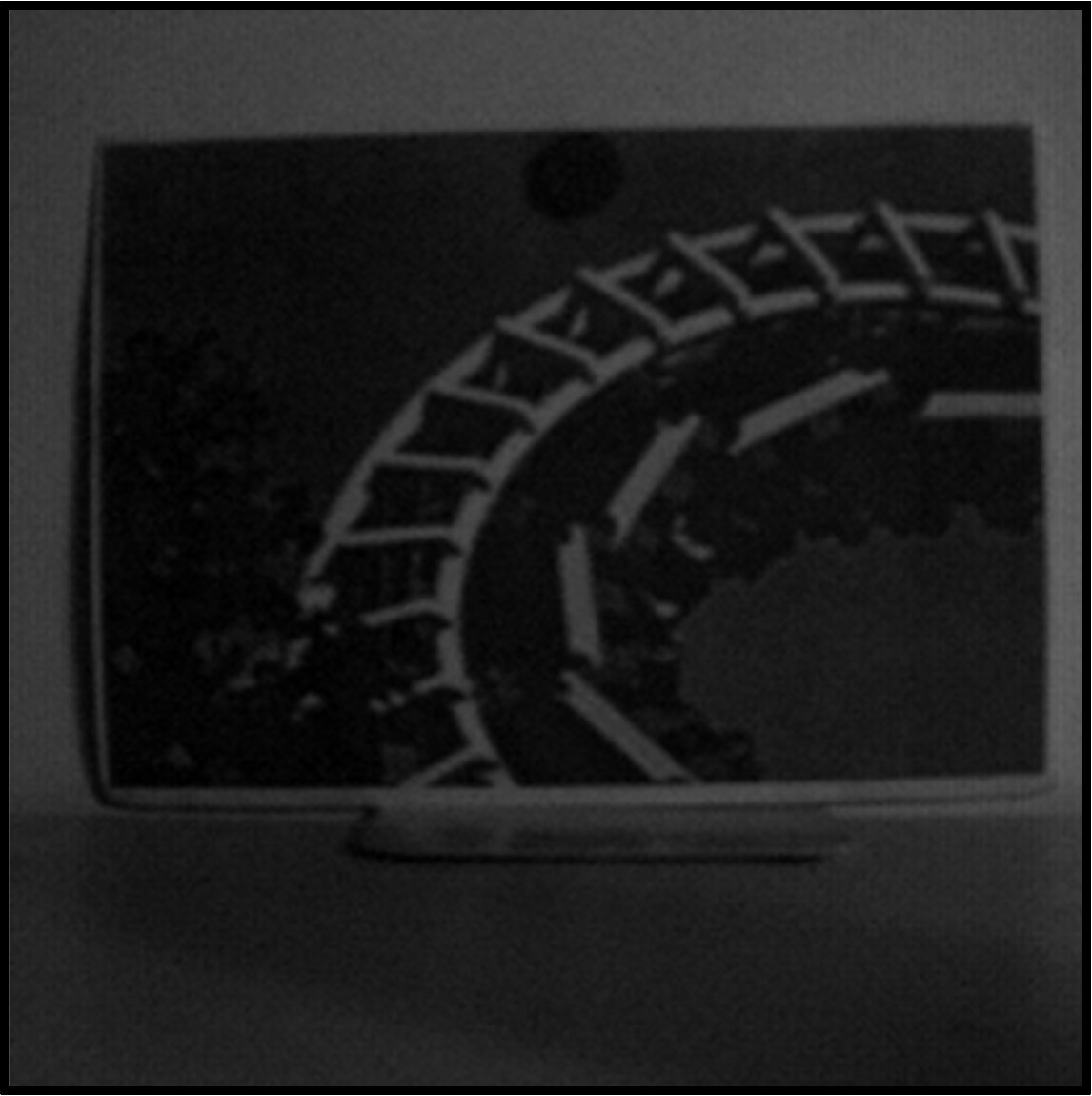}
			\vspace{-1.5em}
			\subcaption*{\scriptsize Reference}
		\end{subfigure}
		\begin{subfigure}[b]{0.24\linewidth}
			\includegraphics[width=\linewidth, trim={0cm 0 0cm 0}]{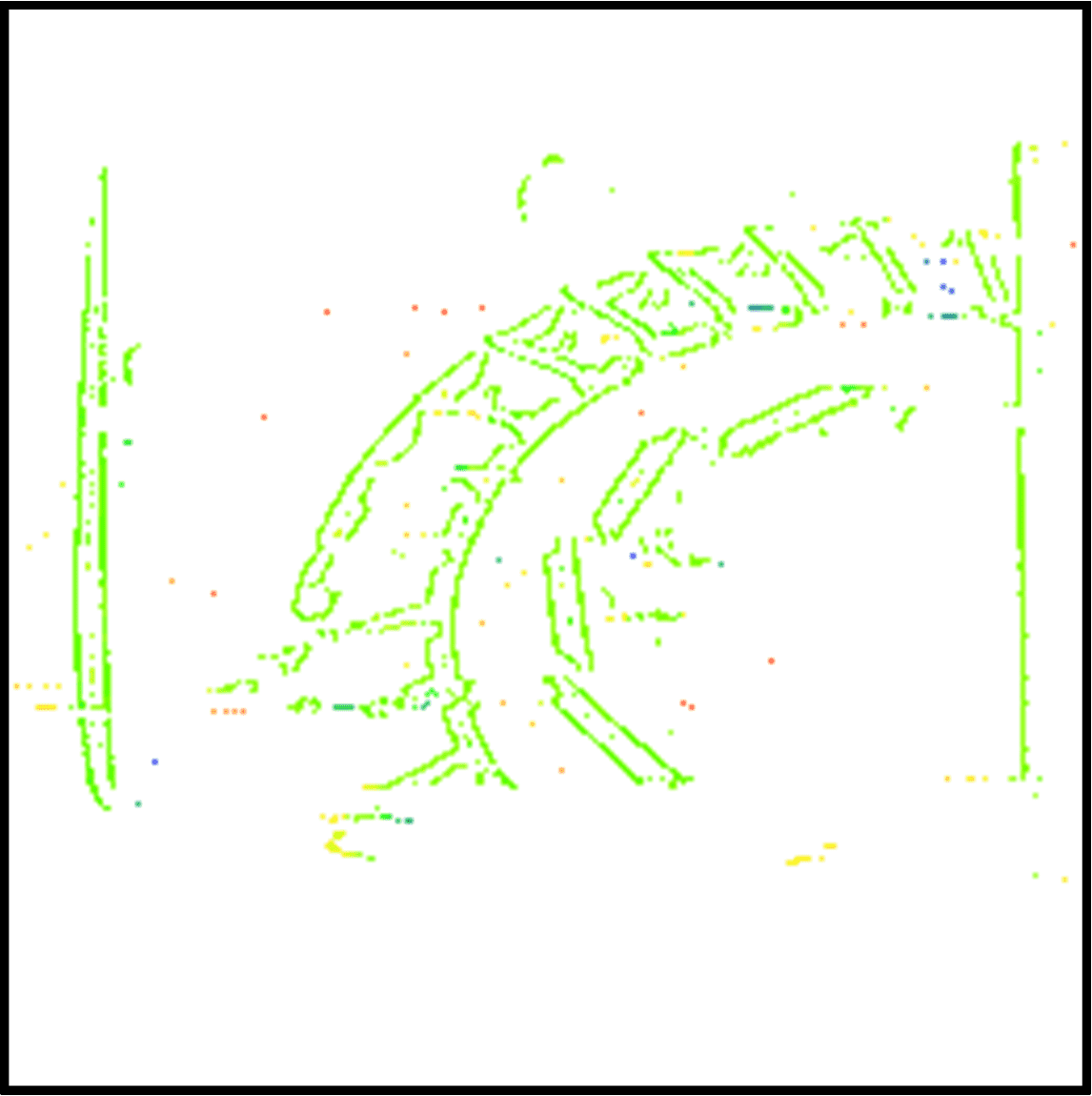}
			\vspace{-1.5em}
			\subcaption*{\scriptsize Depth w/o occ.}
		\end{subfigure}
		\begin{subfigure}[b]{0.24\linewidth}
			\includegraphics[width=\linewidth, trim={0cm 0 0cm 0}]{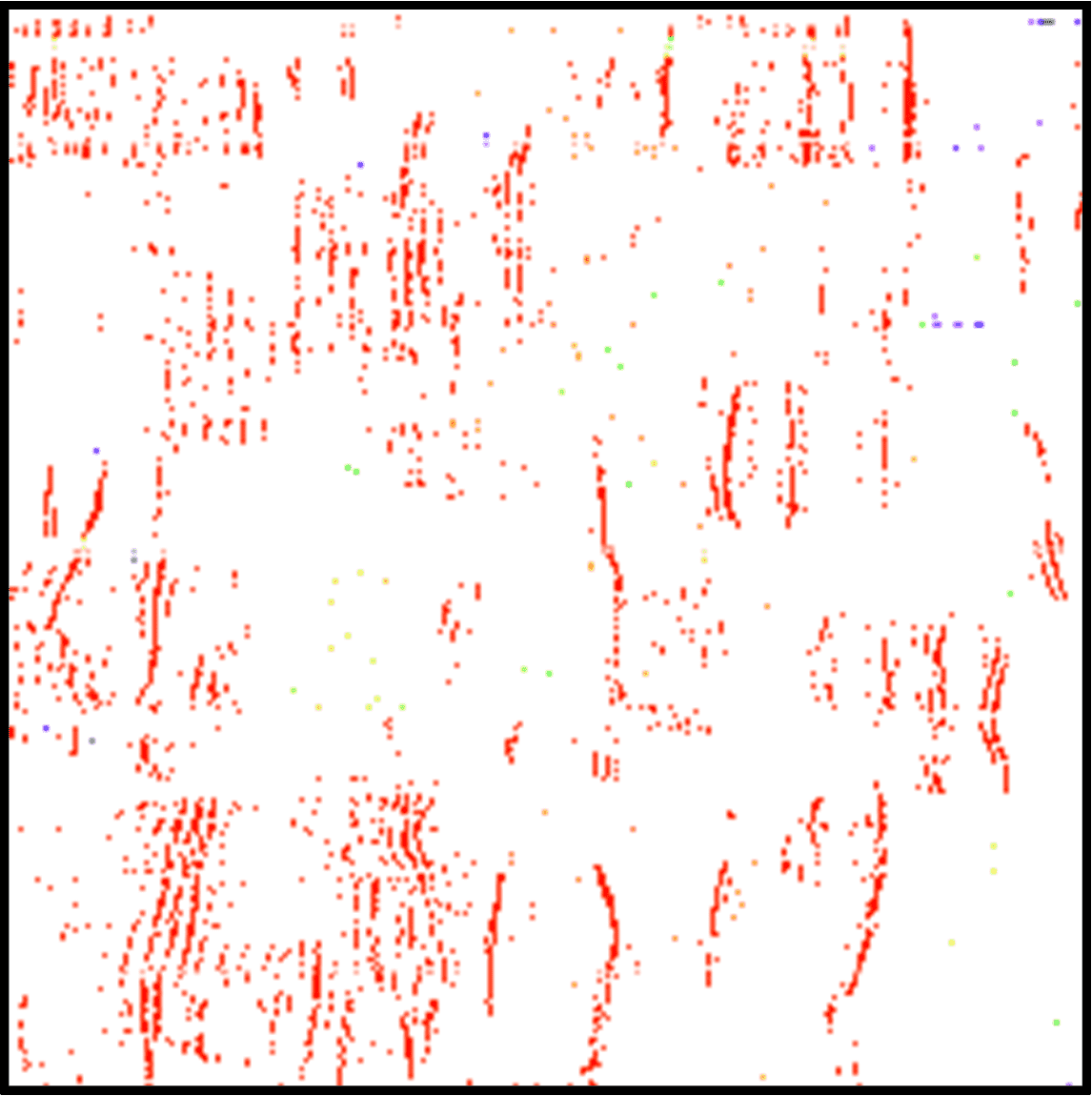}
			\vspace{-1.5em}
			\subcaption*{\scriptsize Depth w/ occ.}
		\end{subfigure}
		\begin{subfigure}[b]{0.24\linewidth}
			\includegraphics[width=\linewidth, trim={0cm 0 0cm 0}]{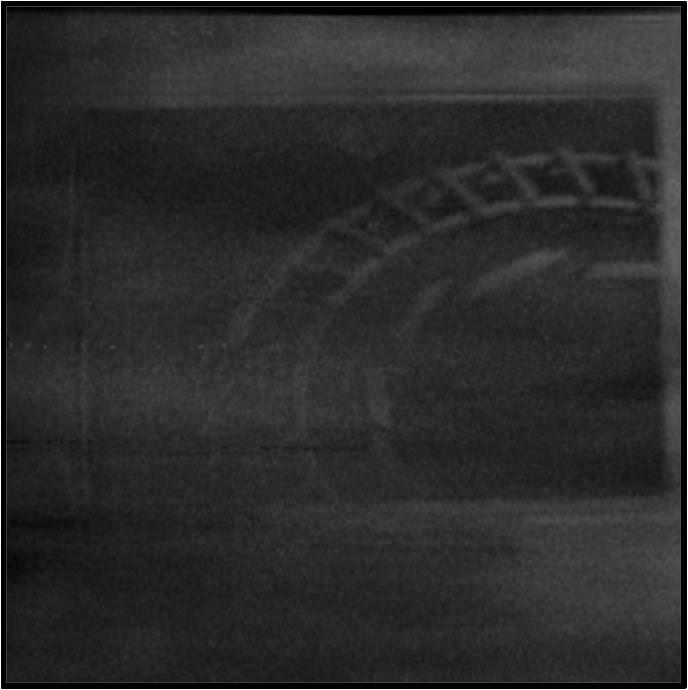}
			\vspace{-1.5em}
			\subcaption*{\scriptsize DSI w/ occ.}
		\end{subfigure}
		\caption{\rmfamily \fontsize{8pt}{0} EMVS results w/ and w/o occlusions}
		\label{fig:EMVS-res}
	\end{subfigure}
	\vspace{-.5em}
    \caption{
    {\color{\colored} 
    Performance of EMVS \cite{rebecq2018emvs} in occlusion-free and densely occluded scenes. (a) In occlusion-free scenes, rays back-projected from the events corresponding to the same target point successfully intersect on the target plane. (b) In densely occluded scene, rays back-projected from the events tend to intersect on the occlusion plane. {\color{\seccolored} (c)
    Results of EMVS in the scenes with (w/ occ.) and without (w/o occ.) dense occlusions, and the disparity space image (DSI) slice at the target depth is also depicted.}
   }
    }
    \label{fig:EMVS}
    \vspace{-1.5em}
\end{figure}

\subsection{Event Refocusing}\label{sec:4-1}
{\color{\colored}
Although the problem of event refocusing has been recently investigated~\cite{Gallego_2018_CVPR,rebecq2018emvs,gallego2019focus,stoffregen2019event,nunes2020entropy,xu2020a}, previous methods are mainly designed for sparsely occluded or occlusion-free scenes, and the dense foreground occlusion in our case brings new challenges to correctly refocusing on the background scenes. For example, EMVS \cite{rebecq2018emvs} can estimate the depth of scene points by locating the high-density regions where several viewing rays back-projected from events intersect.
Thus, EMVS\footnote{We use codes from \url{https://github.com/uzh-rpg/rpg_emvs}.} can easily locate the position of target points in occlusion-free scenes as shown in Figs.~\ref{fig:EMVS-free} and \ref{fig:EMVS-res}. 
However, under dense occlusions, EMVS tends to detect the structure of foreground occlusions instead of the background targets (Fig.~\ref{fig:EMVS-res}), since the occluded scene can only be sparsely and inconsistently observed and the rays back-projected from the acquired events mostly intersect on the occlusion plane as depicted in Fig.~\ref{fig:EMVS-occ}.
{\color{\seccolored}
To deal with this issue, this section presents an auto refocus method to adaptively align signal events under dense occlusions.
}
}

\begin{figure*}[t]
	\centering
	\begin{subfigure}{0.252\linewidth}
	\centering
		\includegraphics[width=\linewidth]{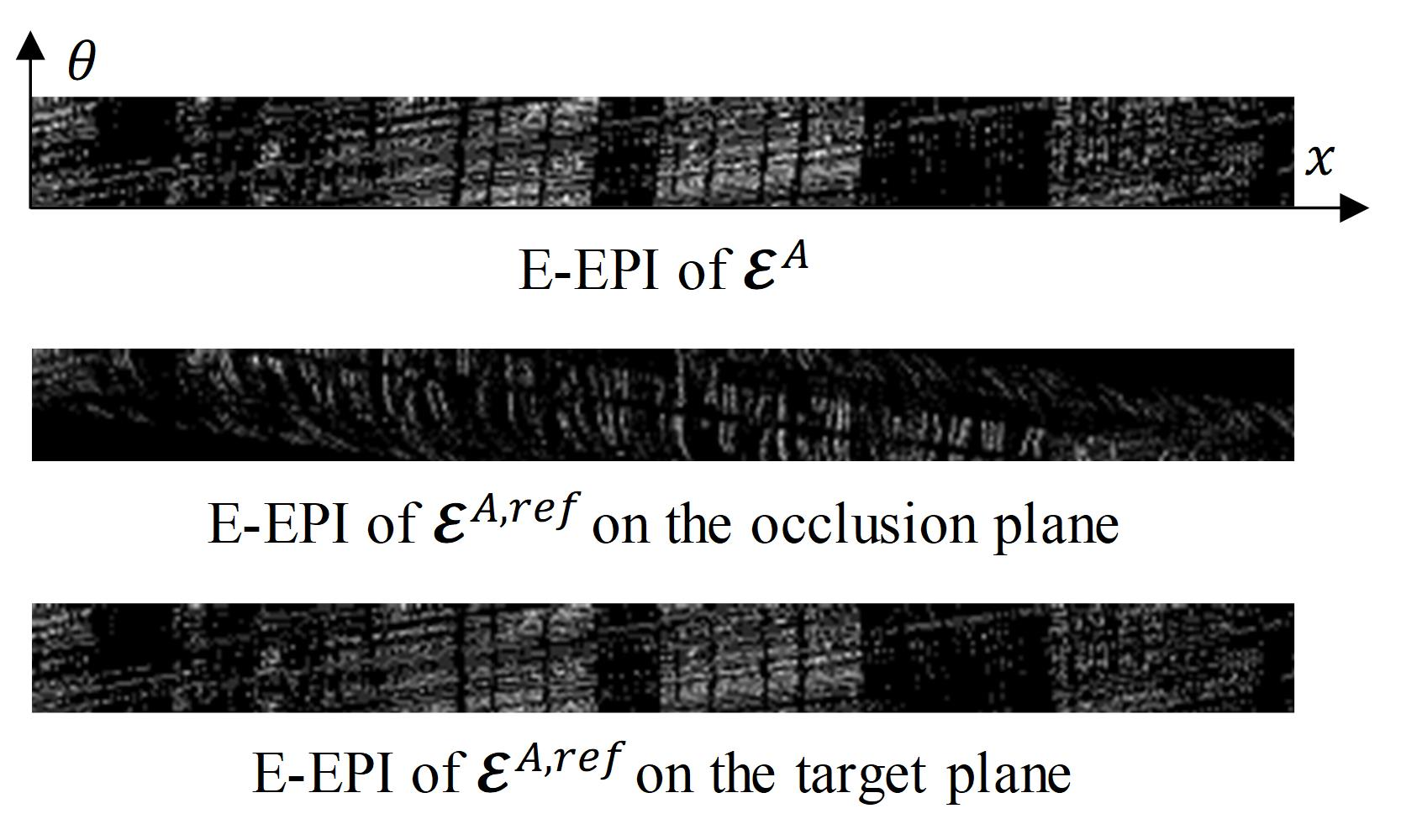}
		\caption{\rmfamily \fontsize{8pt}{0} E-EPIs of event fields}
		\label{refocusMethod-EPI}
	\end{subfigure}
	\begin{subfigure}{0.35\linewidth}
	\centering
		\includegraphics[width=.95\linewidth]{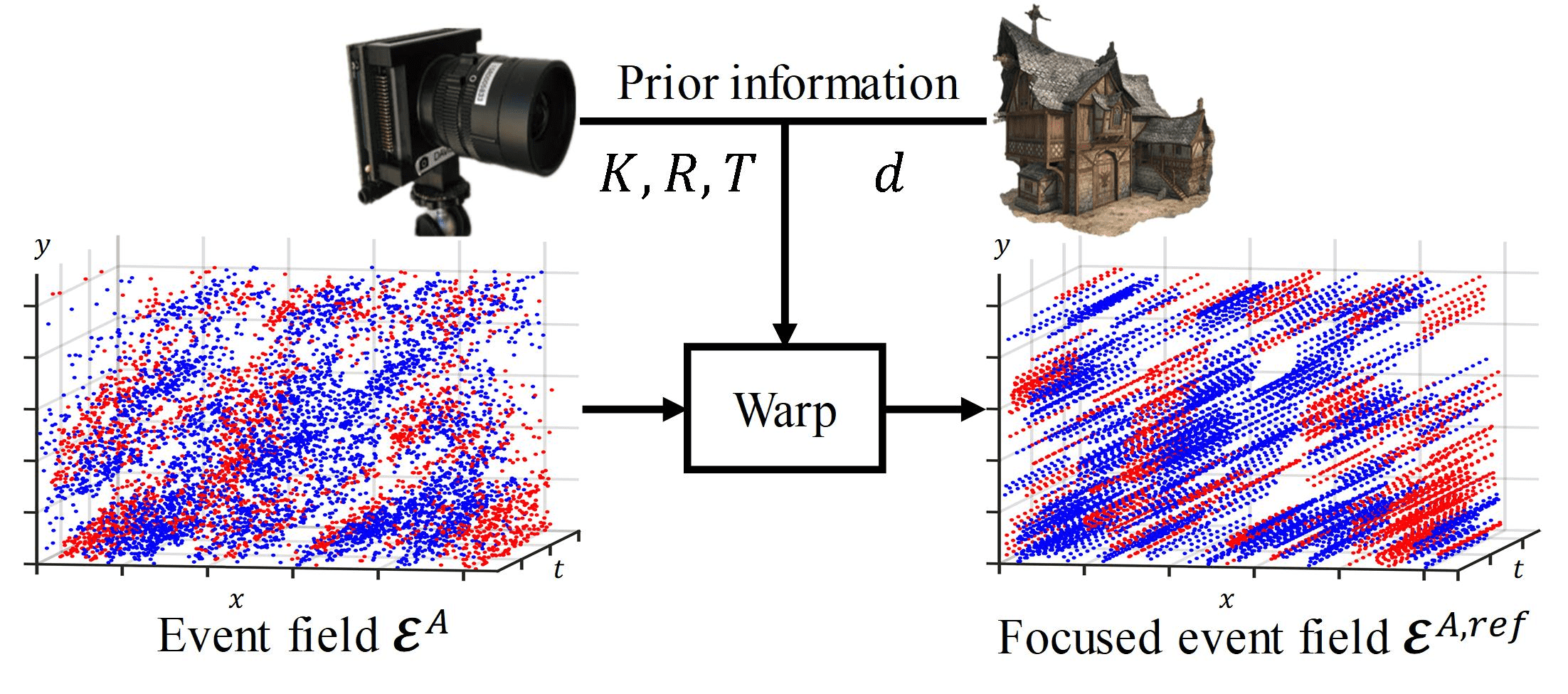}
		\caption{\rmfamily \fontsize{8pt}{0} Manual refocus method \cite{zhang2021event}}
		\label{refocusMethod1}
	\end{subfigure}
	\begin{subfigure}{0.35\linewidth}
	\centering
		\includegraphics[width=.95\linewidth]{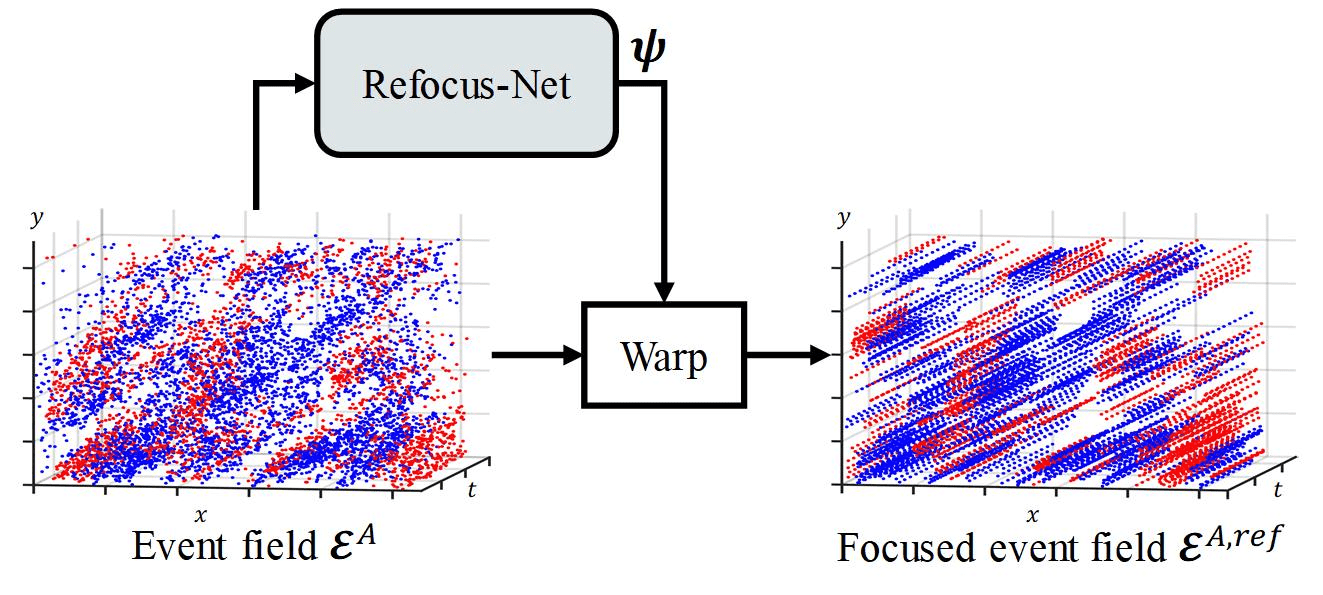}
		\caption{\rmfamily \fontsize{8pt}{0} Auto refocus method}
		\label{refocusMethod2}
	\end{subfigure}
	\vspace{-.5em}
	\caption{
	{\color{\colored}
	Illustration of event refocusing. (a) Based on the example in Fig.~\ref{fig:DiffEPI-common-5EPI}, we draw two E-EPIs of the event fields $\boldsymbol{\mathcal{E}}^{A,ref}$ refocused on the occlusion and target planes. Although some lines in the E-EPI on occlusion plane are not strictly vertical due to the non-uniform surface of our wooden fence occlusions, events are successfully aligned on both occlusion and target planes. However, E-EPI on occlusion plane is spatially sparse since events are aligned on occlusion edges, while E-EPI on target plane is spatially dense as events encode the texture information of occluded scenes.
	}
	(b) Manual refocus via Eq.~\eqref{trans-final} strongly relies on prior information of camera motions and target depth, while (c) auto refocus via spatial transformer (Refocus-Net) can adaptively align signal events, which largely facilitates the E-SAI for real-world scenarios.
	}
	\label{refocusMethod}
	\vspace{-1.5em}
\end{figure*}
\subsubsection{Auto Event Refocusing via Spatial Transformers}\label{sec:autoRefocus}
{\color{\seccolored}
We first define the warping process from the collected event field $\boldsymbol{\mathcal{E}}^{A}$ to the refocused event field $\boldsymbol{\mathcal{E}}^{A,ref}$ as 
\begin{equation}\label{eq-refocus2}
\boldsymbol{\mathcal{E}}^{A,ref} = \mathcal{W}\left(\boldsymbol{\mathcal{E}}^{A}, \boldsymbol{\psi} \right),
\end{equation}
which is achieved by spatial projection for each event from $\mathbf{x}_i$ to $\mathbf{x}_i^{ref}$ parameterized by parameter $\boldsymbol{\psi}$. Since the event camera is moving straightly with uniform velocity in our case, the refocusing formulation Eq.~\eqref{trans-final} can be simplified to
\begin{equation}\label{trans-final1}
\begin{aligned}
     \mathbf{x}_i^{ref} &= \mathbf{x}_{i}+\boldsymbol{\psi} \cdot (t_i-t^{ref}),
\end{aligned}
\end{equation}
where $\boldsymbol{\psi} = \frac{1}{d}[f_x \nu_x,\ f_y \nu_y]^\top$ is the coupled warping parameter with $f_x, f_y$ denoting the pixel focal length and $\nu_x, \nu_y$ indicating the camera speed in horizontal and vertical directions, respectively;
$t_i$ is the timestamp of the $i$-th event; $t^{ref}$ represents the timestamp when the camera is at reference viewpoint $\theta^{ref}$.
After refocusing with the provided warping parameter $\boldsymbol{\psi}$, the events triggered by target $A$ are successfully aligned, while others, \eg, the events generated by occlusions, are scattered out in both temporal and spatial dimensions, achieving a preliminary de-occlusion effect.
}
\par
However, the refocusing process Eq.~\eqref{trans-final1} heavily depends on the prior knowledge of the coupled warping parameter $\boldsymbol{\psi}$ associated with the camera motions and the depth of target scenes, which are difficult to be given in practice. In addition, the refocusing results are also sensitive to the accuracy of the target depth and camera motion, especially for close-view imaging (see Fig.~\ref{speed}). As a result, it is not easy to directly apply Eq.~(\ref{trans-final1}) for event refocusing in practice.
\par 
{\color{\colored}
Fortunately, in our setup with 1D uniform camera motion, we can facilitate event refocusing by analyzing E-EPIs as shown in Fig.~\ref{refocusMethod-EPI}, where spatially aligned events are parallel to the viewpoint dimension $\theta$ and appear vertical in E-EPI. Since a long baseline is often required in SAI systems to imitate the camera with a large aperture, which results in a very shallow depth-of-field, we only consider targets with small depth variation. Therefore, events can be spatially aligned when refocusing on either the occlusion plane or the target plane using the corresponding warping parameter $\boldsymbol{\psi}$. Furthermore, according to Eqs.~\eqref{eq:AA-OO} and \eqref{eq:OA}, events will be respectively aligned on edges and scene textures when refocusing on the occlusion plane and the target plane, leading to different spatial density in E-EPIs as shown in Fig.~\ref{refocusMethod-EPI}. 
Thus, the warping parameter $\boldsymbol{\psi}$ can be uniquely determined based on \textit{event alignment} and \textit{spatial density} in the E-EPI domain for refocusing on the target plane.
}
\par
Based on the above analysis, we propose to address event refocusing via learning-based methods. 
Previous works of spatial transformer networks (STNs) \cite{jaderberg2015spatial,lin2017inverse} have achieved spatial manipulation of data within networks, and the idea behind it can be also exploited for event refocusing. 
Following the methodology of STN \cite{jaderberg2015spatial}, we design a Refocus-Net 
to predict the coupled warping parameter $\boldsymbol{\psi}$
from the collected event field, \ie,
\begin{equation}
    \boldsymbol{\psi} = \operatorname{Refocus-Net}\left(\boldsymbol{\mathcal{E}}^{A}\right).
\end{equation}
Then, one can achieve event refocusing by warping the collected event field with the predicted parameter $\boldsymbol{\psi}$. 
And finally, the warping process Eq.~\eqref{eq-refocus2} can be relaxed,
\begin{equation}\label{auto-refocus-detail}
\boldsymbol{\mathcal{E}}^{A,ref}
=\mathcal{W}\left(\boldsymbol{\mathcal{E}}^{A}, \operatorname{Refocus-Net}\left(\boldsymbol{\mathcal{E}}^{A}\right)\right).
\end{equation}



\begin{figure*}[t]
	\centering
	\includegraphics[width=0.9\textwidth]{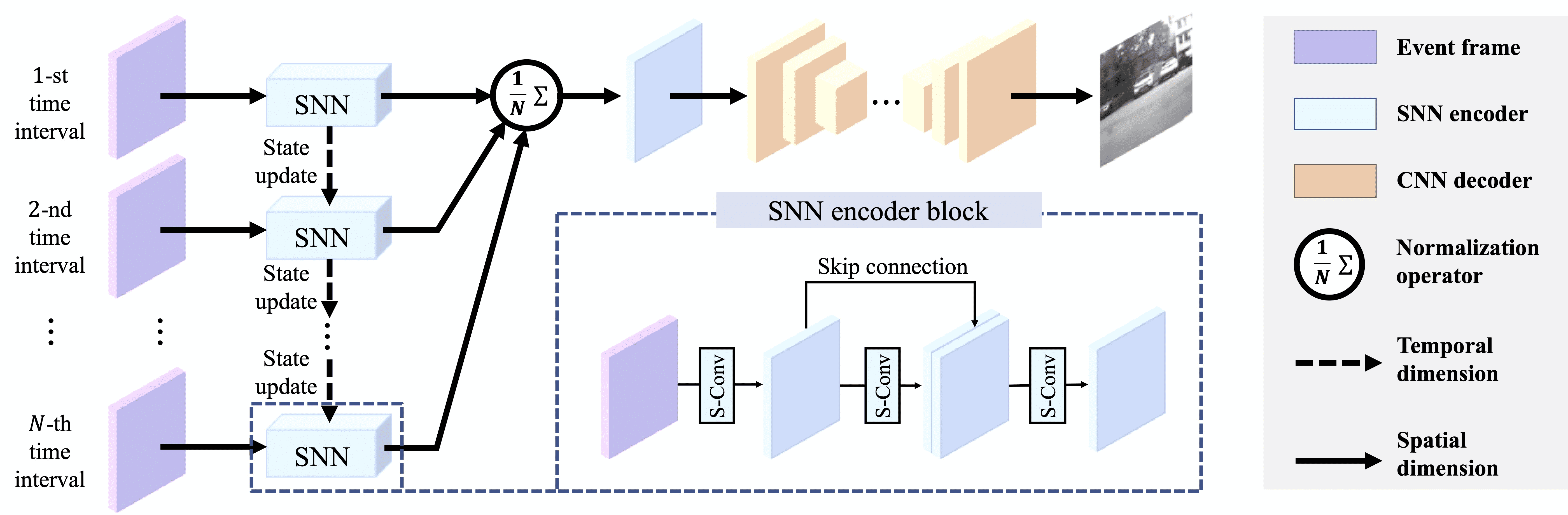}
	\vspace{-.5em}
	\caption{Structure of the hybrid SNN-CNN network. The spatio-temporal information of events is first encoded by SNN blocks, and then transformed to visual images by the CNN decoder.  To reduce the information loss of events, we add skip connections between the outputs of the 1$^{st}$ and 2$^{nd}$ spiking convolution (S-Conv) layers.}
	\label{Hyrbrid}
	\vspace{-1.5em}
\end{figure*}

\par
Comparing Figs.~\ref{refocusMethod1} and \ref{refocusMethod2}, applying Refocus-Net to event refocusing reduces reliance on prior information. Therefore, it is convenient when the target depth is hard to estimate or the camera motion cannot be acquired accurately, enhancing the practicality of E-SAI.

\subsubsection{Training Refocus-Net}\label{sec:train_rn}
Directly training the Refocus-Net with the ground truth warping parameter is usually difficult to converge due to the severe noise issue caused by occlusions. Instead, we propose to train the Refocus-Net with the aid of the reconstruction module. We first train a reconstruction network over the event streams refocused by Eq.~\eqref{trans-final1} with the ground truth $\boldsymbol{\psi}$. By fixing the weights of the reconstruction network, the Refocus-Net can then be trained with the supervision of the reconstruction module and the reference image. One can find more training details in Sec.~\ref{sec:train_detail}.

\subsection{Reconstruction with a Hybrid Network}\label{sec3-2}
According to Eq.~(\ref{eq:OA}), the brightness intensity of the occluded scene is closely related to the number of events. 
{\color{\seccolored} Thus the image of the occluded scene can be recovered by accumulating events after the refocusing process or counting the rays back-projected from events at the target depth, \eg, the disparity space image (DSI) of EMVS \cite{rebecq2018emvs} shown in Fig.~\ref{fig:EMVS-res}. However, the DSI slice is often noisy under dense occlusions as the rays back-projected from noise events $\mathcal{E}^{OO}_{\theta}$ are also counted without filtering.
}
Even though CNN-based methods can be further exploited to alleviate the noise problem, the temporal information inside events cannot be effectively used. Because of this, we propose a hybrid neural network composed of an SNN encoder and a CNN decoder, where both spatial and temporal information of events can be efficiently considered and utilized, as depicted in Fig.~\ref{Hyrbrid}.
\par
\subsubsection{SNN Encoder} 
\begin{figure}[t!]
	\centering
	\includegraphics[width=0.9\linewidth]{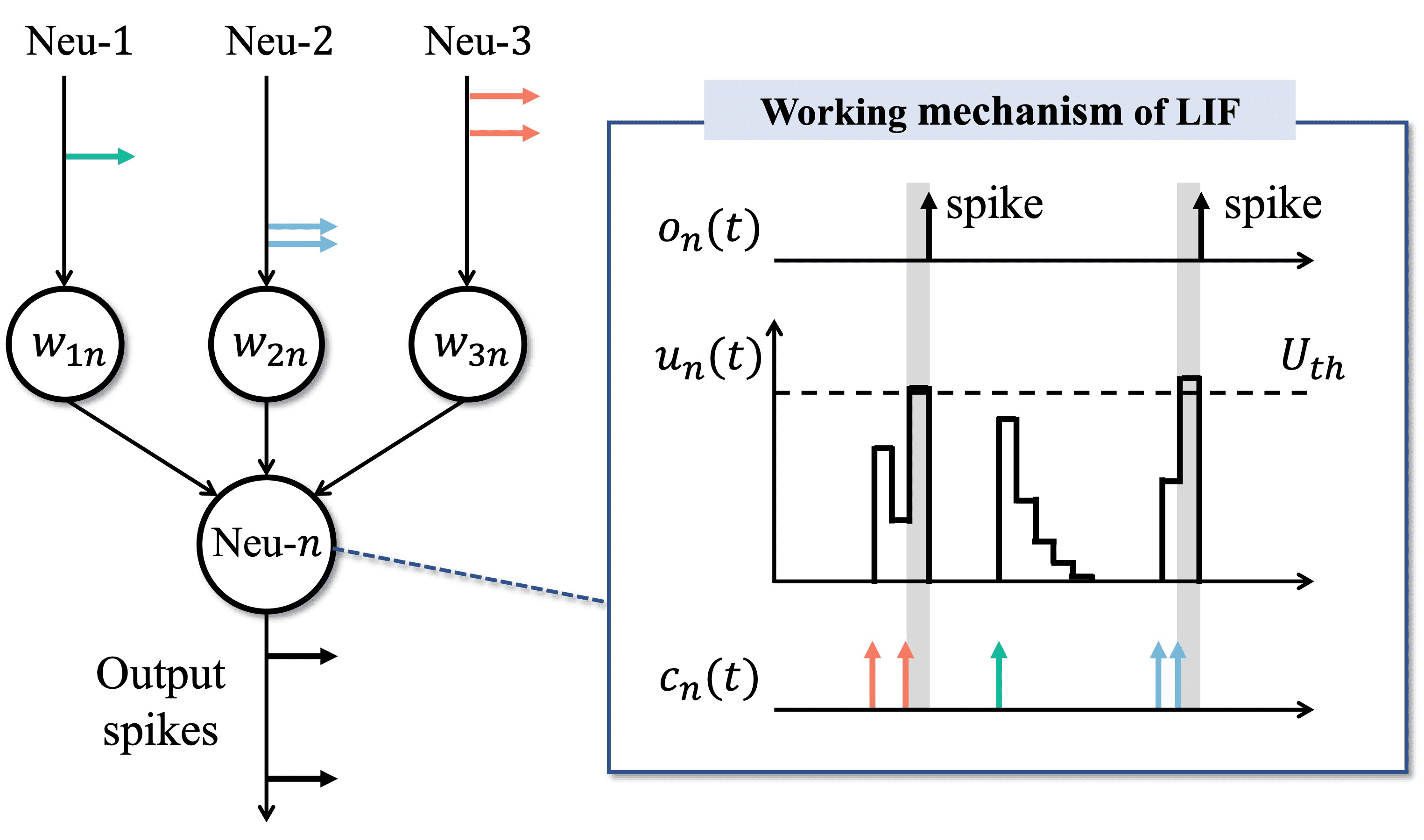}
	\vspace{-.5em}
	\caption{\rmfamily \fontsize{8pt}{0} An illustrative example of the LIF neuron and its working mechanism. The spikes from pre-synaptic neurons are first weighted and then fed into the target neuron-$n$, charging the internal membrane potential $u_n(t)$. Spikes will be fired whenever $u_n(t) > U_{th}$. Thanks to the spike firing and leakage mechanisms, the LIF neuron is able to filter out the isolated spikes, \eg, the noise events scattered out in spatial and temporal dimensions.}
	\label{LIFpic}
	\vspace{-1.5em}
\end{figure}

{\color{\colored}
Although the noise events are dispersed during the refocusing process, their presence still affects the quality of reconstruction.
To deal with it, we implement the SNN encoder using the leaky integrate-and-fire (LIF) model \cite{wuDirectTrainingSpiking2019}, where \textit{spike firing} and \textit{leakage} mechanisms contribute to noise suppression. 
Specifically, the membrane potential of LIF neurons is constantly leaking but occasionally charging when feeding with events, and spikes are fired as long as the membrane potential exceeds the spiking threshold. Thus temporally dense spikes are more likely to activate the LIF neurons than the isolated ones as shown in Fig.~\ref{LIFpic}, which enables LIF neurons to suppress dispersed noise events but preserve aligned signal events.

}

\par 
\noindent {\bf{LIF Neuron.}} Define $u_n^{l}(t)$ as the membrane potential of  the neuron-$n$ on the $l$-th layer at time $t$. The update of membrane potential can be described as
\begin{equation}\label{tmp-mem-up}
	u_n^{l}(t) = \alpha u_n^{l}(t-1) + c_n^{l}(t),
\end{equation}
where $\alpha\in[0,1]$ denotes the decay factor and $c_n^{l}(t)$ is the input current to neuron-$n$.  Considering the convolution operation in spiking layers, Eq.~(\ref{tmp-mem-up}) can be reformulated as:
\begin{equation}\label{mem-up}
	u_n^{l}(t) = \alpha u_n^{l}(t-1) + \sum_{m}w_{mn}o_m^{l-1}(t-1),
\end{equation}
where $o_m^{l-1}(t-1)$ represents the output spike of neuron-$m$ on the $(l-1)$-th layer at time $t-1$, and $w_{mn}$ denotes the synaptic weight between neuron-$m$ and neuron-$n$. Further, we add the reset \& fire mechanism into Eq.~(\ref{mem-up}),
\begin{equation}\label{LIF}
u_n^{l}(t) = \alpha u_n^{l}(t-1)(1-o_n^{l}(t-1)) + \sum_{m}w_{mn}o_m^{l-1}(t-1),
\end{equation}
where the output spike $o_n^{l}(t)$ is defined by
\begin{equation}\label{fire}
o_n^{l}(t) = \left\{
\begin{array}{ll}
1, & \text { if } u_n^{l}(t)>U_{th}, \\
0, & \text{otherwise},
\end{array}\right.
\end{equation}
and $U_{th}$ represents the spiking threshold. Eq.~(\ref{LIF}) indicates that the membrane potential of neuron-$n$ is affected by both its own state and the input spikes. If no new spikes are fed, the membrane potential $u_n^l(t)$ will leak at a certain rate related to the factor $\alpha$. In contrast, if the potential $u_n^l(t)$ is charged up to the spiking threshold $U_{th}$, the potential will be immediately reset to the resting potential $U_{rest}=0$ and simultaneously a spike will be emitted to other neurons. 

\par
\noindent{\bf{SNN Structure.}} 
{\color{\seccolored}
Although spiking neurons are able to directly process asynchronous event data when implemented on neuromorphic hardwares,
many supervised learning methods are based on the discretized frame representation by stacking events to facilitate SNN training \cite{lee2020spike,10.3389/fnins.2018.00331,wuDirectTrainingSpiking2019}. In this work, we mainly follow previous works \cite{10.3389/fnins.2018.00331,wuDirectTrainingSpiking2019} to build our SNN encoder. 
}
As illustrated in Fig.~\ref{Hyrbrid}, our SNN encoder is implemented with 3 neural layers composed of LIF neurons.
To make a balance between computational complexity and information integrity, we present a spatio-temporal representation for events. 
{\color{\seccolored}
The refocused event sequence is evenly divided into a pre-defined number of intervals $N$. In each interval, an event frame is generated by accumulating events over time, and each event frame contains two channels (positive and negative events).
}
Thus, every input group includes $N$ event frames, and the temporal relationship between event frames is retained. Over time, event frames sequentially pass through the spiking layers, and the membrane potential of spiking neurons updates between time intervals. After that, we generate the output of SNN encoder $\mathcal{O}_s$ by normalizing the output spike tensor over time
\begin{equation}
   \mathcal{O}_s  = \frac{1}{N} \sum_{t=1}^{N}  \mathbf{o}(t),
\end{equation}
where $\mathbf{o}(t)$ denotes the output spike tensor of the time interval $t$. 
Since noise events are scattered during refocusing, their influence can be gradually leaked out by the potential update of LIF neurons. Therefore, the noise issue is well alleviated, guaranteeing the reconstruction quality of occluded targets. To avoid the vanishing spike phenomenon in deep SNNs \cite{pandaScalableEfficientAccurate2020}, we instead implement the decoder with CNNs.

\subsubsection{CNN Decoder} Features extracted from the SNN encoder are then fed into a style-transfer network to reconstruct visual images. Here, we adopt the decoder architecture from the generator network used in \cite{zhu2017unpaired} which shows remarkable results in image style transferring, and adjust the kernel size of the output layer to fit the gray-scale images in our case. 
Benefiting from the hybrid structure, the spatio-temporal information of events can be fully utilized by the SNN encoder, and the occluded targets can be effectively reconstructed by the CNN decoder, guaranteeing the overall performance.

\subsubsection{Training Hybrid Network}
The synaptic weights in SNN can be trained in a supervised fashion via the spatio-temporal backpropagation (STBP) technique \cite{10.3389/fnins.2018.00331,wuDirectTrainingSpiking2019}, where the gradient of each pixel can be derived based on time intervals. And CNN can be trained via backpropagation (BP). Thus the SNN and CNN in the proposed hybrid network can be jointly trained. 
\par
To guide the training, we first exploit the idea of perceptual loss \cite{johnson2016perceptual} for high-level feature learning. 
With a pre-trained loss network $\phi$, we denote $\phi_k(X)$ as the output of the $k$-th convolution layer when network $\phi$ processes image $X$. Assume that $\phi_k(X)$ has the shape  $C_{k} \times H_{k}\times W_{k}$, we can formulate the perceptual loss $\mathcal{L}_{per}$ as:
\begin{equation}\label{percep}
	\mathcal{L}_{per}(Y,\hat{Y}) = \sum_{k} \frac{\lambda_k}{C_{k}H_{k}W_{k}} \|\phi_k(Y)-\phi_k(\hat{Y})\|_2^{2},
\end{equation}
where $Y$ represents the output of the hybrid network and $\hat{Y}$ is the corresponding ground truth; $\lambda_k$ denotes the weight of the $k$-th feature map. Rather than encouraging the pixel-wise match between images, the perceptual loss encourages the network to learn the similarity between high-level features, leading to better visual results.

\par 
In the pixel level, we add the pixel loss $\mathcal{L}_{pix}$ to maintain the similarity in low-level features like shape and texture. We express the pixel loss as:
\begin{equation}\label{pixel}
\mathcal{L}_{pix}(Y,\hat{Y}) =\frac{\|Y-\hat{Y}\|_1}{C H W},
\end{equation}
where $C \times H\times W$ represents the shape of $Y$ and $\hat{Y}$. Besides, the total variance loss $\mathcal{L}_{tv}(Y)$ in \cite{mahendran2015understanding} is exploited to encourage the spatial smoothness of reconstruction. Thus, the total loss can be summarized as follows.
\begin{equation}\label{total-loss}
\mathcal{L}(Y,\hat{Y}) = \beta_{per} \mathcal{L}_{per}(Y,\hat{Y}) + \beta_{pix} \mathcal{L}_{pix}(Y,\hat{Y}) + \beta_{tv} \mathcal{L}_{tv}(Y),
\end{equation} 
where $\beta_{per}, \beta_{pix}$ and $\beta_{tv}$ are the weights that control the importance of the corresponding loss function.

\begin{figure}[t!]
	\centering
	\includegraphics[width=0.85\linewidth, trim={0 0 0 40},clip]{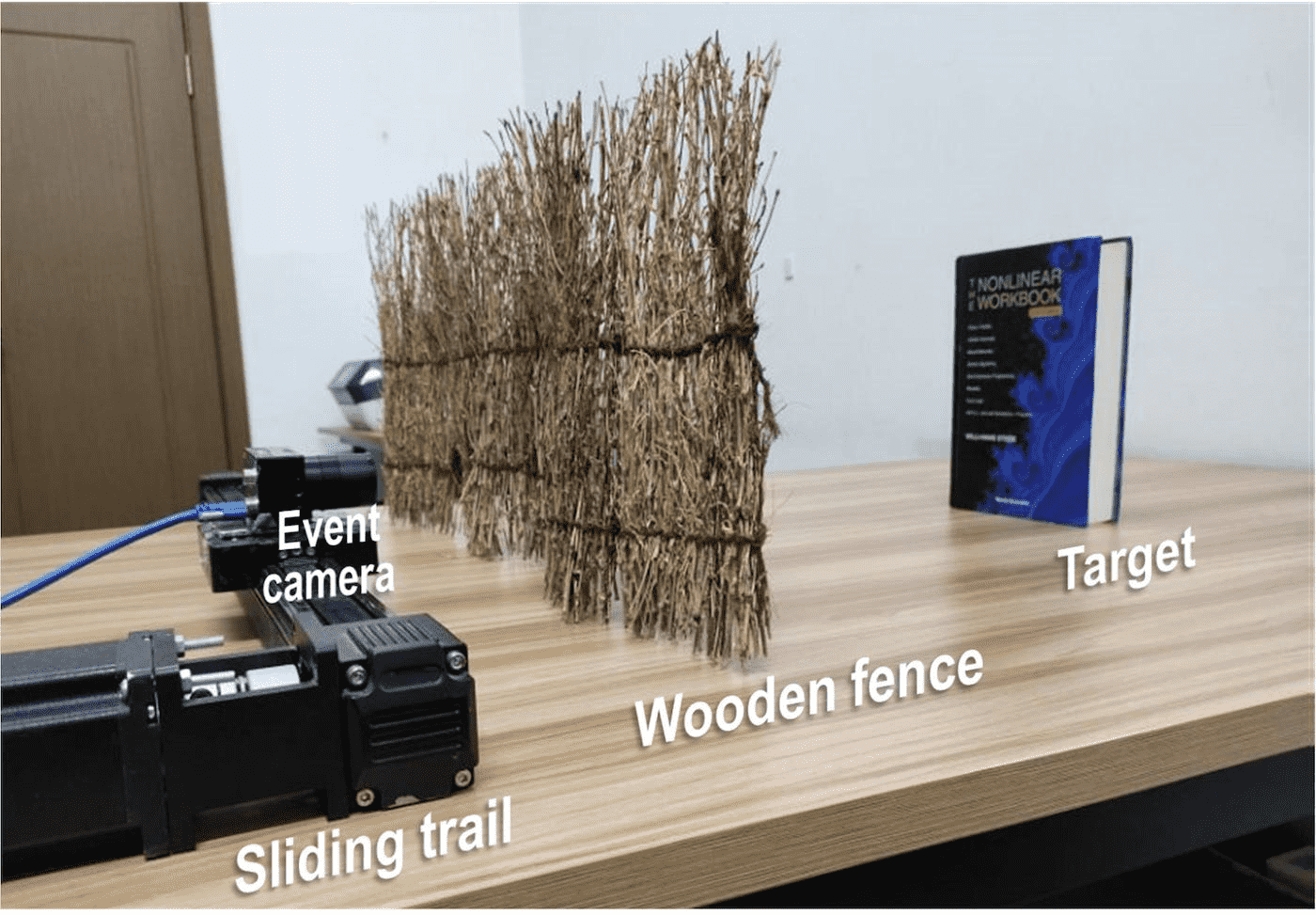}
	\vspace{-.5em}
	\caption{An example of our experimental setup: occlusions (the wooden fence), targets (the book) and an event camera installed on a programmable sliding trail.} 
	\label{scene}
	\vspace{-1.5em}
\end{figure}
\section{SAI Dataset}\label{chapter-5}
Due to the lack of available datasets for comparing F-SAI and E-SAI methods, we build a new SAI dataset containing both image frames and event streams. A DAVIS346 camera \cite{lichtsteiner128Times1282008} is employed for data collection since it can output both events and gray-scale active pixel sensor (APS) frames. 
\par 
As displayed in Fig.~\ref{scene}, we install the event camera on a programmable sliding trail and employ a wooden fence as dense occlusions. When the camera moves linearly on the sliding trail, the triggered events can be collected from different viewpoints, and APS frames are captured simultaneously by the DAVIS346 camera. We also collect the APS frames without occlusions (occlusion-free APS frames) for reference. Then event sequences and APS frames with occlusions are spatially matched with occlusion-free APS frames. To achieve this, we first generate an event 
frame by accumulating refocused events over time and then choose the occlusion-free image with the highest structural similarity (SSIM) \cite{wang2003multiscale} as the ground truth.
{\color{\colored}
All APS frames are collected under a fixed range of exposure time (25.0-38.4 ms) and each event stream experiences around 0.7 seconds under constant camera moving speed ($17.7$ cm/s).
}
\par

In the SAI dataset, a large variety of targets are considered, including 2D printed pictures and 3D objects in simple and complex real-world scenarios. They are occluded by the wooden fence to imitate the very dense occlusions, as shown in Fig.~\ref{scene}, and some of them are captured under extreme lighting conditions, \ie, under/over exposure scenes.
For clarity, we divide the SAI dataset into two main categories according to the shooting scenes: \emph{indoor} and \emph{outdoor}. The \emph{indoor} dataset contains printed pictures and simple objects, while the \emph{outdoor} dataset only contains real complex scenes. 
{\color{\colored}
For under exposure scenes, we first acquire the occluded frames and events in \emph{indoor} environments with the lights off, and then turn on the lights to capture the reference image. Regarding over exposure scenes, we collect the occluded frames and events using DAVIS346 in sunny \emph{outdoor} environments and capture the reference images with an iPhone 11 Pro in HDR mode. Thus, there is no spatially matched occlusion-free APS frame for the extreme lighting scenes due to the over/under exposure problem. }

\par 
In summary, the SAI dataset is built with 588 groups of data, including 488 groups for \emph{indoor} and 100 groups for \emph{outdoor}. 
To quantify the occlusion density in our SAI dataset, we first generate occlusion mask by subtracting the occlusion-free images from the corresponding occluded frames, and then calculate the proportion of occlusion pixels in the mask as occlusion density at the reference viewpoint. Overall, the occlusion density in our SAI dataset ranges from 73.5\% to 99.8\%, with the average value equal to 90.8\%. \textcolor{\seccolored}{
More details of our SAI dataset can be found in the supplementary material.}
And our SAI dataset is released at \url{https://dvs-whu.cn/projects/esai/}.

\begin{figure*}[t!]
	\centering
	\begin{subfigure}[b]{0.133\linewidth}
		\includegraphics[width=\linewidth, trim={0cm 0 0cm 0}]{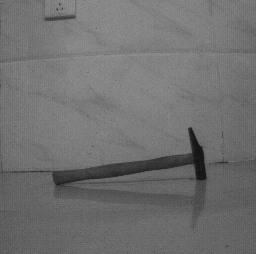}
		\vspace{-1em}\\
		\includegraphics[width=\linewidth, trim={0cm 0 0cm 0}]{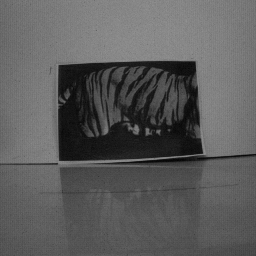}
		\vspace{-1em}\\
		\includegraphics[width=\linewidth, trim={0cm 0 0cm 0}]{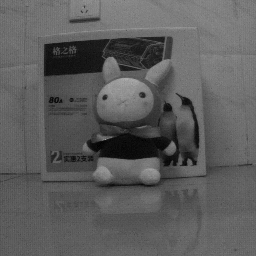}
		\vspace{-1.5em}
		\subcaption*{\scriptsize Reference}
	\end{subfigure}
	\begin{subfigure}[b]{0.133\linewidth}
		\includegraphics[width=\linewidth]{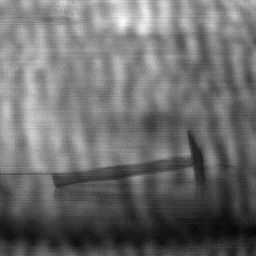}
		\vspace{-1em}\\
		\includegraphics[width=\linewidth, trim={0cm 0 0cm 0}]{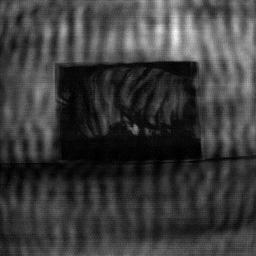}
		\vspace{-1em}\\
		\includegraphics[width=\linewidth, trim={0cm 0 0cm 0}]{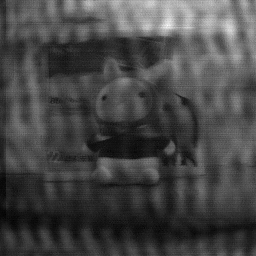}
		\vspace{-1.5em}
		\subcaption*{\scriptsize F-SAI+ACC}
	\end{subfigure}
	\begin{subfigure}[b]{0.133\linewidth}
		\includegraphics[width=\linewidth, trim={0cm 0 0cm 0}]{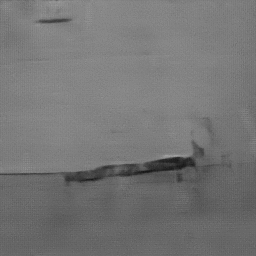}
		\vspace{-1em}\\
		\includegraphics[width=\linewidth, trim={0cm 0 0cm 0}]{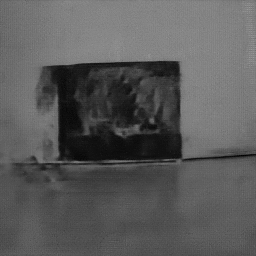}
		\vspace{-1em}\\
		\includegraphics[width=\linewidth, trim={0cm 0 0cm 0}]{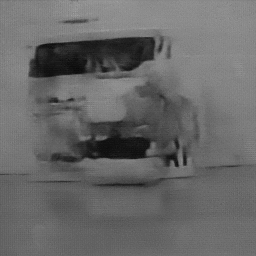}
		\vspace{-1.5em}
		\subcaption*{\scriptsize F-SAI+CNN}
	\end{subfigure}
	\begin{subfigure}[b]{0.133\linewidth}
		\includegraphics[width=\linewidth, trim={0cm 0 0cm 0}]{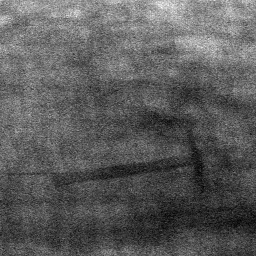}
		\vspace{-1em}\\
		\includegraphics[width=\linewidth, trim={0cm 0 0cm 0}]{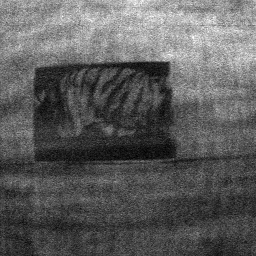}
		\vspace{-1em}\\
		\includegraphics[width=\linewidth, trim={0cm 0 0cm 0}]{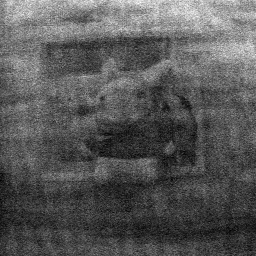}
		\vspace{-1.5em}
		\subcaption*{\scriptsize E-SAI+ACC}
	\end{subfigure}
	\begin{subfigure}[b]{0.133\linewidth}
		\includegraphics[width=\linewidth, trim={0cm 0 0cm 0}]{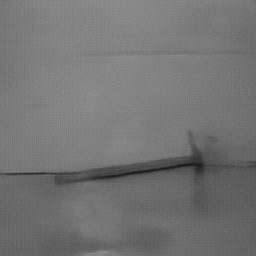}
		\vspace{-1em}\\
		\includegraphics[width=\linewidth, trim={0cm 0 0cm 0}]{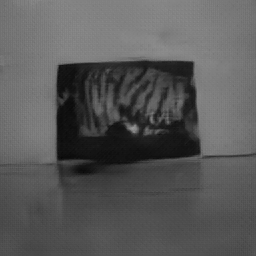}
		\vspace{-1em}\\
		\includegraphics[width=\linewidth, trim={0cm 0 0cm 0}]{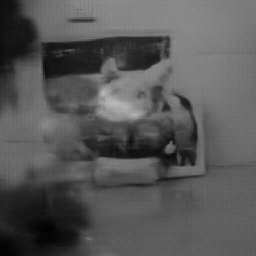}
		\vspace{-1.5em}
		\subcaption*{\scriptsize E-SAI+CNN}
	\end{subfigure}
	\begin{subfigure}[b]{0.133\linewidth}
		\includegraphics[width=\linewidth, trim={0cm 0 0cm 0}]{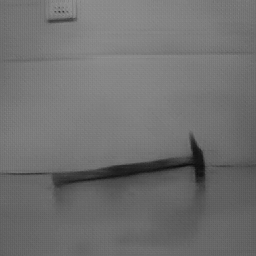}
		\vspace{-1em}\\
		\includegraphics[width=\linewidth, trim={0cm 0 0cm 0}]{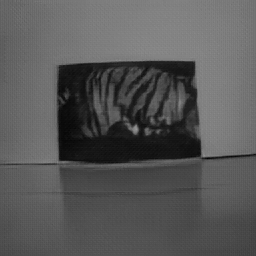}
		\vspace{-1em}\\
		\includegraphics[width=\linewidth, trim={0cm 0 0cm 0}]{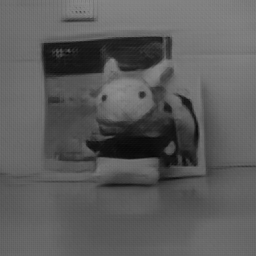}
		\vspace{-1.5em}
		\subcaption*{\scriptsize E-SAI+Hybrid (M)}
	\end{subfigure}
	\begin{subfigure}[b]{0.133\linewidth}
		\includegraphics[width=\linewidth, trim={0cm 0 0cm 0}]{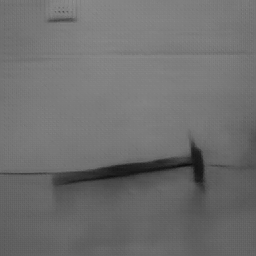}
		\vspace{-1em}\\
		\includegraphics[width=\linewidth, trim={0cm 0 0cm 0}]{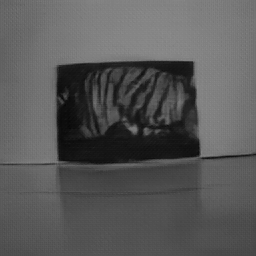}
		\vspace{-1em}\\
		\includegraphics[width=\linewidth, trim={0cm 0 0cm 0}]{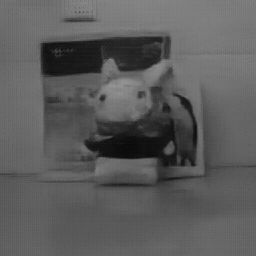}
		\vspace{-1.5em}
		\subcaption*{\scriptsize E-SAI+Hybrid (A)}
	\end{subfigure}
    \vspace{-.5em}
	\caption{Qualitative comparisons between F-SAI and E-SAI algorithms under very dense occlusions for \emph{indoor} dataset. } 
	\label{Indoor}
	\vspace{-1em}
\end{figure*}

\begin{figure*}[t!]
	\centering
	\begin{subfigure}[b]{0.133\linewidth}
		\includegraphics[width=\linewidth, trim={0cm 0 0cm 0}]{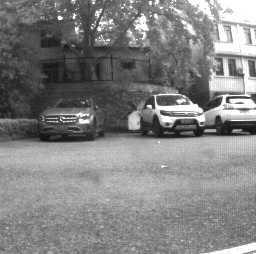}
		\vspace{-1em}\\
		\includegraphics[width=\linewidth, trim={0cm 0 0cm 0}]{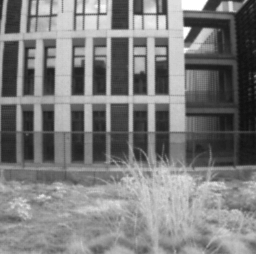}
		\vspace{-1em}\\
		\includegraphics[width=\linewidth, trim={0cm 0 0cm 0}]{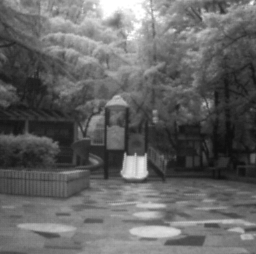}
		\vspace{-1.5em}
		\subcaption*{\scriptsize Reference}
	\end{subfigure}
	\begin{subfigure}[b]{0.133\linewidth}
		\includegraphics[width=\linewidth]{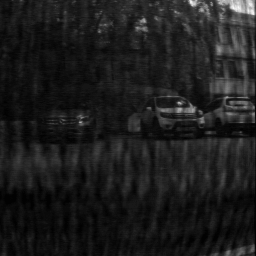}
		\vspace{-1em}\\
		\includegraphics[width=\linewidth, trim={0cm 0 0cm 0}]{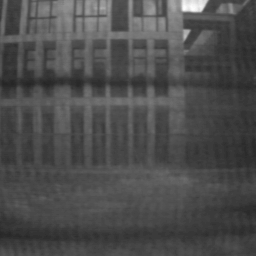}
		\vspace{-1em}\\
		\includegraphics[width=\linewidth, trim={0cm 0 0cm 0}]{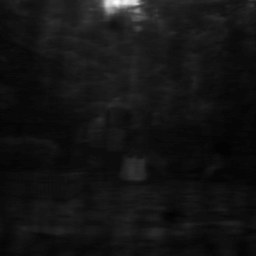}
		\vspace{-1.5em}
		\subcaption*{\scriptsize F-SAI+ACC}
	\end{subfigure}
	\begin{subfigure}[b]{0.133\linewidth}
		\includegraphics[width=\linewidth, trim={0cm 0 0cm 0}]{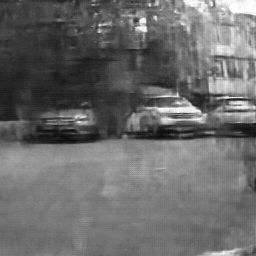}
		\vspace{-1em}\\
		\includegraphics[width=\linewidth, trim={0cm 0 0cm 0}]{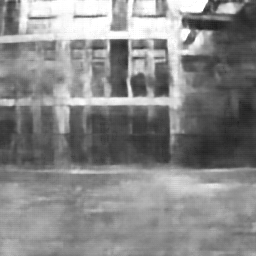}
		\vspace{-1em}\\
		\includegraphics[width=\linewidth, trim={0cm 0 0cm 0}]{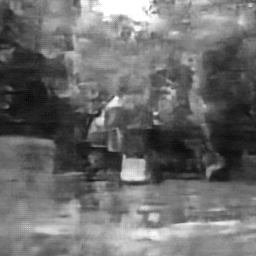}
		\vspace{-1.5em}
		\subcaption*{\scriptsize F-SAI+CNN}
	\end{subfigure}
	\begin{subfigure}[b]{0.133\linewidth}
		\includegraphics[width=\linewidth, trim={0cm 0 0cm 0}]{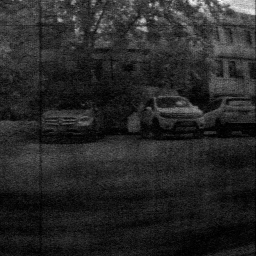}
		\vspace{-1em}\\
		\includegraphics[width=\linewidth, trim={0cm 0 0cm 0}]{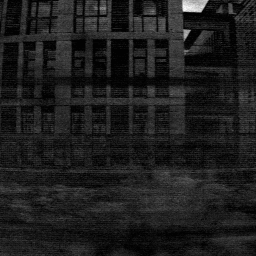}
		\vspace{-1em}\\
		\includegraphics[width=\linewidth, trim={0cm 0 0cm 0}]{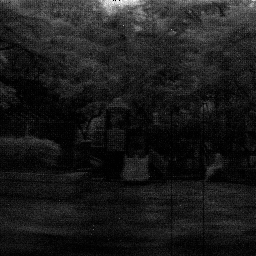}
		\vspace{-1.5em}
		\subcaption*{\scriptsize E-SAI+ACC}
	\end{subfigure}
	\begin{subfigure}[b]{0.133\linewidth}
		\includegraphics[width=\linewidth, trim={0cm 0 0cm 0}]{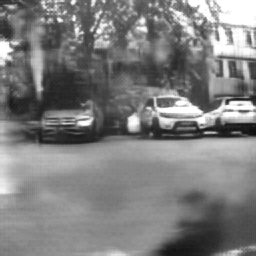}
		\vspace{-1em}\\
		\includegraphics[width=\linewidth, trim={0cm 0 0cm 0}]{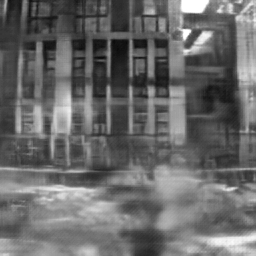}
		\vspace{-1em}\\
		\includegraphics[width=\linewidth, trim={0cm 0 0cm 0}]{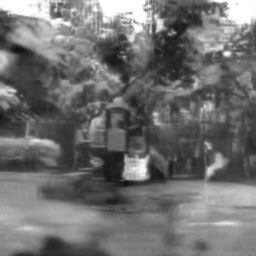}
		\vspace{-1.5em}
		\subcaption*{\scriptsize E-SAI+CNN}
	\end{subfigure}
	\begin{subfigure}[b]{0.133\linewidth}
		\includegraphics[width=\linewidth, trim={0cm 0 0cm 0}]{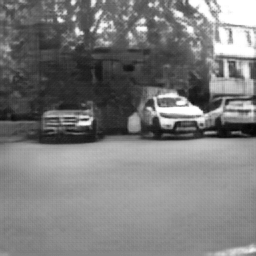}
		\vspace{-1em}\\
		\includegraphics[width=\linewidth, trim={0cm 0 0cm 0}]{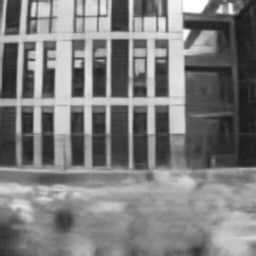}
		\vspace{-1em}\\
		\includegraphics[width=\linewidth, trim={0cm 0 0cm 0}]{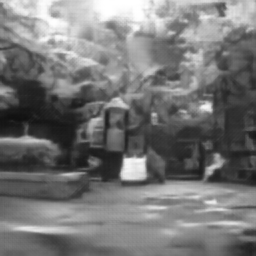}
		\vspace{-1.5em}
		\subcaption*{\scriptsize E-SAI+Hybrid (M)}
	\end{subfigure}
	\begin{subfigure}[b]{0.133\linewidth}
		\includegraphics[width=\linewidth, trim={0cm 0 0cm 0}]{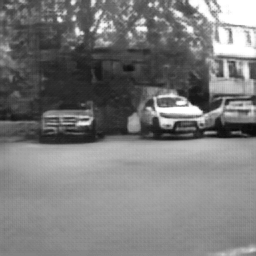}
		\vspace{-1em}\\
		\includegraphics[width=\linewidth, trim={0cm 0 0cm 0}]{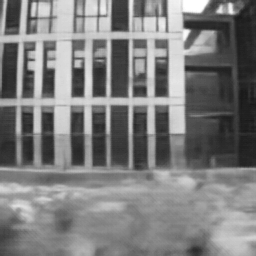}
		\vspace{-1em}\\
		\includegraphics[width=\linewidth, trim={0cm 0 0cm 0}]{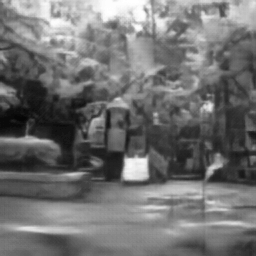}
		\vspace{-1.5em}
		\subcaption*{\scriptsize E-SAI+Hybrid (A)}
	\end{subfigure}
	\vspace{-.5em}
	\caption{Qualitative comparisons between F-SAI and E-SAI algorithms under very dense occlusions for \emph{outdoor} dataset. } 
	\label{Outdoor}
    \vspace{-1em}
\end{figure*}

\begin{figure*}[t!]
	\centering
	\begin{subfigure}[b]{0.225\linewidth}
        \begin{subfigure}{1\linewidth}
            \includegraphics[width=1\linewidth]{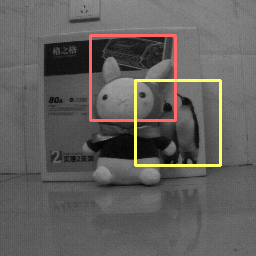}
        \end{subfigure}
		\vspace{.2em}
		\\
		\begin{subfigure}{1\linewidth}
            \includegraphics[width=1\linewidth]{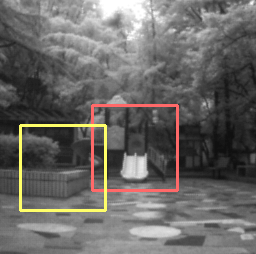}
        \end{subfigure}
		\vspace{-.5em}
		\subcaption*{\scriptsize Reference}
	\end{subfigure}
	\begin{subfigure}[b]{0.112\linewidth}
        \begin{subfigure}{1\linewidth}
            \includegraphics[width=1\linewidth]{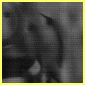}
            \\
            \includegraphics[width=1\linewidth]{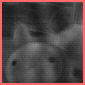}
        \end{subfigure}
		\vspace{.2em}
		\\
		\begin{subfigure}{1\linewidth}
            \includegraphics[width=1\linewidth]{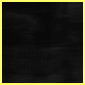}
            \\
            \includegraphics[width=1\linewidth]{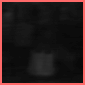}
        \end{subfigure}
		\vspace{-.5em}
		\subcaption*{\scriptsize F-SAI+ACC}
	\end{subfigure}
	\begin{subfigure}[b]{0.112\linewidth}
        \begin{subfigure}{1\linewidth}
            \includegraphics[width=1\linewidth]{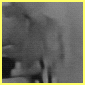}
            \\
            \includegraphics[width=1\linewidth]{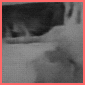}
        \end{subfigure}
		\vspace{.2em}
		\\
		\begin{subfigure}{1\linewidth}
            \includegraphics[width=1\linewidth]{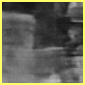}
            \\
            \includegraphics[width=1\linewidth]{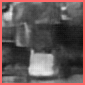}
        \end{subfigure}
		\vspace{-.5em}
		\subcaption*{\scriptsize F-SAI+CNN}
	\end{subfigure}
	\begin{subfigure}[b]{0.112\linewidth}
        \begin{subfigure}{1\linewidth}
            \includegraphics[width=1\linewidth]{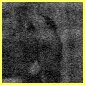}
            \\
            \includegraphics[width=1\linewidth]{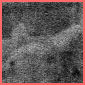}
        \end{subfigure}
		\vspace{.2em}
		\\
		\begin{subfigure}{1\linewidth}
            \includegraphics[width=1\linewidth]{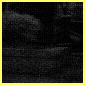}
            \\
            \includegraphics[width=1\linewidth]{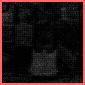}
        \end{subfigure}
		\vspace{-.5em}
		\subcaption*{\scriptsize E-SAI+ACC}
	\end{subfigure}
		\begin{subfigure}[b]{0.112\linewidth}
        \begin{subfigure}{1\linewidth}
            \includegraphics[width=1\linewidth]{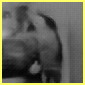}
            \\
            \includegraphics[width=1\linewidth]{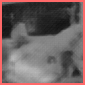}
        \end{subfigure}
		\vspace{.2em}
		\\
		\begin{subfigure}{1\linewidth}
            \includegraphics[width=1\linewidth]{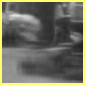}
            \\
            \includegraphics[width=1\linewidth]{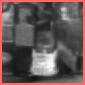}
        \end{subfigure}
		\vspace{-.5em}
		\subcaption*{\scriptsize E-SAI+CNN}
	\end{subfigure}
    	\begin{subfigure}[b]{0.112\linewidth}
        \begin{subfigure}{1\linewidth}
            \includegraphics[width=1\linewidth]{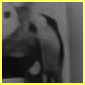}
            \\
            \includegraphics[width=1\linewidth]{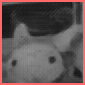}
        \end{subfigure}
		\vspace{.2em}
		\\
		\begin{subfigure}{1\linewidth}
            \includegraphics[width=1\linewidth]{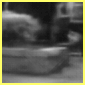}
            \\
            \includegraphics[width=1\linewidth]{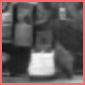}
        \end{subfigure}
		\vspace{-.5em}
		\subcaption*{\scriptsize E-SAI+Hybrid (M)}
	\end{subfigure}
		\begin{subfigure}[b]{0.112\linewidth}
        \begin{subfigure}{1\linewidth}
            \includegraphics[width=1\linewidth]{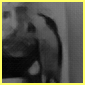}
            \\
            \includegraphics[width=1\linewidth]{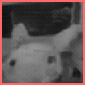}
        \end{subfigure}
		\vspace{.2em}
		\\
		\begin{subfigure}{1\linewidth}
            \includegraphics[width=1\linewidth]{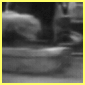}
            \\
            \includegraphics[width=1\linewidth]{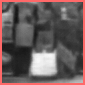}
        \end{subfigure}
		\vspace{-.5em}
		\subcaption*{\scriptsize E-SAI+Hybrid (A)}
	\end{subfigure}
	\vspace{-.5em}
	\caption{ Comparisons of F-SAI and E-SAI with different reconstruction methods. Details are zoomed in for better view.}
	\label{DetailVis}
	\vspace{-.5em}
\end{figure*}

\section{Experiments and Analysis}
This section evaluates and analyzes the proposed E-SAI method. In Sec.~\ref{sec:6-1}, we first present the experimental settings, including the evaluation prototype and implementation details. The performance of different F-SAI and E-SAI methods are then compared in Sec.~\ref{sec:6-2}, under dense occlusions and extreme lighting conditions. 
After that, we analyze the effectiveness of refocusing and reconstruction modules of our proposed E-SAI method respectively in Secs.~\ref{sec:exp-refocus} and \ref{sec:exp-recon}. Finally, we discuss the limitations of our E-SAI method and the related future work in Sec.~\ref{sec:exp-limitations}.

\subsection{Experimental Settings} \label{sec:6-1}
\subsubsection{Evaluation Prototype} 
For the frame-based SAI, we employ a representative traditional method \cite{vaishUsingPlaneParallax2004} (\textbf{F-SAI+ACC}) and the recent state-of-the-art learning-based method \cite{wangDeOccNetLearningSee2019} (\textbf{F-SAI+CNN}) for comparison. We use $30$ APS frames with occlusions for reconstruction in F-SAI+ACC and stack them in chronological order as the input of F-SAI+CNN. For the event-based SAI, we compare three different reconstruction methods, including the accumulation method (\textbf{E-SAI+ACC}), CNN-based method (\textbf{E-SAI+CNN}), and the proposed hybrid network (\textbf{E-SAI+Hybrid}). We denote E-SAI+Hybrid with manual refocus and auto refocus methods as E-SAI+Hybrid (M) and E-SAI+Hybrid (A), respectively. 
The detailed network architecture is provided in the supplementary material.

{\color{\colored}
For E-SAI+ACC, we directly accumulate both positive and negative events along the time dimension after refocusing on the target plane. Specifically, for the refocused event field $\boldsymbol{\mathcal{E}}^{A,ref}$, we define $\boldsymbol{\mathcal{E}}^{A,ref}({\bf x})\triangleq \left\{e_i|e_i\in \boldsymbol{\mathcal{E}}^{A,ref}, {\bf x}_i={\bf x}\right\}$ as the set of events at pixel ${\bf x}$ and generate the event frame $
        {E}({\bf x}) \triangleq \left\vert \mathcal{E}^{A,ref}({\bf x}) \right\vert.
    $
    Due to the lack of event threshold $\eta$, we further apply a min-max normalization to obtain the E-SAI+ACC results,
     $$
    E_{\textnormal{ACC}} =\frac{{E}-\operatorname{min}({E})}{\operatorname{max}({E})-\operatorname{min}({E})}.
    $$

}
Regarding the E-SAI+CNN method, the refocused event frames are stacked as a $2N$-channel tensor (corresponding to $2$ polarities) for network input. For the sake of fairness, we also build a pure CNN counterpart by simply replacing the SNN encoder with a 3-layer CNN, with the same number of network parameters as the SNN encoder. Therefore, we can evaluate the effectiveness of the SNN encoder by comparing E-SAI+Hybrid with E-SAI+CNN. 

\subsubsection{Training Details}\label{sec:train_detail}
Networks implemented in this paper are all trained on NVIDIA GeForce RTX 2080 Ti GPUs with batch size 12 for around 500 epochs, and the initializer in \cite{he2015delving} is applied. The Adam optimizer \cite{kingma2014adam} is used with the initial learning rate setting to $5\times {10}^{-4}$ and the SGDR schedule \cite{loshchilov2016sgdr} by setting $T_{max}=64$ (reset the learning rate every 64 epochs).

{\color{\colored}
\noindent{\bf Data Augmentation.} We choose $90\%$ scenes of the SAI dataset for training and leave the rest for the testing phase. 
In training reconstruction networks, we apply flipping (horizontal, vertical, and horizontal-vertical) and random rotating (random angles ranging from -15 to 15 degrees) to ground truth images and the refocused event frames for data augmentation. Meanwhile, the reflection padding technique is applied to reduce boundary artifacts. For the training of Refocus-Net, we only perform flipping for augmentation since rotation will break the parallax structure. And the horizontal flipping is performed together with the temporal order flipping of event streams to maintain the original parallax structure of event field.
}

\noindent{\bf F-SAI+CNN.} The pre-trained models of F-SAI+CNN in \cite{wangDeOccNetLearningSee2019} are oriented to a two-dimensional camera array. Sliding the camera in our experimental setting corresponds to a $1\times 30$ camera array. For fair comparisons, we re-train F-SAI+CNN over our SAI dataset with the official codes\footnote{Official codes at \url{https://github.com/YingqianWang/DeOccNet}.}. 

\noindent{\bf E-SAI Networks.} To train E-SAI+CNN and E-SAI+Hybrid, we divide each event sequence into 30 slices with equal time intervals, \ie, $N=30$, and set the loss weights as $\left[\beta_{per},\beta_{pix},\beta_{tv}\right]=\left[1,32,2\times {10}^{-4}\right]$. The 16-layer VGG network \cite{simonyan2014very} pre-trained on the ImageNet dataset \cite{russakovsky2015imagenet} is employed to calculate the perceptual loss based on the 2-nd, 4-th, 7-th, and 10-th convolution layers and the corresponding loss weights are respectively $\left[\lambda_2,\lambda_4,\lambda_7,\lambda_{10}\right]=\left[1\times {10}^{-1},1/21,10/21,10/21\right]$.

\begin{figure*}[t!]
	\centering
	\begin{subfigure}{0.332\linewidth}
		\includegraphics[width=.87\linewidth,height=.72\linewidth]{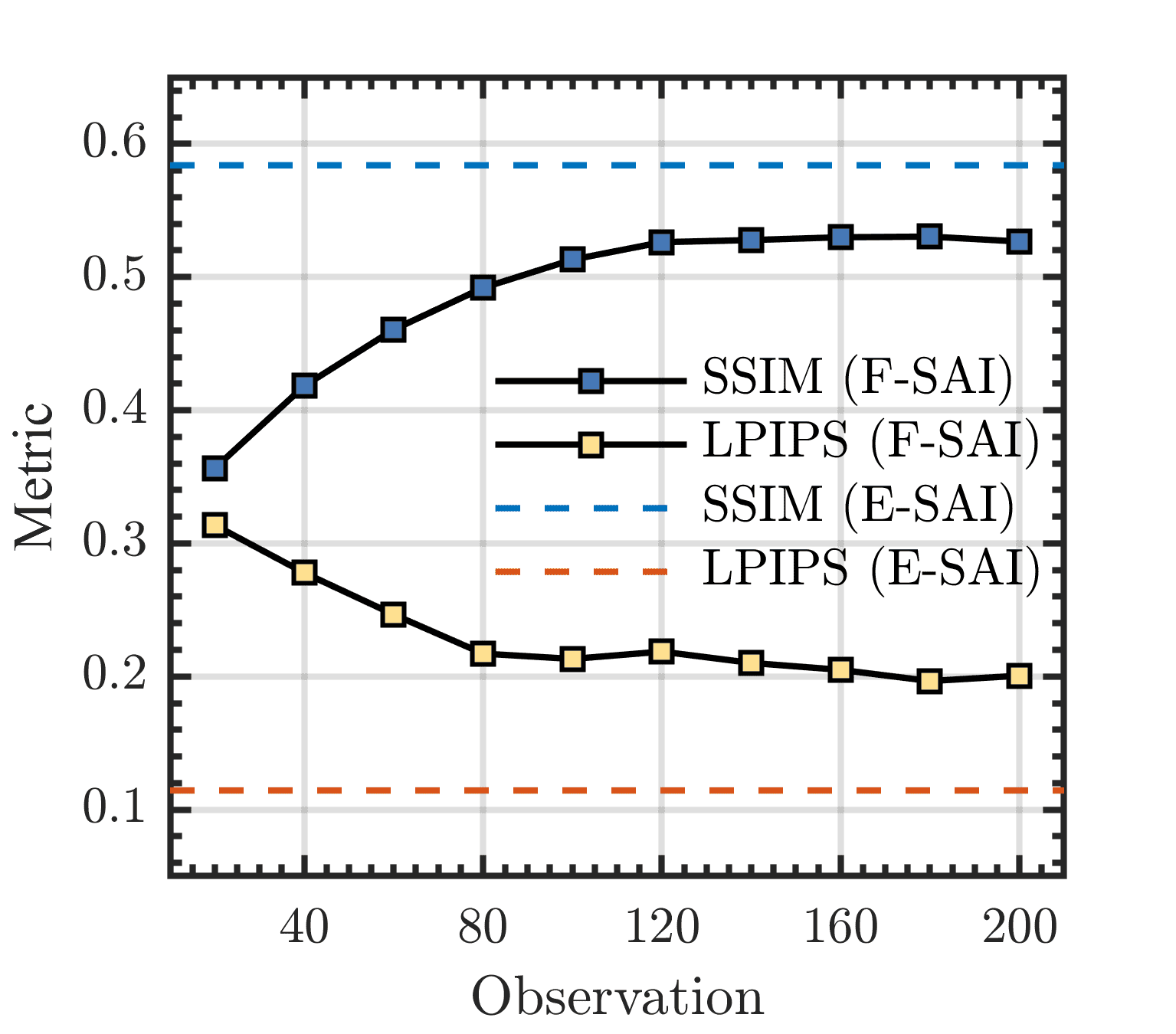}
		\caption{\rmfamily \fontsize{8pt}{0} Quantitative results}
		\label{fig:Slow-quan}
	\end{subfigure}	
	\begin{subfigure}{0.6\linewidth}
		\begin{subfigure}[b]{0.18\linewidth}
			\includegraphics[width=\linewidth, trim={0cm 0 0cm 0}]{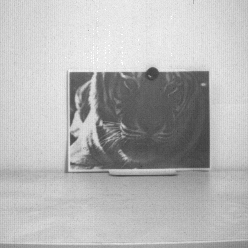}
			\vspace{-1em}\\
			\includegraphics[width=\linewidth, trim={0cm 0 0cm 0}]{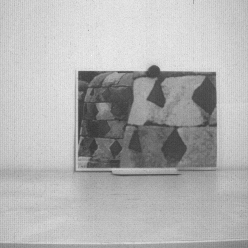}
			\vspace{-1.5em}
			\subcaption*{\scriptsize Reference}
		\end{subfigure}
		\begin{subfigure}[b]{0.18\linewidth}
			\includegraphics[width=\linewidth, trim={0cm 0 0cm 0}]{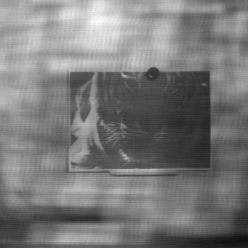}
			\vspace{-1em}\\
			\includegraphics[width=\linewidth, trim={0cm 0 0cm 0}]{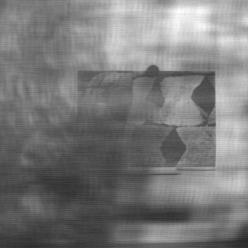}
			\vspace{-1.5em}
			\subcaption*{\scriptsize F-SAI (20)}
		\end{subfigure}
		\begin{subfigure}[b]{0.18\linewidth}
			\includegraphics[width=\linewidth, trim={0cm 0 0cm 0}]{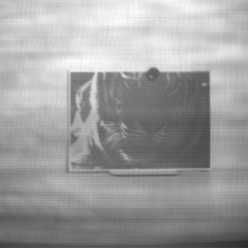}
			\vspace{-1em}\\
			\includegraphics[width=\linewidth, trim={0cm 0 0cm 0}]{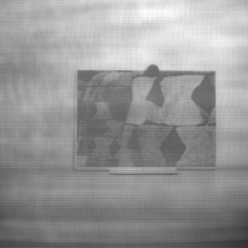}
			\vspace{-1.5em}
			\subcaption*{\scriptsize F-SAI (100)}
		\end{subfigure}
		\begin{subfigure}[b]{0.18\linewidth}
			\includegraphics[width=\linewidth, trim={0cm 0 0cm 0}]{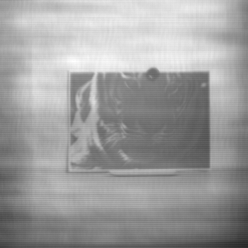}
			\vspace{-1em}\\
			\includegraphics[width=\linewidth, trim={0cm 0 0cm 0}]{fig2/figure-slow/FSAI-011-100.png}
			\vspace{-1.5em}
			\subcaption*{\scriptsize F-SAI (200)}
		\end{subfigure}
		\begin{subfigure}[b]{0.18\linewidth}
			\includegraphics[width=\linewidth, trim={0cm 0 0cm 0}]{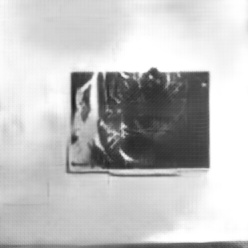}
			\vspace{-1em}\\
			\includegraphics[width=\linewidth, trim={0cm 0 0cm 0}]{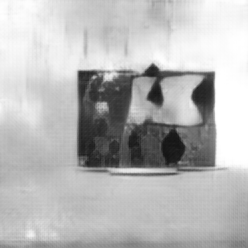}
			\vspace{-1.5em}
			\subcaption*{\scriptsize E-SAI+Hybrid}
		\end{subfigure}
		\caption{\rmfamily \fontsize{8pt}{0} Qualitative results}
		\label{fig:Slow-quali}
	\end{subfigure}
	\vspace{-.5em}
    \caption{
    {\color{\colored}
    Quantitative (a) and Qualitative (b) results of F-SAI+ACC under different numbers of observations, where F-SAI (20/100/200) represents the F-SAI+ACC result with 20/100/200 views and results of E-SAI are provided for comparison.
    }
    }
    \label{fig:Slow}
    \vspace{-1em}
\end{figure*}


\par 
\noindent{\bf Refocus-Net.} The Refocus-Net is fed with a $2N$-channel tensor composed of $2N$ event frames from the unfocused event stream and outputs the warping parameter $\boldsymbol{\psi}$. We first pre-train the E-SAI+Hybrid network with manually refocused event streams using ground truth warping parameters $\boldsymbol{\psi}$. After that, fixing the weights of the pre-trained E-SAI+Hybrid network and replacing the manual refocus module with the Refocus-Net, we train the Refocus-Net individually by backpropagating the image reconstruction loss.  
The Refocus-Net is finally trained for 200 epochs using the same training strategy (Adam optimizer, batch size 12 and SGDR schedule) and loss function as the hybrid network, except we set the initial learning rate to $3\times {10}^{-4}$.

\subsection{Comparisons between F-SAI and E-SAI}\label{sec:6-2}
In this subsection, we first evaluate the performance of SAI methods respectively over the indoor and outdoor datasets with dense occlusions. Comparisons of the proposed E-SAI+Hybrid networks with manual and auto refocusing methods to the state-of-the-art methods are made qualitatively and quantitatively. After that, the superiority of E-SAI to F-SAI under extreme lighting conditions is also validated.

{\color{\colored}
\subsubsection{Results with Dense Occlusions}\label{sec:dense_occ}
{\color{\seccolored}
The qualitative results on indoor and outdoor scenes are presented in Figs.~\ref{Indoor} and \ref{Outdoor}, respectively.
}
In the indoor experiments, we mainly test F-SAI and E-SAI methods with simple objects. As displayed in Fig.~\ref{Indoor}, the reconstruction results of F-SAI methods are severely disturbed by dense occlusions where a lot of details are missing or blurred. This is because the signal information for F-SAI, \ie, $I_{\theta}^{A}$ in Eq.~\eqref{FSAI}, is heavily contaminated by occlusions, and a lot of redundant information like the texture of occlusions is recorded in each input frames, as shown in Fig.~\ref{First}c. Thus, although F-SAI+CNN can achieve a better de-occlusion effect than F-SAI+ACC via deep learning-based methods, the results still suffer from detail losses and artifacts.
Compared with F-SAI, E-SAI methods can retain more details and produce results with better visual effects since the major signal events for E-SAI, \ie, $\mathcal{E}_{\theta}^{OA}$ in Eq.~\eqref{ESAI}, can be effectively triggered by dense occlusions. To reveal the advantages of our hybrid network, we compare the results of E-SAI with different reconstruction techniques in Fig.~\ref{DetailVis} and Tab.~\ref{tab}. It is difficult for E-SAI+ACC to produce satisfactory quantitative results since the emission of events is based on the log-scale brightness change, which differs from the intensity directly recorded in reference images. Through learning the mapping relationship between the event domain and the image domain, this problem can be well resolved by E-SAI+CNN and E-SAI+Hybrid. However, E-SAI+CNN fed directly with the stacked event frames cannot efficiently deal with the temporal information of asynchronous events, thus degrading the visual quality with detail losses, artifacts, and saturation. These issues can be well mitigated by the hybrid architecture where the SNN encoder utilizes temporal information. Furthermore, LIF neurons in SNN can efficiently leak out the influence of noise events which are either emitted randomly or scattered after the refocusing process. Consequently, the proposed hybrid network generates images with more uniform structure and realistic details.

\par 
For the outdoor dataset, we consider more general targets, including cars, fields, and buildings. Compared with the indoor scenes, outdoor lighting conditions are much more complicated, making it harder for F-SAI methods to generate sharp results, as shown in Fig.~\ref{Outdoor}.
Similarly, complex lighting conditions also degrade the performance of E-SAI due to the increased noise events, \eg, $\mathcal{E}^{OO}_\theta$. The rising number of noise events makes the target indistinguishable in the results of E-SAI+ACC and brings more disturbances to E-SAI+CNN, deteriorating the reconstruction quality with severe saturation problems. 
Thanks to the hybrid SNN-CNN architecture, noise events can be alleviated from the temporal dimension. In Tab.~\ref{tab}, our E-SAI+Hybrid method excels its pure CNN counterpart with a 4 dB increase in PSNR, a 24\% improvement in SSIM, and a 44\% decrease in LPIPS. This shows that the use of an SNN encoder achieves a better denoising effect and improves the overall reconstruction performance, which is consistent with the qualitative results shown in Figs.~\ref{Outdoor} and \ref{DetailVis}.

\begin{figure*}[t]
	\centering
	\begin{subfigure}[b]{0.133\linewidth}
		\includegraphics[width=\linewidth, trim={0cm 0 0cm 0}]{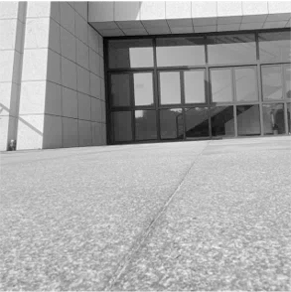}
		\vspace{-1em}\\
		\includegraphics[width=\linewidth, trim={0cm 0 0cm 0}]{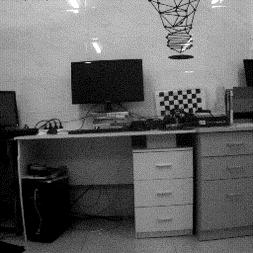}
		\vspace{-1.5em}
		\subcaption*{\scriptsize Reference}
	\end{subfigure}
	\begin{subfigure}[b]{0.133\linewidth}
		\includegraphics[width=\linewidth]{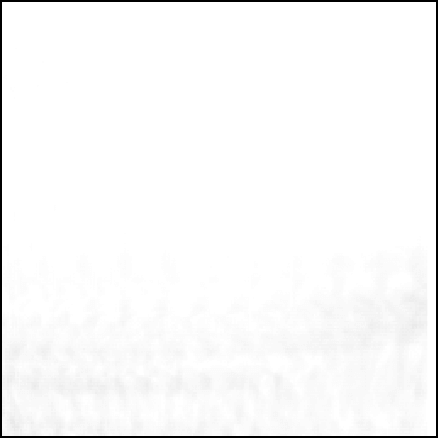}
		\vspace{-1em}\\
		\includegraphics[width=\linewidth, trim={0cm 0 0cm 0}]{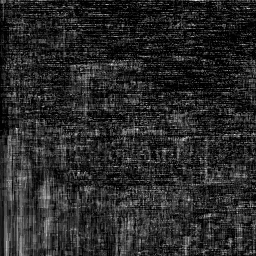}
		\vspace{-1.5em}
		\subcaption*{\scriptsize F-SAI+ACC}
	\end{subfigure}
	\begin{subfigure}[b]{0.133\linewidth}
		\includegraphics[width=\linewidth, trim={0cm 0 0cm 0}]{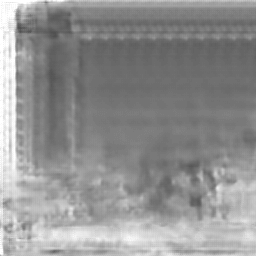}
		\vspace{-1em}\\
		\includegraphics[width=\linewidth, trim={0cm 0 0cm 0}]{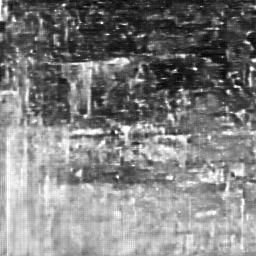}
		\vspace{-1.5em}
		\subcaption*{\scriptsize F-SAI+CNN}
	\end{subfigure}
	\begin{subfigure}[b]{0.133\linewidth}
		\includegraphics[width=\linewidth, trim={0cm 0 0cm 0}]{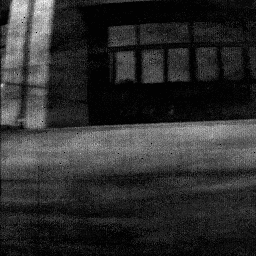}
		\vspace{-1em}\\
		\includegraphics[width=\linewidth, trim={0cm 0 0cm 0}]{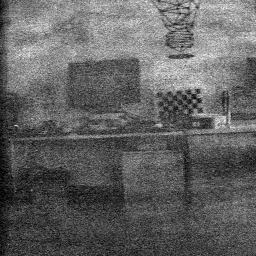}
		\vspace{-1.5em}
		\subcaption*{\scriptsize E-SAI+ACC}
	\end{subfigure}
	\begin{subfigure}[b]{0.133\linewidth}
		\includegraphics[width=\linewidth, trim={0cm 0 0cm 0}]{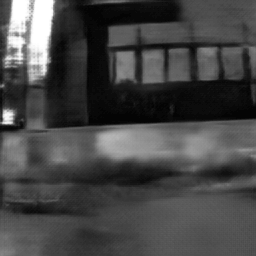}
		\vspace{-1em}\\
		\includegraphics[width=\linewidth, trim={0cm 0 0cm 0}]{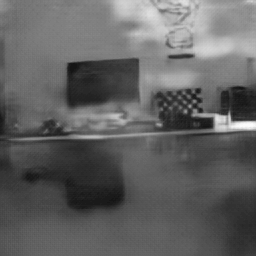}
		\vspace{-1.5em}
		\subcaption*{\scriptsize E-SAI+CNN}
	\end{subfigure}
	\begin{subfigure}[b]{0.133\linewidth}
		\includegraphics[width=\linewidth, trim={0cm 0 0cm 0}]{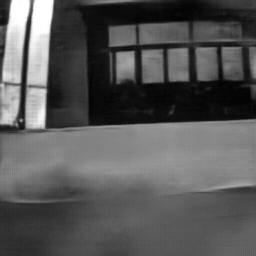}
		\vspace{-1em}\\
		\includegraphics[width=\linewidth, trim={0cm 0 0cm 0}]{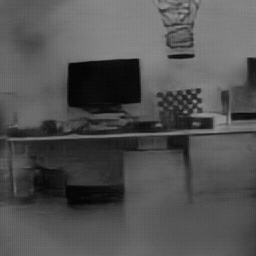}
		\vspace{-1.5em}
		\subcaption*{\scriptsize E-SAI+Hybrid (M)}
	\end{subfigure}
	\begin{subfigure}[b]{0.133\linewidth}
		\includegraphics[width=\linewidth, trim={0cm 0 0cm 0}]{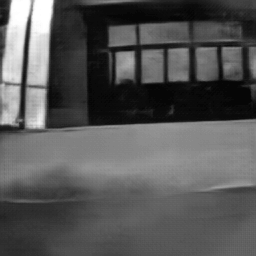}
		\vspace{-1em}\\
		\includegraphics[width=\linewidth, trim={0cm 0 0cm 0}]{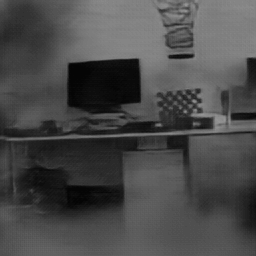}
		\vspace{-1.5em}
		\subcaption*{\scriptsize E-SAI+Hybrid (A)}
	\end{subfigure}
	\vspace{-.5em}
	\caption{Qualitative comparisons between F-SAI and E-SAI algorithms under extreme lighting conditions, where the reference images are captured under well lighting conditions. } 
	\label{ExtremeLight}
	\vspace{-.5em}
\end{figure*}
\begin{table*}[th]
\centering
\renewcommand{\arraystretch}{1.3}
\caption{
{\color{\seccolored}
Quantitative comparisons of F-SAI and E-SAI. Results are the average over all test sequences of the corresponding datasets. Our E-SAI+Hybrid outperforms the state-of-the-art SAI method (F-SAI+CNN) with a 3-5 dB improvement in peak signal to noise ratio (PSNR) and a 24\%-56\% decrease in perceptual distance (LPIPS) \cite{zhang2018perceptual} on our SAI dataset.
}
\vspace{-.5em}
}
\begin{tabular}{c|p{1.3cm}<{\centering}p{1.3cm}<{\centering}p{1.3cm}<{\centering}|p{1.3cm}<{\centering}p{1.3cm}<{\centering}p{1.3cm}<{\centering}}
\hline
\multirow{2}{*}{\textbf{Method}} & \multicolumn{3}{c|}{\textbf{Indoor}} & \multicolumn{3}{c}{\textbf{Outdoor}} \\ \cline{2-7} 
 & \multicolumn{1}{c|}{PSNR(dB) $\uparrow$} & \multicolumn{1}{c|}{SSIM $\uparrow$} & LPIPS $\downarrow$ & \multicolumn{1}{c|}{PSNR(dB) $\uparrow$} & \multicolumn{1}{c|}{SSIM $\uparrow$} & LPIPS $\downarrow$\\ \hline
F-SAI+ACC \cite{vaishUsingPlaneParallax2004} & 14.22 & 0.3484 & 0.2955 & 11.86 & 0.3925 & 0.2821 \\
F-SAI+CNN \cite{wangDeOccNetLearningSee2019} & 25.01 & 0.7933 & 0.1245 & \underline{17.39} & 0.5616 & 0.1347 \\ \hline
E-SAI+ACC & 14.96 & 0.2608 & 0.3219 & 10.26 & 0.2956 & 0.2875 \\
E-SAI+CNN & 26.53 & 0.7653 & 0.0871 & 15.98 & 0.5240 & 0.1838 \\
E-SAI+Hybrid (M) \cite{zhang2021event}& \textbf{30.71} & \textbf{0.8311} & \textbf{0.0374} & \textbf{20.39} & \textbf{0.7037} & \textbf{0.0981} \\ 
E-SAI+Hybrid (A) & \underline{29.55} & \underline{0.8086} & \underline{0.0546} & \textbf{20.39} & \underline{0.6961} & \underline{0.1013} \\ \hline
\end{tabular}
\label{tab}
\vspace{-1em}
\end{table*}
{\color{\colored}
To further verify the superiority of E-SAI over F-SAI, we investigate the performance of F-SAI+ACC with more angular sampling.
We collect up to 200 frames for 10 indoor scenes during camera movement. Fig.~\ref{fig:Slow} shows that the reconstruction performance of F-SAI+ACC becomes better when the input rises from 20 to 120 frames since more target information, \ie, $I^{A}_{\theta}$ in Eq.~\eqref{FSAI}, is acquired from multiple viewpoints.
With sufficient angular sampling, the effect of $\mathcal{M}^{O}$ is alleviated and the occluded scenes can be fully observed as shown in Fig.~\ref{fig:Slow-quali} with F-SAI (100). 
However, further increasing the number of observations, \eg, 120-200 views, hardly results in better reconstruction quality of F-SAI since the disturbance from dense occlusions, \ie, $I_{\theta}^{O}$, exists at each view and simultaneously increases with more input frames.
Therefore, despite more angular sampling brings more target information for reconstruction, the performance of F-SAI is limited by dense occlusions.
Compared with F-SAI, our E-SAI method produces visual images with better contrast and less noise, validating its superiority over F-SAI under dense occlusions.

}

}

\subsubsection{Results with Extreme Lighting Conditions}
Extreme lighting conditions often degrade the reconstruction quality of F-SAI methods or even lead to failure reconstruction, as shown in Fig.~\ref{ExtremeLight}. This is because the light from the occluded target cannot be correctly measured due to the over/under exposure problems encountered with frame-based cameras, as displayed in Fig.~\ref{First}c. By contrast, E-SAI methods do not suffer from over/under exposure problems thanks to the high dynamic range of event cameras. Therefore, the light information of occluded targets can be reliably captured by event cameras and effectively reconstructed via our E-SAI methods under extreme lighting conditions.

\par
We compare the performance of E-SAI methods in extreme lighting scenes. Although the refocusing process scatters out the noise events, their presence still degrades the image quality of E-SAI+ACC since all the events are directly accumulated in the results. For the learning-based E-SAI methods, the saturation issue becomes more severe in E-SAI+CNN because the distribution of the collected event field in extreme lighting scenes differs from that in standard lighting scenes. By exploiting the SNN encoder to utilize the spatio-temporal information inside events, this issue can be mitigated by E-SAI+Hybrid, thus generating results with more natural textures and better contrast.

\subsection{Analysis of Refocus Module}\label{sec:exp-refocus}
{\color{\colored} An in-depth analysis to the refocus module will be given in this subsection, including the influence of the refocusing accuracy to the reconstruction performance, the comparison of auto and manual refocus methods, and the effectiveness of the proposed Refocus-Net.
}
\subsubsection{Influence of Warping Parameter Error}\label{sec:influ_of_psi}
In this subsection, we investigate the robustness of E-SAI+Hybrid to the estimation error of the camera poses $R,\ T$ and the target depth $d$.
Since our data is mainly recorded under the fronto-parallel camera motions as depicted in Fig.~\ref{refocus}, we only need to consider the warping parameter $\boldsymbol{\psi}$ in the horizontal direction ($\psi$ represents the horizontal warping parameter hereinafter).
Denoting the ground truth $\psi$ as $\hat{\psi}$, we apply the warping projection Eq.~\eqref{trans-final1} to two random pairs of data respectively selected from indoor and outdoor datasets with the parameter ratio $\psi/\hat{\psi}$ varying from $0.4$ to $1.6$, and reconstruct the corresponding visual images with our hybrid network.
\par
\begin{figure}[t!]
	\centering
	\begin{subfigure}{0.2\textwidth}
		\includegraphics[width=\textwidth]{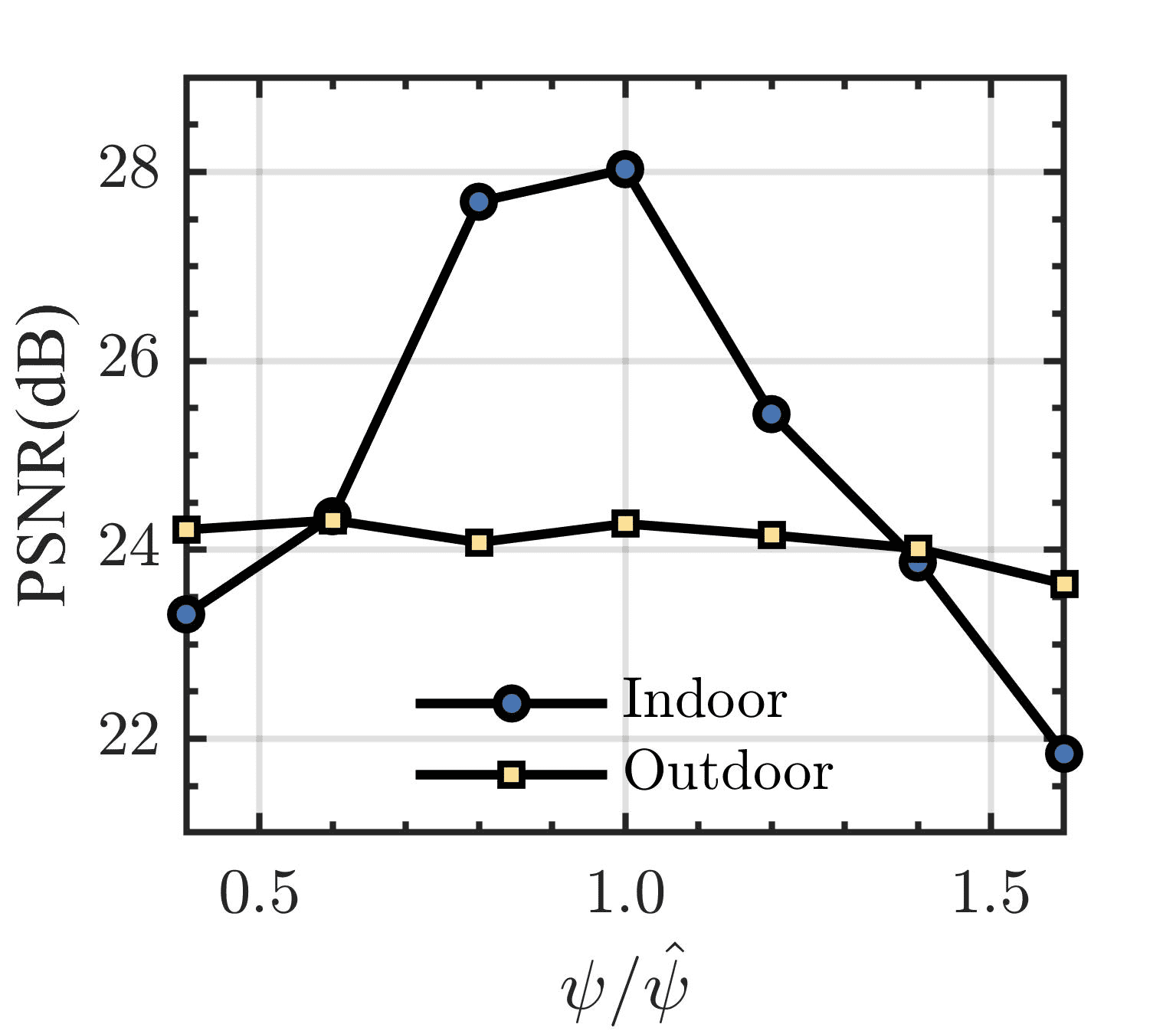}
		\vspace{-1.5em}
		\caption{\rmfamily \fontsize{8pt}{0} PSNR results}
		\label{speed-psnr}
	\end{subfigure}
	\begin{subfigure}{0.2\textwidth}
		\includegraphics[width=\textwidth]{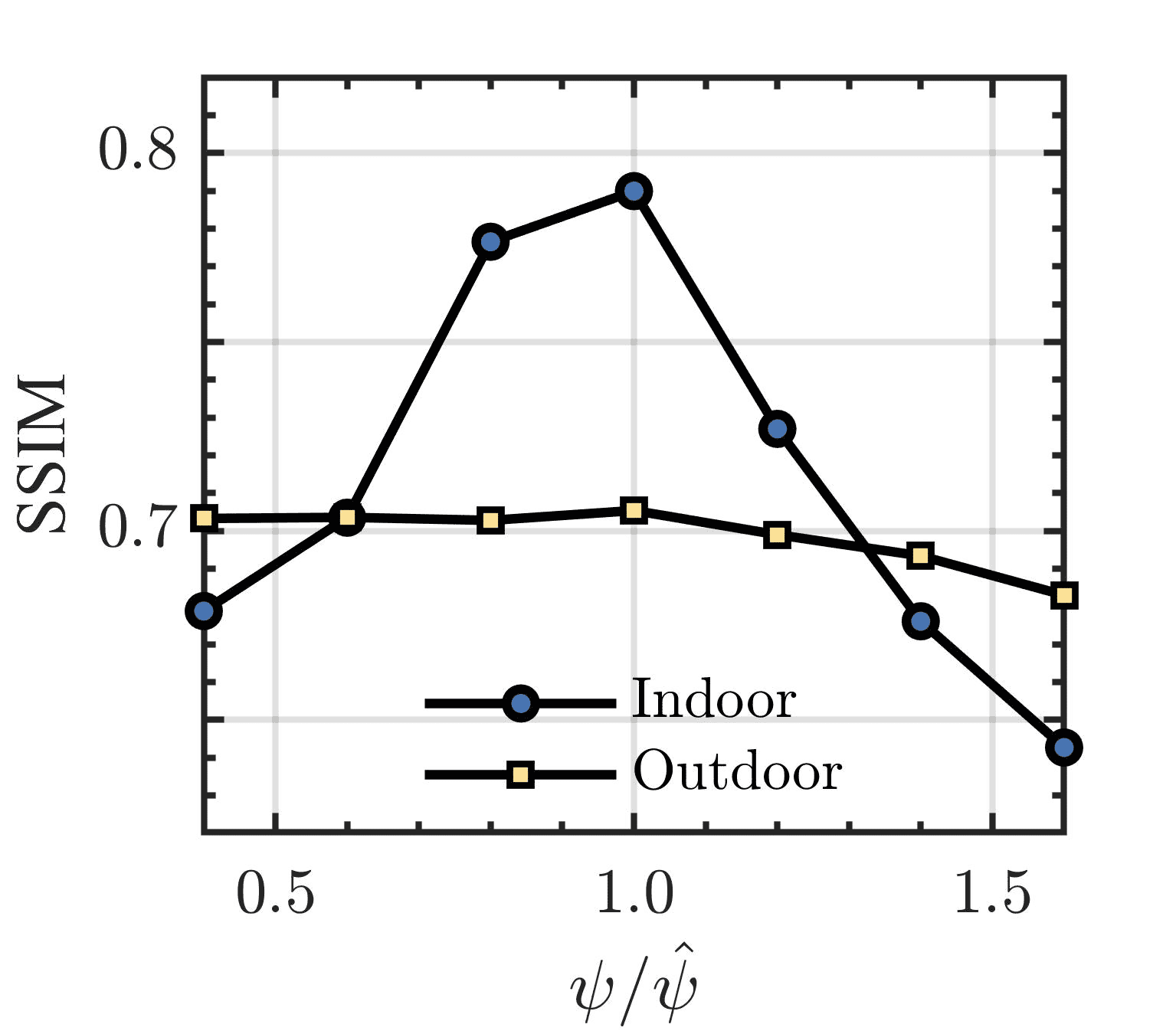}
		\vspace{-1.5em}
		\caption{\rmfamily \fontsize{8pt}{0} SSIM results}
		\label{speed-ssim}
	\end{subfigure}
	\\
	\begin{subfigure}[b]{\linewidth}
	\centering
	\begin{subfigure}[b]{0.22\linewidth}
		\includegraphics[width=1\linewidth]{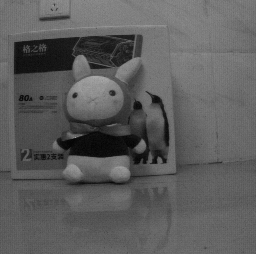}
	\end{subfigure}
	\begin{subfigure}[b]{0.22\linewidth}
		\includegraphics[width=1\linewidth]{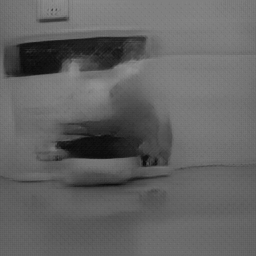}
	\end{subfigure}
	\begin{subfigure}[b]{0.22\linewidth}
		\includegraphics[width=1\linewidth]{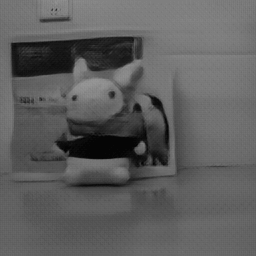}
	\end{subfigure}
	\begin{subfigure}[b]{0.22\linewidth}
		\includegraphics[width=1\linewidth]{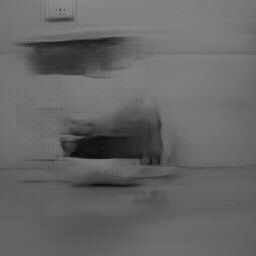}
	\end{subfigure}
	\\
	\vspace{.2em}
		\begin{subfigure}[b]{0.22\linewidth}
			\includegraphics[width=\linewidth, trim={0cm 0 0cm 0}]{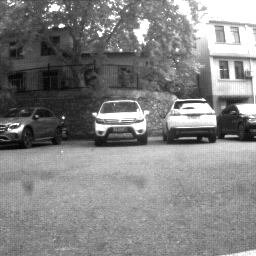}
			\vspace{-1.5em}
			\subcaption*{\centering \scriptsize Reference}
		\end{subfigure}
		\begin{subfigure}[b]{0.22\linewidth}
			\includegraphics[width=\linewidth, trim={0cm 0 0cm 0}]{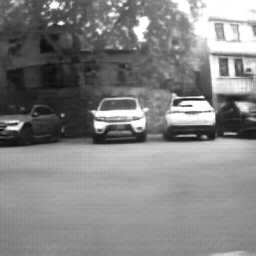}
			\vspace{-1.5em}
			\subcaption*{\centering \scriptsize $\psi/\hat{\psi}=0.4$}
		\end{subfigure}
		\begin{subfigure}[b]{0.22\linewidth}
			\includegraphics[width=\linewidth, trim={0cm 0 0cm 0}]{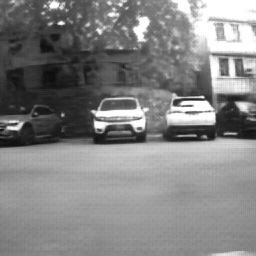}
			\vspace{-1.5em}
			\subcaption*{\centering \scriptsize $\psi/\hat{\psi}=1$}
		\end{subfigure}
		\begin{subfigure}[b]{0.22\linewidth}
			\includegraphics[width=\linewidth, trim={0cm 0 0cm 0}]{fig2/figure12/12-7.png}
			\vspace{-1.5em}
			\subcaption*{\centering \scriptsize $\psi/\hat{\psi}=1.6$}
		\end{subfigure}
		\caption{\rmfamily \fontsize{8pt}{0} Reconstruction results}
		\label{speedres}
	\end{subfigure}
	\vspace{-2em}
	\caption{Influence of the warping parameter $\psi$: (a-b) Quantitative and (c) qualitative results of indoor and outdoor scenes with different $\psi/\hat{\psi}$.} 
	\label{speed}
	\vspace{-1em}
\end{figure}

As illustrated in Figs.~\ref{speed-psnr} and \ref{speed-ssim}, the reconstruction quality of indoor data is severely degraded if the warping parameter is under/over estimated. Since targets of the indoor dataset are usually close to the camera (\ie, close-view targets), even a small error of the warping parameter will cause a significant pixel shift of events on the imaging plane. As a consequence, it makes the failure alignments of the signal events and thus leads to severe blurs and missing details in the final results as shown in Fig.~\ref{speedres}. On the other hand, E-SAI on the outdoor dataset mainly contains far-view targets. Therefore, the corresponding E-SAI performance is less sensitive to the estimation error of the warping parameter than that of the indoor dataset.

\begin{table}[t!]
\centering
\renewcommand{\arraystretch}{1.3}
\caption{
{\color{\seccolored}
APSE of Refocus-Net over all \emph{Indoor} and \emph{Outdoor} test sequences and the corresponding performance drops of E-SAI+Hybrid from (M) \cite{zhang2021event} to (A).
}
}
\vspace{-.5em}
\begin{tabular}{c|cccc}
\hline
 \multicolumn{1}{m{1.3cm}<{\centering}|}{\textbf{Scenes}} &  \multicolumn{1}{m{1.3cm}<{\centering}}{APSE}  & \multicolumn{1}{m{1.3cm}<{\centering}}{PSNR Drop} &  \multicolumn{1}{m{1.3cm}<{\centering}}{SSIM Drop} &  \multicolumn{1}{m{1.3cm}<{\centering}}{LPIPS Drop} \\ \hline
\multicolumn{1}{m{1.3cm}<{\centering}|}{\textbf{Indoor}} & \multicolumn{1}{m{1.3cm}<{\centering}}{\textcolor{\seccolored}{0.722}} & \multicolumn{1}{m{1.3cm}<{\centering}}{1.16} & \multicolumn{1}{m{1.3cm}<{\centering}}{0.0225} & \multicolumn{1}{m{1.3cm}<{\centering}}{0.0172} \\
\multicolumn{1}{m{1.3cm}<{\centering}|}{\textbf{Outdoor}} & \multicolumn{1}{p{1.3cm}<{\centering}}{\textcolor{\seccolored}{1.047}} & \multicolumn{1}{m{1.3cm}<{\centering}}{0} & \multicolumn{1}{m{1.3cm}<{\centering}}{0.0076} & \multicolumn{1}{m{1.3cm}<{\centering}}{0.0032} 
\\ \hline
\end{tabular}
\label{tab:refocusExp}
\vspace{-1.5em}
\end{table}
\subsubsection{Auto vs. Manual Refocus Module}
According to above discussions, the prior information of the camera velocity and target depth is essential for the refocus module, especially when encountering close-view target scenes. 
To investigate the performance of the proposed auto refocus method, 
we respectively employ Eq.~\eqref{trans-final} (manual refocus) and a pre-trained Refocus-Net (auto refocus) to generate the refocused events and apply the same hybrid network for reconstruction. 
Thus, we can evaluate the performance of Refocus-Net based on both reconstruction quality and event refocusing accuracy. 
{\color{\colored}
To quantitatively assess the refocusing accuracy, we introduce the average pixel shift error (APSE),
\[ 
\text{APSE} \triangleq \frac{1}{N_e} \sum_{i=1}^{N_e}\left\|(\boldsymbol{\psi}_i-\hat{\boldsymbol{\psi}}_i)\Delta t_{i}\right\|,
\]
where \textcolor{\seccolored}{$\Delta t_i=|t_i - t^{ref}|$}, $N_e$ denotes the total number of events input to the Refocus-Net, and $\boldsymbol{\psi}_i, \hat{\boldsymbol{\psi}}_i$ indicate the predicted and ground truth warping parameters, respectively. 
APSE measures the pixel alignment error caused by auto refocus method compared to the manual refocus one, and lower values indicate more accurate pixel alignment. Note we only compute the APSE of horizontal warping parameter here since our experiments only includes horizontal motion.
}
\par
\begin{figure}[!t] 
\centering
	\begin{subfigure}[b]{0.3\linewidth}
		\includegraphics[width=\linewidth, trim={0cm 0 0cm 0}]{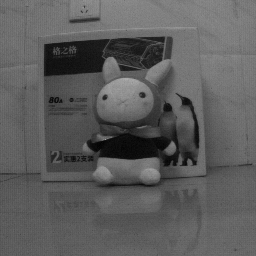}
		\vspace{-1em}\\
		\includegraphics[width=\linewidth, trim={0cm 0 0cm 0}]{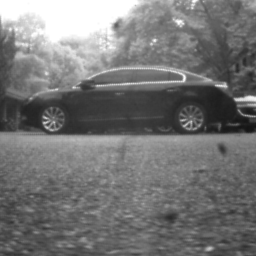}
		\vspace{-1.5em}
		\subcaption*{\scriptsize Reference}
	\end{subfigure}
	\begin{subfigure}[b]{0.3\linewidth}
		\includegraphics[width=\linewidth, trim={0cm 0 0cm 0}]{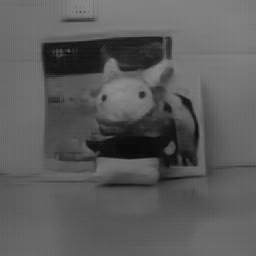}
		\vspace{-1em}\\
		\includegraphics[width=\linewidth, trim={0cm 0 0cm 0}]{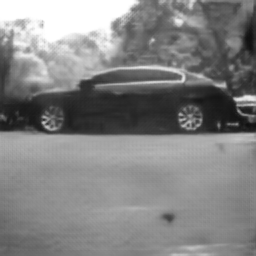}
		\vspace{-1.5em}
		\subcaption*{\scriptsize  E-SAI+Hybrid (M)}
	\end{subfigure}
	\begin{subfigure}[b]{0.3\linewidth}
		\begin{tikzpicture}[inner sep=0]
			\node {\includegraphics[width=\linewidth, trim={0cm 0 0cm 0}]{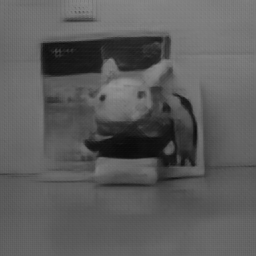}};
			\node [anchor=south,fill=black!10!white,opacity=0.75,inner sep=2] at (0,-1.3){\textcolor{red}{\scriptsize APSE: 0.790}};
		\end{tikzpicture}
		\vspace{-1em}\\
		\begin{tikzpicture}[inner sep=0]
			\node {\includegraphics[width=\linewidth, trim={0cm 0 0cm 0}]{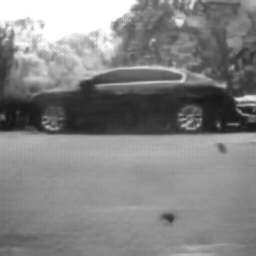}};
			\node [anchor=south,fill=black!10!white,opacity=0.75,inner sep=2] at (0,-1.3) {\textcolor{red}{\scriptsize APSE: 1.467}};
		\end{tikzpicture}
		\vspace{-1.5em}
		\subcaption*{\scriptsize E-SAI+Hybrid (A)}
	\end{subfigure}
		\vspace{-0.5em}
	\caption{Qualitative results generated by the E-SAI+Hybrid with manual refocus (M) and auto refocus (A). } 
	\label{refocusExp}
	\vspace{-1.5em}
\end{figure}

 
In Tab. \ref{tab:refocusExp}, applying the auto refocus method results in an average APSE of 0.7-1 pixels, which is generally acceptable from the qualitative perspective as depicted in Fig.~\ref{refocusExp}. 
Although the APSE of the outdoor dataset is relatively high, the corresponding performance degradation is 
still tolerable since outdoor data is less sensitive to the error of $\psi$ (strongly related to APSE) as discussed in Sec.~\ref{sec:influ_of_psi}.
Thus, under fronto-parallel camera motions shown in Fig.~\ref{refocus}, our auto refocus method can largely relax the dependence on camera motions and target depth with only slight performance degradation and thus facilitate practical usage.

\begin{figure}[t!]
	\centering
	\begin{subfigure}{0.32\linewidth}
		\includegraphics[width=\textwidth]{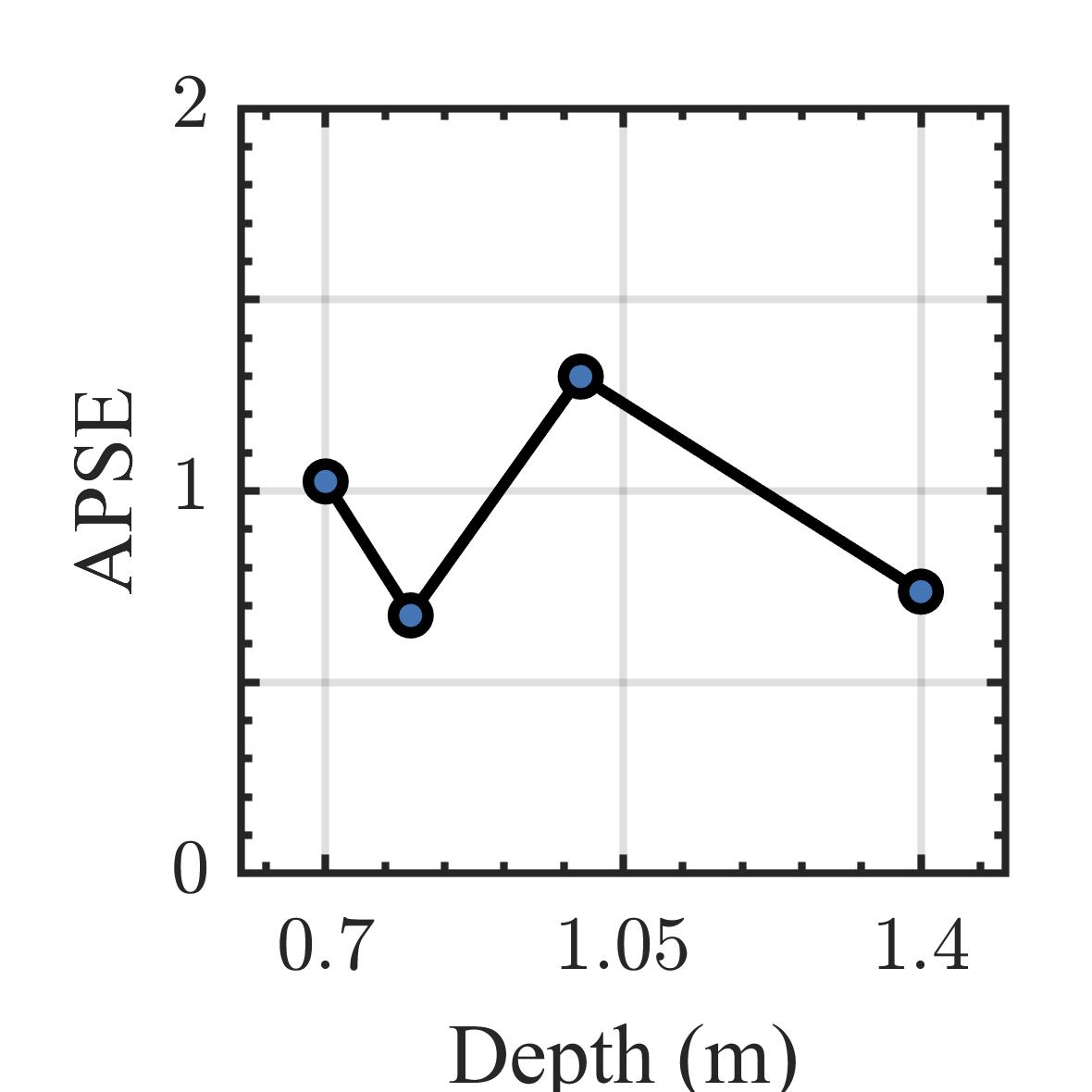}
		\caption{\rmfamily \fontsize{8pt}{0} APSE results}
		\label{DiffDepth_MPSE}
	\end{subfigure}
	\begin{subfigure}{0.32\linewidth}
		\includegraphics[width=\textwidth]{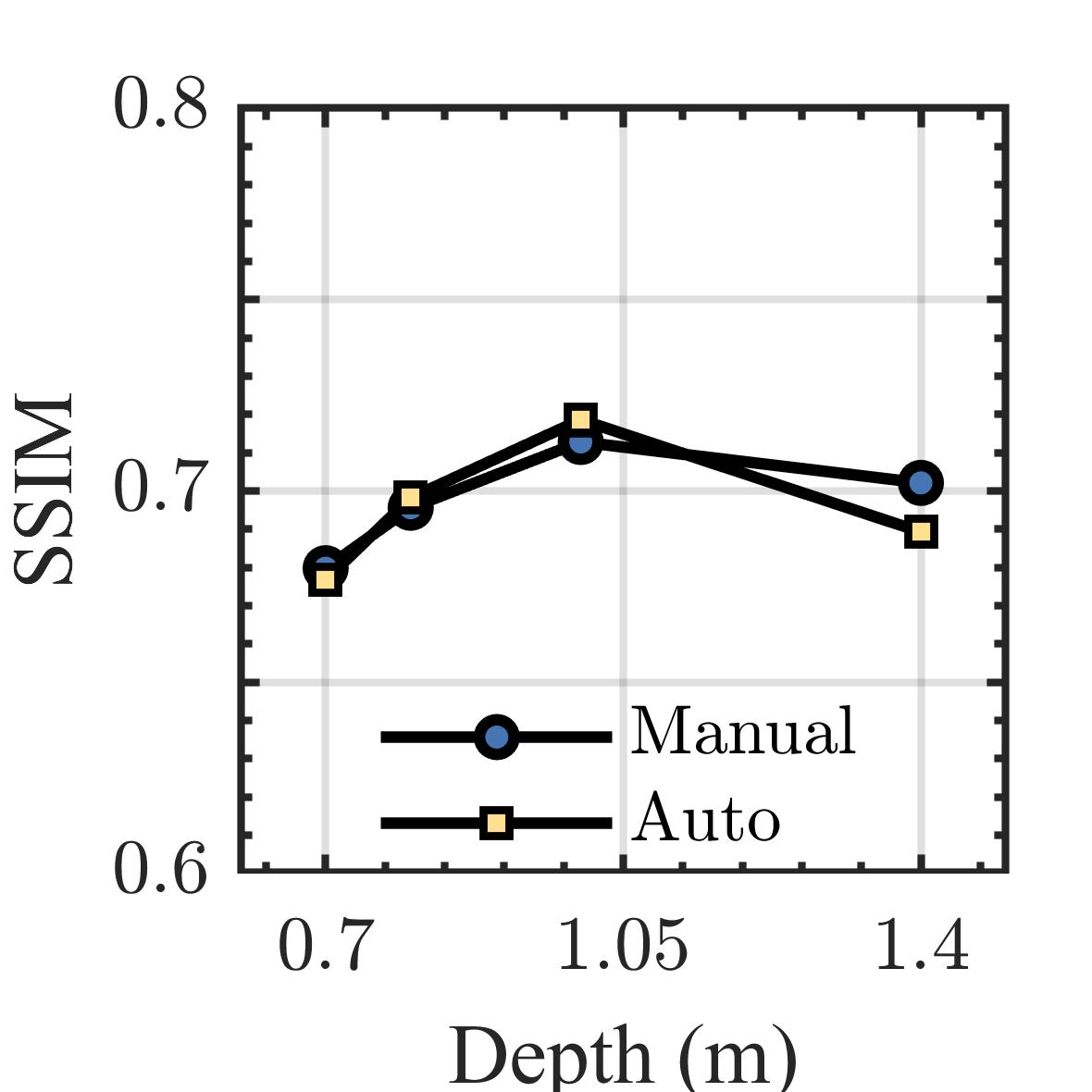}
		\caption{\rmfamily \fontsize{8pt}{0} SSIM results}
		\label{DiffDepth_SSIM}
	\end{subfigure}
	\begin{subfigure}{0.32\linewidth}
		\includegraphics[width=\textwidth]{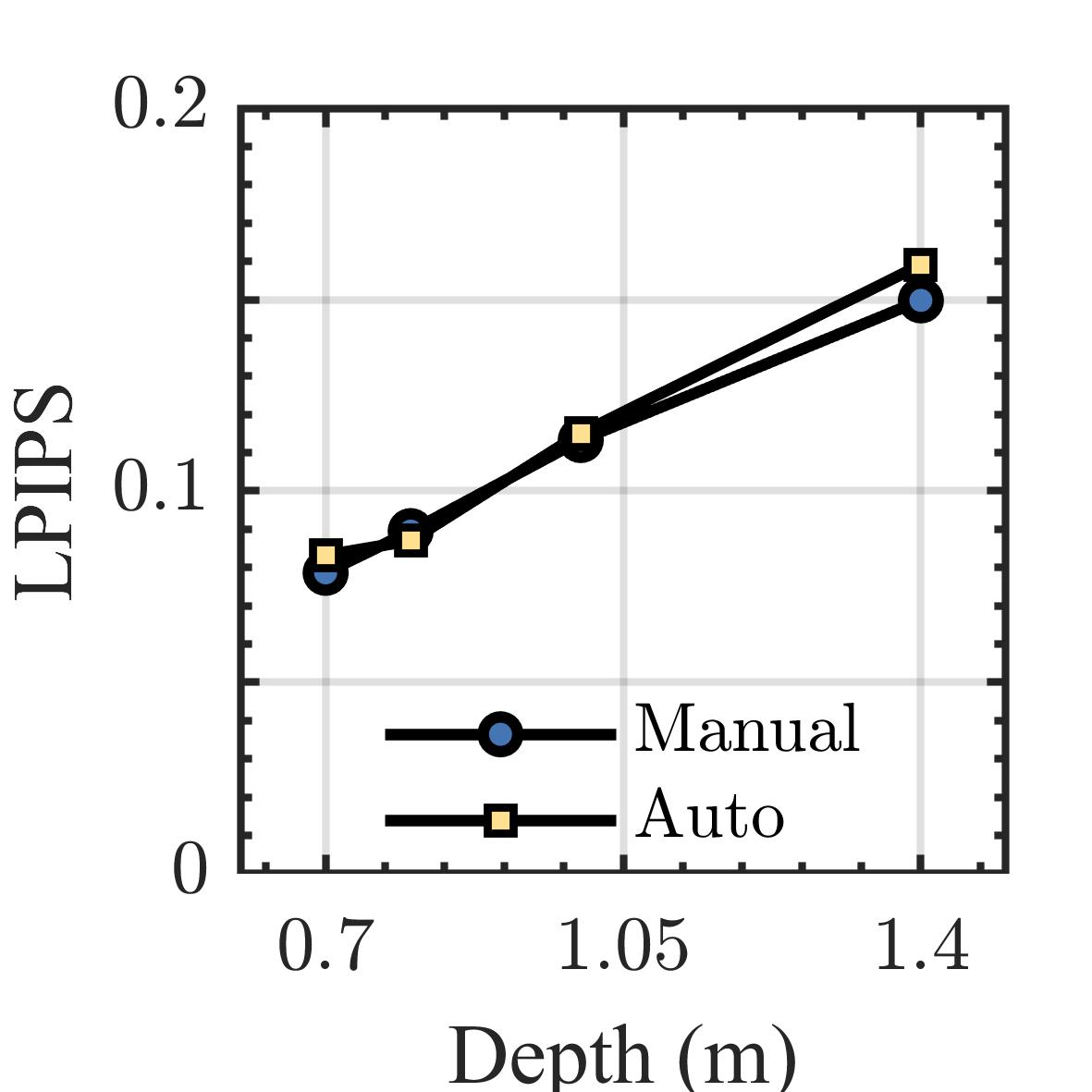}
		\caption{\rmfamily \fontsize{8pt}{0} LPIPS results}
		\label{DiffDepth_LPIPS}
	\end{subfigure}
	\vspace{-.5em}
    \caption{Quantitative comparisons under different target depths: (a) Refocus error (APSE) of Refocs-Net; (b-c) Reconstruction performance of E-SAI+Hybrid (A/M). }
    \label{DiffDepthExp}
    \vspace{-1.5em}
\end{figure}

\begin{figure*}[t!]
	\centering
	\begin{subfigure}{0.306\textwidth}
		\includegraphics[width=.87\textwidth]{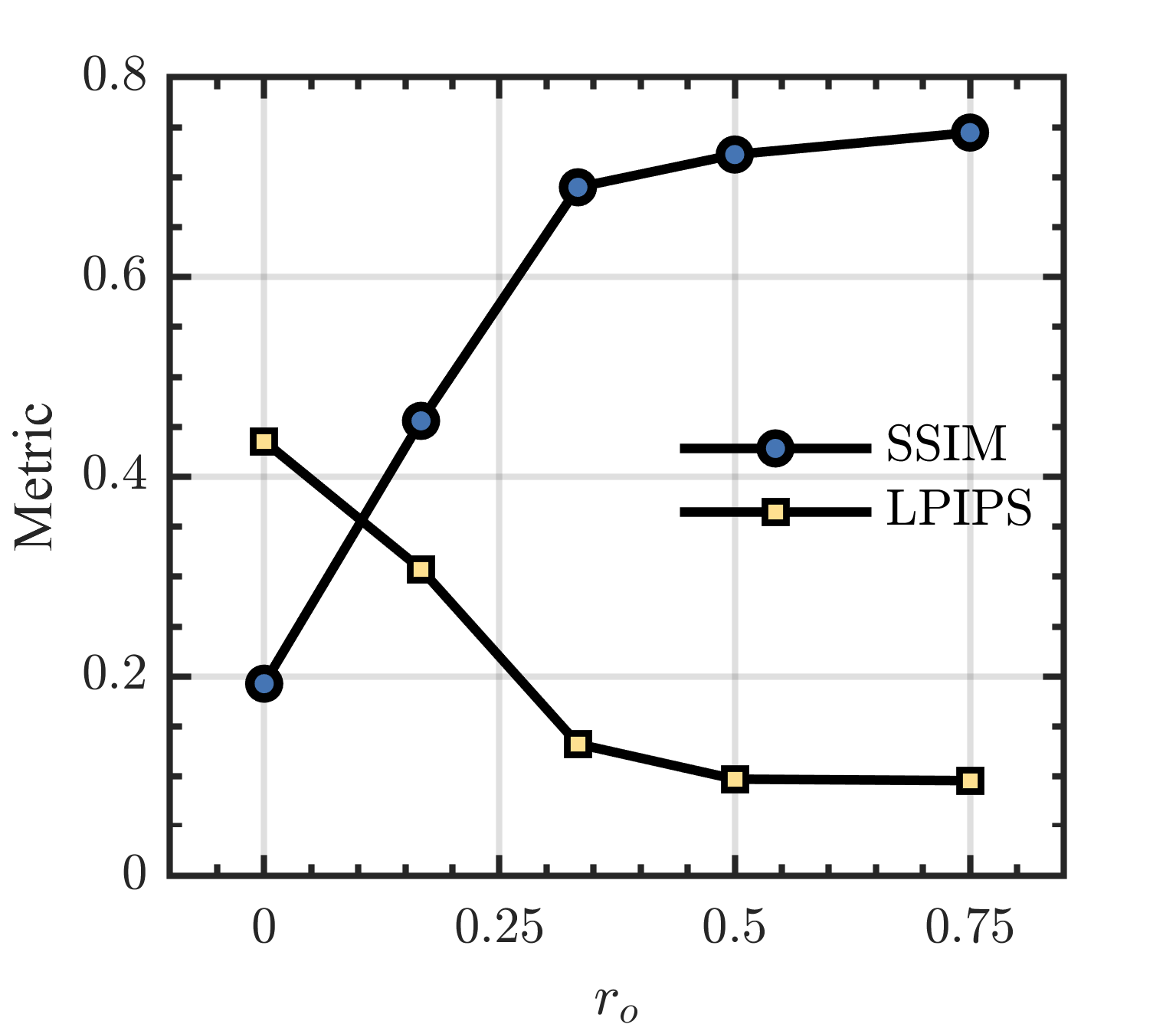}
		\vspace{-0.5em}
		\caption{\rmfamily \fontsize{8pt}{0} Quantitative results}
		\label{DiffOcc}
	\end{subfigure}	
	\begin{subfigure}{0.630\textwidth}
		\begin{subfigure}[b]{0.15\textwidth}
			\includegraphics[width=\textwidth, trim={0cm 0 0cm 0}]{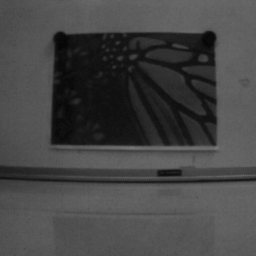}
			\vspace{-1em}\\
			\includegraphics[width=\textwidth, trim={0cm 0 0cm 0}]{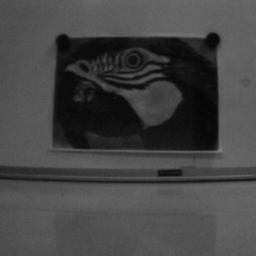}
			\vspace{-1.2em}
			\subcaption*{\scriptsize Reference}
		\end{subfigure}
		\begin{subfigure}[b]{0.15\textwidth}
			\includegraphics[width=\textwidth, trim={0cm 0 0cm 0}]{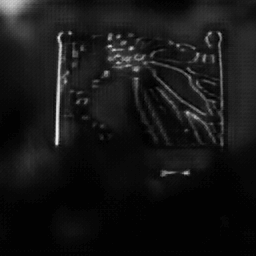}
			\vspace{-1em}\\
			\includegraphics[width=\textwidth, trim={0cm 0 0cm 0}]{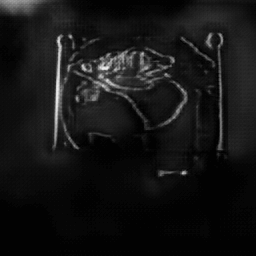}
			\vspace{-1.2em}
			\subcaption*{\scriptsize $r_o =0$}
		\end{subfigure}
		\begin{subfigure}[b]{0.15\textwidth}
			\includegraphics[width=\textwidth, trim={0cm 0 0cm 0}]{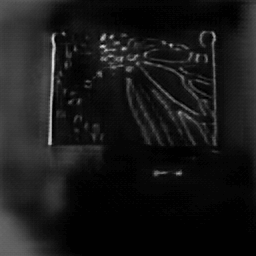}
			\vspace{-1em}\\
			\includegraphics[width=\textwidth, trim={0cm 0 0cm 0}]{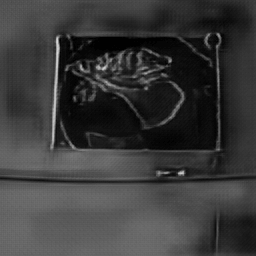}
			\vspace{-1.2em}
			\subcaption*{\scriptsize $r_o =1/6$}
		\end{subfigure}
		\begin{subfigure}[b]{0.15\textwidth}
			\includegraphics[width=\textwidth, trim={0cm 0 0cm 0}]{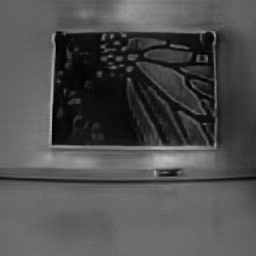}
			\vspace{-1em}\\
			\includegraphics[width=\textwidth, trim={0cm 0 0cm 0}]{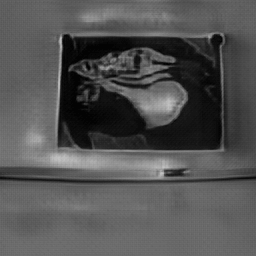}
			\vspace{-1.2em}
			\subcaption*{\scriptsize $r_o =1/3$}
		\end{subfigure}
		\begin{subfigure}[b]{0.15\textwidth}
			\includegraphics[width=\textwidth, trim={0cm 0 0cm 0}]{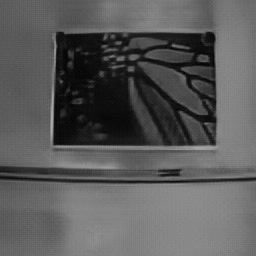}
			\vspace{-1em}\\
			\includegraphics[width=\textwidth, trim={0cm 0 0cm 0}]{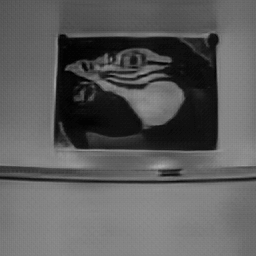}
			\vspace{-1.2em}
			\subcaption*{\scriptsize $r_o =1/2$}
		\end{subfigure}
		\begin{subfigure}[b]{0.15\textwidth}
			\includegraphics[width=\textwidth, trim={0cm 0 0cm 0}]{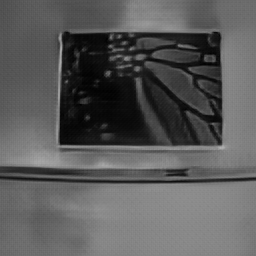}
			\vspace{-1em}\\
			\includegraphics[width=\textwidth, trim={0cm 0 0cm 0}]{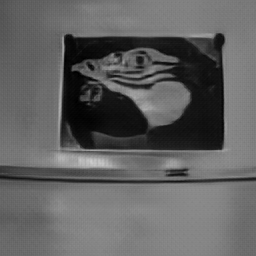}
			\vspace{-1.2em}
			\subcaption*{\scriptsize $r_o =3/4$}
		\end{subfigure}
		\caption{\rmfamily \fontsize{8pt}{0} Qualitative results}
		\label{DiffOccPic}
	\end{subfigure}
	\vspace{-.5em}
    \caption{
    {\color{\colored}
    Performance of E-SAI+Hybrid under different occlusion densities ($r_o$ indicates the ratio of the occluded area).}
    }
    \vspace{-1em}
    \label{DiffOccExp}
\end{figure*}

\subsubsection{Effectiveness of the Refocus-Net}
We further validate the effectiveness of the Refocus-Net on the same targets at different depths $d$ varying from 0.7 to 1.4 meters. It can also be regarded as testing Refocus-Net with different camera speeds or camera intrinsic parameters since they are coupled to the warping parameter $\psi$. The proposed auto refocus method, \ie, Refocus-Net, can successfully estimate the warping parameter $\psi$ and achieve promising performance under different target depths, as shown in Fig.~\ref{DiffDepth_MPSE}. Thus E-SAI+Hybrid method with Refocus-Net can achieve comparable performance to that with the manual refocus approach, as shown in Fig.~\ref{DiffDepth_SSIM} and \ref{DiffDepth_LPIPS}. And the Refocus-Net can even boost the performance of E-SAI+Hybrid and achieve better results than the manual refocusing method. This is because the spatial matching method employed in our dataset (detailed in Sec.~\ref{chapter-5}) is based on the SSIM value calculated from the noisy E-SAI+ACC results and the corresponding occlusion-free APS frames, and thus may occasionally cause a mismatch. By contrast, the Refocus-Net is supervised directly by the image reconstruction error, as discussed in Sec.~\ref{sec:train_rn}.

\subsection{Analysis of Reconstruction Module}\label{sec:exp-recon}
{\color{\colored}The effectiveness of the reconstruction module is evaluated in this subsection, including the influences of occlusion density and occlusion texture, the ablation study on the proposed E-SAI+Hybrid, and the analysis of hyper-parameters.}
\subsubsection{Influence of Occlusion Density}
Based on the results in Sec.~\ref{sec:dense_occ}, the proposed E-SAI+Hybrid method achieves promising results on our SAI dataset with dense occlusions, but how well does it perform under more general or sparse occlusions?
To investigate this point, we replace the wooden fence with cuttable cardboards as the occlusion for quantitative evaluation and separately analyze the performance of the reconstruction module under the different occlusion densities.
To facilitate the quantification of occlusion density, we introduce the metric $r_o \triangleq A_o / A_m$ where $A_o,\  A_m$ respectively denote the area of occlusions and the total observation area during camera moving. Then we conduct experiments with $r_o$ ranging from 0 (occlusion-free) to 3/4.

\par 
{\color{\colored}
As shown in Fig.~\ref{DiffOccExp}, the proposed E-SAI+Hybrid performs better in both qualitative and quantitative perspectives when the occlusion density increases. 
In occlusion-free scenes, the collected events are mainly caused by edges of target scenes with high-contrast regions, \ie, $\mathcal{E}_{\theta}^{AA}$ in Eq.~\eqref{ESAI}, which will be aligned on target edges after the refocusing process, leading to edge-like reconstructions as shown in Fig.~\ref{DiffOccPic} with $r_o=0$. Thus for the occlusion-free scenes, one can directly reconstruct the visual image using event to video translation methods, \eg, E2VID \cite{rebecq2019high}.
When occlusion ratio increases, more signal events can be collected, \ie, $\mathcal{E}_{\theta}^{OA}$, which encodes the overall scene texture of targets since it responds to both low and high contrast target regions by exploiting foreground occlusions as brightness reference. 
As depicted in Fig.~\ref{DiffOccExp}, the overall texture of target scene can be restored when $r_o\geq 1/3$, and the performance of E-SAI becomes better as the occlusion ratio increases.
}

\begin{figure}[t!]
	\centering
		\begin{subfigure}[b]{0.225\linewidth}
			\includegraphics[width=\linewidth, trim={0cm 0 0cm 0}]{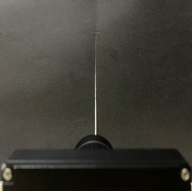}
			\vspace{-1.75em}
			\subcaption*{\centering \scriptsize Single-tiny-slit \hspace{\textwidth} occlusion}
		\end{subfigure}
		\begin{subfigure}[b]{0.225\linewidth}
			\includegraphics[width=\linewidth, trim={0cm 0 0cm 0}]{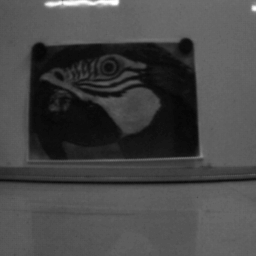}
			\vspace{-1.75em}
			\subcaption*{\centering \scriptsize Reference \hspace{\textwidth} SSIM / LPIPS}
		\end{subfigure}
		\begin{subfigure}[b]{0.225\linewidth}
			\includegraphics[width=\linewidth, trim={0cm 0 0cm 0}]{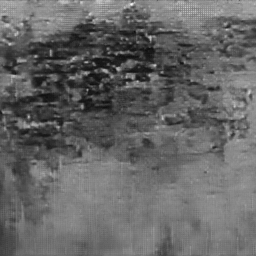}
			\vspace{-1.75em}
			\subcaption*{\centering \scriptsize F-SAI+CNN \hspace{\textwidth} 0.2562 / 0.2859}
		\end{subfigure}
		\begin{subfigure}[b]{0.225\linewidth}
			\includegraphics[width=\linewidth, trim={0cm 0 0cm 0}]{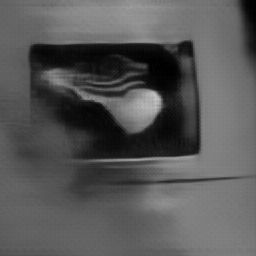}
			\vspace{-1.75em}
			\subcaption*{\centering \scriptsize \bf E-SAI+Hybrid \hspace{\textwidth} 0.6468 / 0.1575}
		\end{subfigure}\\
		\begin{subfigure}[b]{0.225\linewidth}
			\includegraphics[width=\linewidth, trim={0cm 0 0cm 0}]{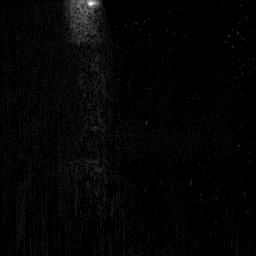}
			\vspace{-1.75em}
			\subcaption*{\centering \scriptsize Occluded \hspace{\textwidth}  view}
		\end{subfigure}
		\begin{subfigure}[b]{0.225\linewidth}
			\includegraphics[width=\linewidth, trim={0cm 0 0cm 0}]{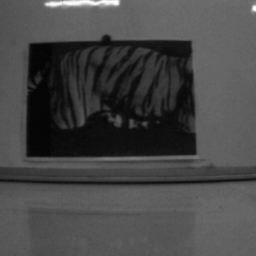}
			\vspace{-1.75em}
			\subcaption*{\centering \scriptsize Reference \hspace{\textwidth} SSIM / LPIPS}
		\end{subfigure}
		\begin{subfigure}[b]{0.225\linewidth}
			\includegraphics[width=\linewidth, trim={0cm 0 0cm 0}]{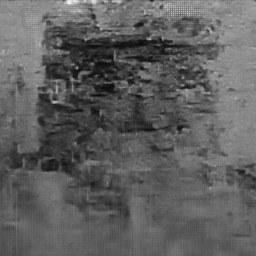}
			\vspace{-1.75em}
			\subcaption*{\centering \scriptsize F-SAI+CNN \hspace{\textwidth} 0.2918 / 0.2812}
		\end{subfigure}
		\begin{subfigure}[b]{0.225\linewidth}
			\includegraphics[width=\linewidth, trim={0cm 0 0cm 0}]{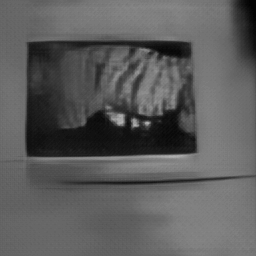}
			\vspace{-1.75em}
			\subcaption*{\centering \scriptsize \bf E-SAI+Hybrid \hspace{\textwidth} 0.6830 / 0.1329}
		\end{subfigure}
		\vspace{-.5em}
	\caption{Comparisons between the state-of-the-art SAI method (F-SAI+CNN) and the proposed E-SAI+Hybrid to see through a single-tiny-slit.} 
	\label{SingleGap}
    \vspace{-1.5em}
\end{figure}

We also considered an extremely occluded scene where the camera can only observe the target scene through a single tiny slit on the cardboard, as shown in Fig.~\ref{SingleGap}. Compared to the state-of-the-art F-SAI+CNN method, the proposed E-SAI+Hybrid gains a 142\% increase in SSIM and a 48\% decrease in LPIPS. Qualitatively, F-SAI+CNN fails to reconstruct the occluded scenes while E-SAI+Hybrid can produce visually acceptable results with clear shapes and natural textures. 
Thanks to the low temporal latency, the event camera can ``scan'' the target scene through the tiny slit, acquiring enough information to guarantee the performance of E-SAI+Hybrid.

\begin{figure}[t!]
	\centering
	\begin{subfigure}{1\linewidth}
	\centering
	    \begin{subfigure}[b]{0.3\linewidth}
			\includegraphics[width=\linewidth, trim={0cm 0 0cm 0}]{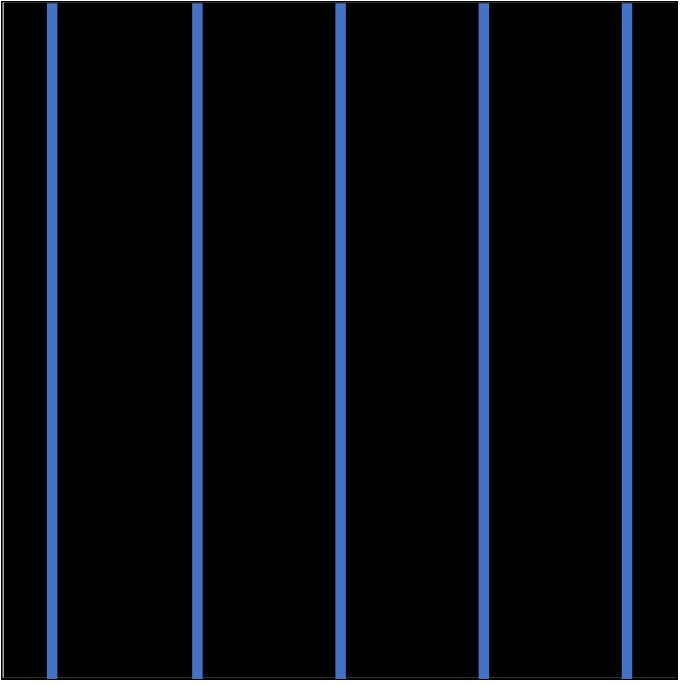}
			\vspace{-1.5em}
			\subcaption*{\scriptsize Black cardboard}
		\end{subfigure}
		\begin{subfigure}[b]{0.3\linewidth}
			\includegraphics[width=\linewidth, trim={0cm 0 0cm 0}]{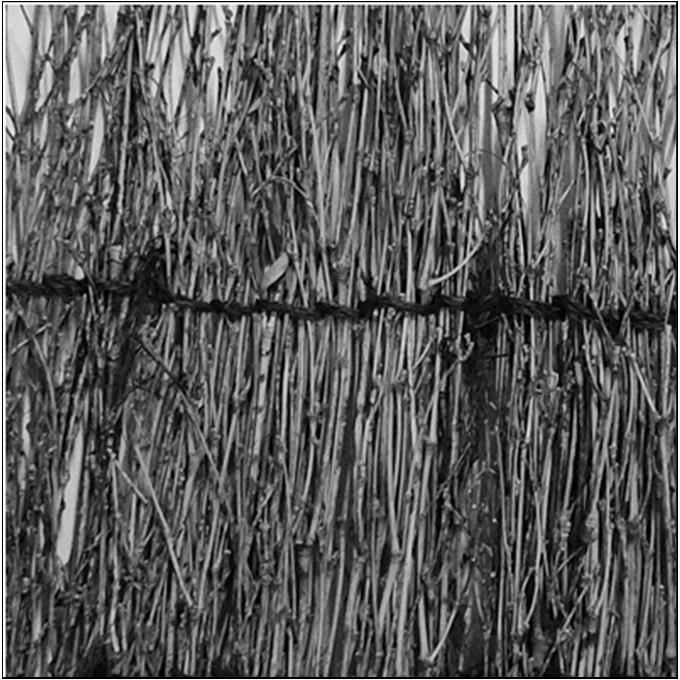}
			\vspace{-1.5em}
			\subcaption*{\scriptsize Wooden fence}
		\end{subfigure}
		\begin{subfigure}[b]{0.3\linewidth}
			\includegraphics[width=\linewidth, trim={0cm 0 0cm 0}]{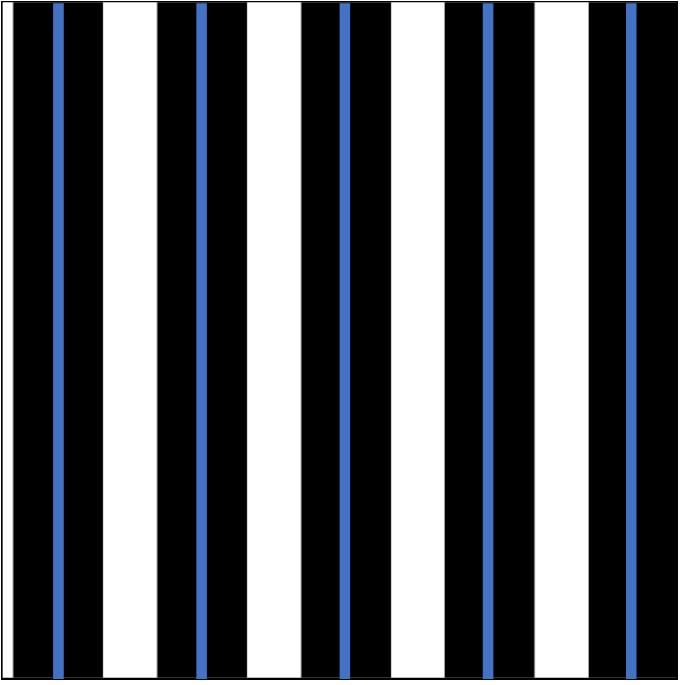}
			\vspace{-1.5em}
			\subcaption*{\scriptsize Vertical stripes}
		\end{subfigure}
		\vspace{-.5em}
		\caption{\rmfamily \fontsize{8pt}{0} Illustration of foreground occlusions}
		\label{fig:TexturedOcc-Occ}
	\end{subfigure}	
	\begin{subfigure}{1\linewidth}
	\centering
		\begin{subfigure}[b]{0.225\linewidth}
			\includegraphics[width=\linewidth, trim={0cm 0 0cm 0}]{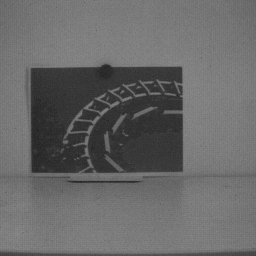}
			\vspace{-1em}\\
			\includegraphics[width=\linewidth, trim={0cm 0 0cm 0}]{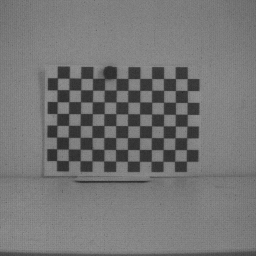}
			\vspace{-1.5em}
			\subcaption*{\scriptsize Reference}
		\end{subfigure}
		\begin{subfigure}[b]{0.225\linewidth}
			\includegraphics[width=\linewidth, trim={0cm 0 0cm 0}]{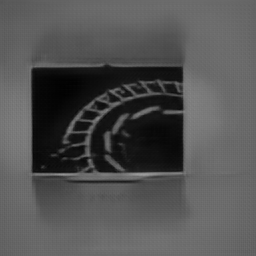}
			\vspace{-1em}\\
			\includegraphics[width=\linewidth, trim={0cm 0 0cm 0}]{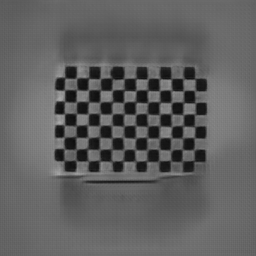}
			\vspace{-1.5em}
			\subcaption*{\scriptsize Black cardboard}
		\end{subfigure}
		\begin{subfigure}[b]{0.225\linewidth}
			\includegraphics[width=\linewidth, trim={0cm 0 0cm 0}]{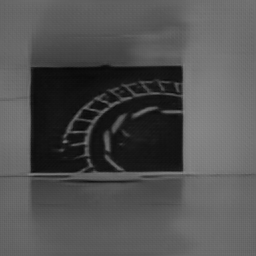}
			\vspace{-1em}\\
			\includegraphics[width=\linewidth, trim={0cm 0 0cm 0}]{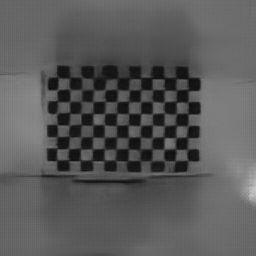}
			\vspace{-1.5em}
			\subcaption*{\scriptsize Wooden fence}
		\end{subfigure}
		\begin{subfigure}[b]{0.225\linewidth}
			\includegraphics[width=\linewidth, trim={0cm 0 0cm 0}]{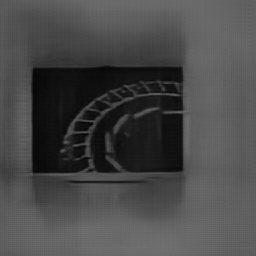}
			\vspace{-1em}\\
			\includegraphics[width=\linewidth, trim={0cm 0 0cm 0}]{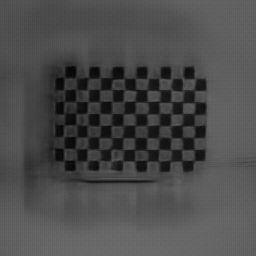}
			\vspace{-1.5em}
			\subcaption*{\scriptsize Vertical stripes}
		\end{subfigure}
		\vspace{-.5em}
		\caption{\rmfamily \fontsize{8pt}{0} Qualitative results under different occlusions}
		\label{fig:TexturedOcc-Res}
	\end{subfigure}
	
	\begin{subfigure}{1\linewidth}
	\begin{subfigure}{.48\linewidth}
	\centering
	    \includegraphics[width=1\linewidth, trim={0cm 0 0cm 0}]{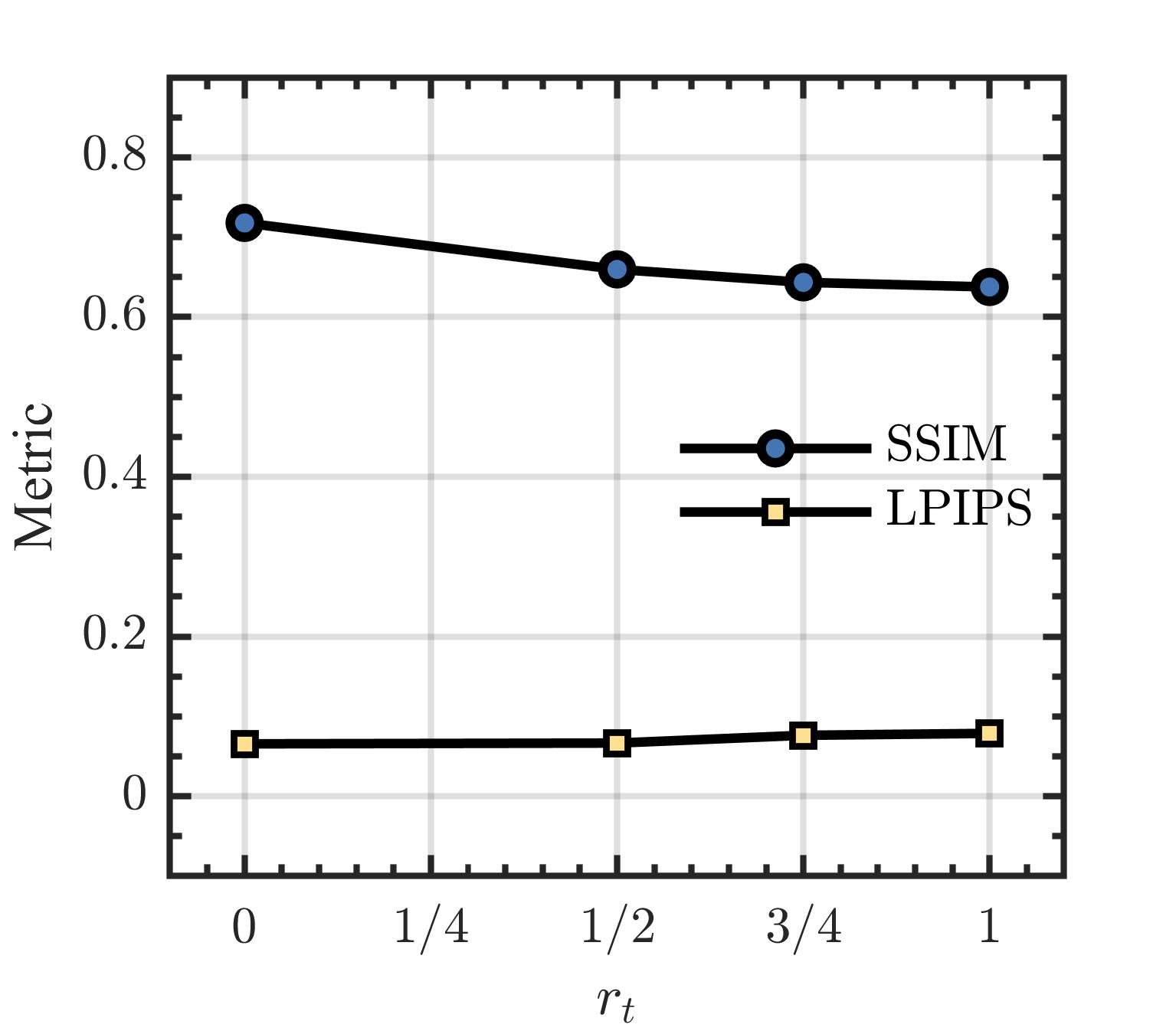}
	\end{subfigure}
	\begin{subfigure}{.5\linewidth}
	\centering
	\begin{subfigure}[b]{0.42\linewidth}
			\includegraphics[width=\linewidth, trim={0cm 0 0cm 0}]{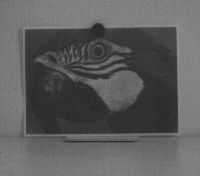}
			\vspace{-1.5em}
			\subcaption*{\scriptsize Reference}
			\includegraphics[width=\linewidth, trim={0cm 0 0cm 0}]{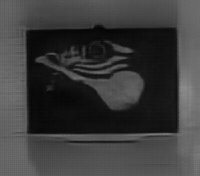}
			\vspace{-1.5em}
			\subcaption*{\scriptsize $r_t=1/2$}
		\end{subfigure}
		\begin{subfigure}[b]{0.42\linewidth}
			\includegraphics[width=\linewidth, trim={0cm 0 0cm 0}]{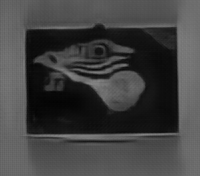}
			\vspace{-1.5em}
			\subcaption*{\scriptsize $r_t=0$}
			\includegraphics[width=\linewidth, trim={0cm 0 0cm 0}]{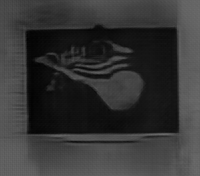}
			\vspace{-1.5em}
			\subcaption*{\scriptsize $r_t=1$}
		\end{subfigure}
	\end{subfigure}
	\vspace{-.5em}
	\caption{Results under different amount of occlusion textures}
	\label{fig:TexturedOccInflu}
	\end{subfigure}
    \vspace{-.5em}
    \caption{
    {\color{\colored} Influence of occlusion textures.
   (a) Different types of occlusions with the blue lines indicating the cropped slits where the event camera can see the occluded scene. (b) Reconstruction results under different foreground occlusions. (c) We employ the high contrast vertical stripes in (a) with different $r_t$ as occlusions and provide the corresponding results, where $r_t$ denotes the ratio of the number of occlusion edges for $\mathcal{E}_{\theta}^{OO}$ (boundaries of stripes) to that for  $\mathcal{E}_{\theta}^{OA}$ (boundaries of cropped slits).
   }
    }
    \label{fig:TexturedOcc}
    \vspace{-1em}
\end{figure}

{\color{\colored}
\subsubsection{Influence of Occlusion Texture}
The noise events $\mathcal{E}_{\theta}^{OO}$ triggered by the high contrast occlusion edges will interfere with  reconstruction.
To study the influence of occlusion texture, we perform experiments with two additional occlusions as depicted in Fig.~\ref{fig:TexturedOcc-Occ}, where black cardboard triggers few noise events with homogeneous texture and vertical stripes emit a large number of noise events $\mathcal{E}^{OO}_{\theta}$ with high contrast stripe boundaries.
\par 
As shown in Fig.~\ref{fig:TexturedOcc-Res}, the reconstruction performance is related to the number of noise events, where our E-SAI method generates the most reliable texture and brightness under the occlusion of black cardboard while the results of vertical stripes are disturbed by heavy noise events $\mathcal{E}_{\theta}^{OO}$. To further qualify the influence of occlusion texture on our E-SAI method, we employ multiple vertical stripes and denote $r_t$ as the ratio of the number of occlusion edges for $\mathcal{E}_{\theta}^{OO}$ (boundaries of stripes) to that for $\mathcal{E}_{\theta}^{OA}$ (boundaries of cropped slits) with higher $r_t$ indicating more foreground texture to emit $\mathcal{E}_{\theta}^{OO}$. As shown in Fig.~\ref{fig:TexturedOccInflu}, although noise events $\mathcal{E}_{\theta}^{OO}$ will bring disturbances to the reconstruction results, the overall texture of occluded scenes is still successfully restored by our E-SAI method even under severe noise disturbances, \eg, $r_t=1$.

}

{\color{\colored}
\subsubsection{Ablation Studies of the Hybrid Network}

In this section, we analyze the hybrid architecture of our reconstruction network. For encoders, we implement two counterpart networks by replacing our 3-layer SNN encoder with a 3-layer CNN, \ie, E-SAI+CNN, and a 3-layer RNN (using ConvLSTM as \cite{rebecq2019high}) for comparison. The results in Tab.~\ref{tab:DiffEncDec} demonstrate that the RNN encoder surpasses the CNN one by efficiently utilizing the temporal information of events in a recurrent manner. Compared with RNN, our SNN encoder not only exploits spatio-temporal information in events but also alleviates the influence of noise events via the leakage and spike firing mechanisms in spiking neurons, achieving the best performance. 
\par 
Although SNN shows promising results in processing spatio-temporal events, we opt for the hybrid architecture instead of pure SNNs since (i) \textit{vanishing spike phenomenon}: it is commonly observed in deep SNNs that spike activities dramatically reduce as the network depth grows \cite{pandaScalableEfficientAccurate2020}, which often causes information loss in the results; (ii) \textit{low numerical precision}: SNN outputs sparse and binary spike signals (in input and hidden layers), which are usually not sufficient for high-quality image restoration tasks.
In our experiments, we design two decoders with 5-layer SNN and 7-layer SNN, denoted by SNN-5 and SNN-7, respectively. As depicted in Fig.~\ref{fig:DiffEncDec}, SNN-5 cannot restore high quality details of target scenes due to the low numerical precision issue, and SNN-7 totally fails in reconstruction since spikes vanish in the output layer. 
Apparently, our proposed hybrid architecture not only alleviates the disturbances from noise events but also prevents the vanishing spike phenomenon and low numerical precision problems of SNN, and thus achieves the best reconstruction performance.

\begin{figure*}[t!]
	\centering
	\centering
		\begin{subfigure}[b]{0.156\textwidth}
			\includegraphics[width=\textwidth, trim={0cm 0 0cm 0}]{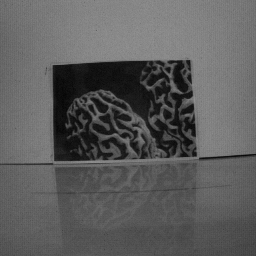}
			\vspace{-1em}\\
			\includegraphics[width=\textwidth, trim={0cm 0 0cm 0}]{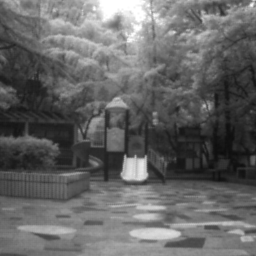}
			\vspace{-1.5em}
			\subcaption*{\scriptsize Reference}
		\end{subfigure}
		\begin{subfigure}[b]{0.156\textwidth}
			\includegraphics[width=\textwidth, trim={0cm 0 0cm 0}]{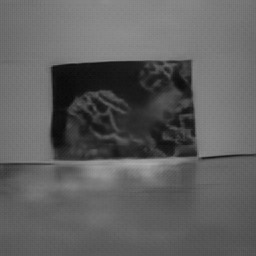}
			\vspace{-1em}\\
			\includegraphics[width=\textwidth, trim={0cm 0 0cm 0}]{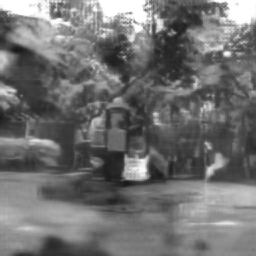}
			\vspace{-1.5em}
			\subcaption*{\scriptsize CNN+CNN}
		\end{subfigure}
		\begin{subfigure}[b]{0.156\textwidth}
			\includegraphics[width=\textwidth, trim={0cm 0 0cm 0}]{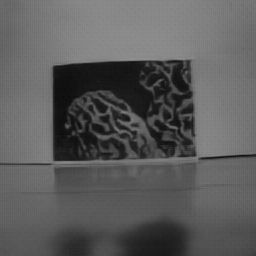}
			\vspace{-1em}\\
			\includegraphics[width=\textwidth, trim={0cm 0 0cm 0}]{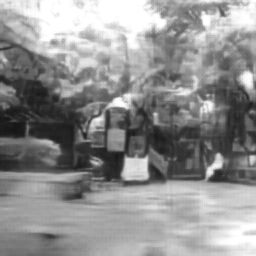}
			\vspace{-1.5em}
			\subcaption*{\scriptsize RNN+CNN}
		\end{subfigure}
		\begin{subfigure}[b]{0.156\textwidth}
			\includegraphics[width=\textwidth, trim={0cm 0 0cm 0}]{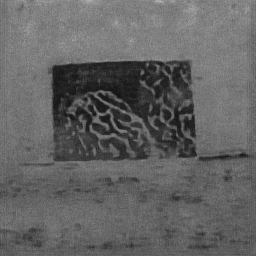}
			\vspace{-1em}\\
			\includegraphics[width=\textwidth, trim={0cm 0 0cm 0}]{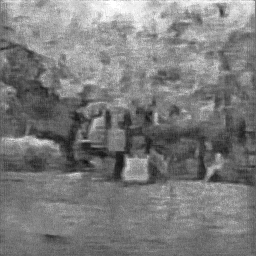}
			\vspace{-1.5em}
			\subcaption*{\scriptsize SNN+SNN-5}
		\end{subfigure}
		\begin{subfigure}[b]{0.156\textwidth}
			\includegraphics[width=\textwidth, trim={0cm 0 0cm 0}]{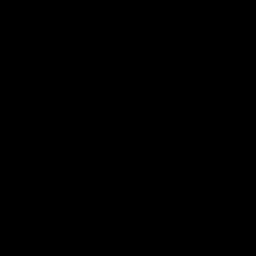}
			\vspace{-1em}\\
			\includegraphics[width=\textwidth, trim={0cm 0 0cm 0}]{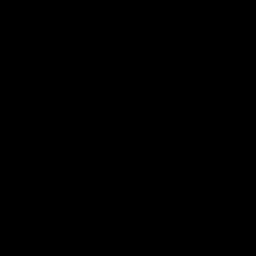}
			\vspace{-1.5em}
			\subcaption*{\scriptsize SNN+SNN-7}
		\end{subfigure}
		\begin{subfigure}[b]{0.156\textwidth}
			\includegraphics[width=\textwidth, trim={0cm 0 0cm 0}]{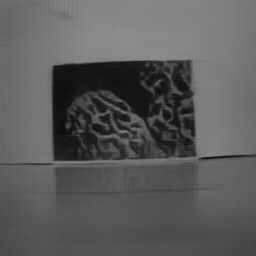}
			\vspace{-1em}\\
			\includegraphics[width=\textwidth, trim={0cm 0 0cm 0}]{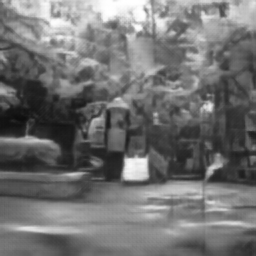}
			\vspace{-1.5em}
			\subcaption*{\scriptsize SNN+CNN (Ours)}
		\end{subfigure}
		\vspace{-.5em}
    \caption{
    {\color{\colored}
   Qualitative results of E-SAI with different reconstruction networks.
   }
    }
    \vspace{-1em}
    \label{fig:DiffEncDec}
\end{figure*}

\begin{table}[t!]
    \centering
    \renewcommand{\arraystretch}{1.3}
    \caption{
    {\color{\seccolored}
    Quantitative results averaged over all test sequences with the same Refocus-Net and different reconstruction networks (Recon-Nets), where SNN-5 and SNN-7 indicate 5-layer and 7-layer spiking neural networks, respectively.
    }
    }
    \vspace{-.5em}
\begin{tabular}{c|c|c}
\hline
\multicolumn{1}{c|}{\textbf{Recon-Net}}  & \multicolumn{1}{c|}{\textbf{Indoor}}                                    & \multicolumn{1}{c}{\textbf{Outdoor}}                                    \\ \hline
Enc.+Dec. & PSNR / SSIM / LPIPS            &  PSNR / SSIM / LPIPS          \\ \hline
CNN+CNN     & 26.53 / 0.7653 / 0.0871          & 15.98 / 0.5240 / 0.1838          \\
RNN+CNN     & \underline{28.38} / \underline{0.7993} / \underline{0.0571}          & \underline{18.26} / \underline{0.6477} / \underline{0.1244}          \\
SNN+SNN-5   & 23.34 / 0.5015 / 0.1298          & 15.61 / 0.4334 / 0.1964          \\
SNN+SNN-7   & 7.33 / 0.0008 / 0.7173          & 6.36 / 0.0006 / 0.8610          \\
SNN+CNN     & \textbf{29.55} / \textbf{0.8086} / \textbf{0.0546} & \textbf{20.39} / \textbf{0.6961} / \textbf{0.1013} \\ \hline
\end{tabular}
\label{tab:DiffEncDec}
\vspace{-.5em}
\end{table}

\begin{table}[t!]
    \centering
    \renewcommand{\arraystretch}{1.3}
    \caption{
    {\color{\seccolored}
    Quantitative results of E-SAI with different numbers of time intervals $N$, averaged over all test sequences.
    }
    }
    \vspace{-.5em}
   \begin{tabular}{c|c|c}
\hline
\multirow{2}{*}{\textbf{N}} & \multicolumn{1}{c|}{\textbf{Indoor}}                                    & \multicolumn{1}{c}{\textbf{Outdoor}}                                    \\ \cline{2-3} 
                            & PSNR / SSIM / LPIPS          &  PSNR / SSIM / LPIPS           \\ \hline
1                           & 29.30 / 0.8073 / 0.0491          & 17.80 / 0.5714 / 0.1635          \\
3                           & 29.89 / 0.8156  / 0.0421          & 18.99 / 0.6067 / 0.1412          \\
6                           & 30.22 / 0.8176 / 0.0398          & 19.47 / 0.6139 / 0.1377          \\
15                          & 30.60 / 0.8218 / 0.0377          & 20.31 / 0.6586 / 0.1081          \\
30                          & \textbf{30.71} / \textbf{0.8311} / \textbf{0.0374} & \textbf{20.39} / \textbf{0.7037} / \textbf{0.0981} \\ \hline
\end{tabular}
\label{tab:DiffN}
\vspace{-1em}
\end{table}

\begin{figure}[t!]
	\centering
	\begin{subfigure}[b]{0.225\linewidth}
			\includegraphics[width=\linewidth, trim={0cm 0 0cm 0}]{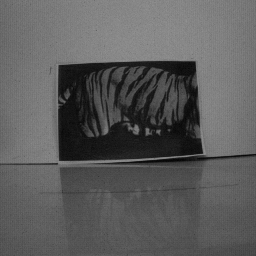}
			\vspace{-1em}
			\\
			\includegraphics[width=\linewidth, trim={0cm 0 0cm 0}]{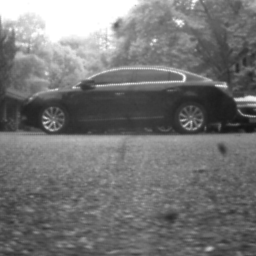}
			\vspace{-1.5em}
			\subcaption*{\scriptsize Reference}
		\end{subfigure}
		\begin{subfigure}[b]{0.225\linewidth}
			\includegraphics[width=\linewidth, trim={0cm 0 0cm 0}]{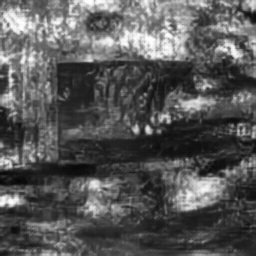}
			\vspace{-1em}
			\\
			\includegraphics[width=\linewidth, trim={0cm 0 0cm 0}]{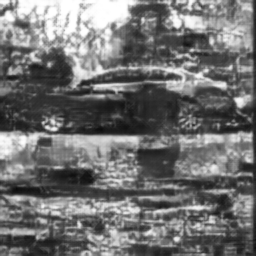}
			\vspace{-1.5em}
			\subcaption*{\scriptsize 5}
		\end{subfigure}
		\begin{subfigure}[b]{0.225\linewidth}
			\includegraphics[width=\linewidth, trim={0cm 0 0cm 0}]{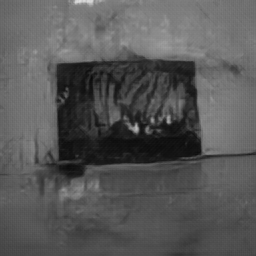}
			\vspace{-1em}
			\\
			\includegraphics[width=\linewidth, trim={0cm 0 0cm 0}]{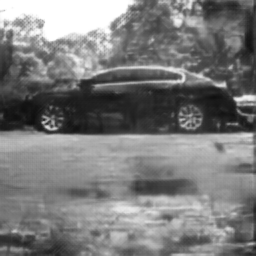}
			\vspace{-1.5em}
			\subcaption*{\scriptsize 15}
		\end{subfigure}
		\begin{subfigure}[b]{0.225\linewidth}
			\includegraphics[width=\linewidth, trim={0cm 0 0cm 0}]{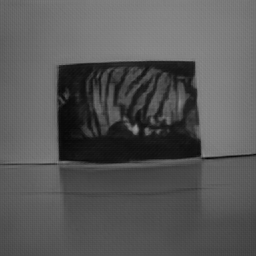}
			\vspace{-1em}
			\\
			\includegraphics[width=\linewidth, trim={0cm 0 0cm 0}]{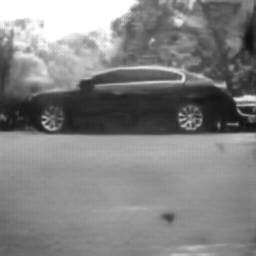}
			\vspace{-1.5em}
			\subcaption*{\scriptsize 30}
		\end{subfigure}
	\vspace{-.5em}
	\caption{
	{\color{\colored}
	Qualitative results of E-SAI with events of the first 5, 15, and 30 time intervals.} 
	}
	\label{fig:DiffStep}
    \vspace{-1em}
\end{figure}
\subsubsection{Analysis of Hyper-parameters}
The number of time intervals $N$ controls the granularity of temporal information in input events. In Tab.~\ref{tab:DiffN}, we compare the performance of E-SAI+Hybrid with different $N$. When $N=1$, the temporal information in events is discarded and only the spike firing mechanism in Eq.~\eqref{fire} functions, which benefits noise suppression since the spiking threshold acts as a blocker for noise events. As $N$ increases, the reconstruction performance becomes better since both leakage and spike firing mechanisms contribute to noise suppression. In addition, the leakage mechanism gains more performance improvements with larger $N$ since finer temporal granularity helps spiking neurons to leak out the influence of noise events. 
Thus, we set $N=30$ to balance reconstruction performance and computational efficiency. 
\par
We also study the performance of E-SAI+Hybrid with partial input events under fixed $N=30$. Specifically, we first record the SNN outputs with events at the first 5, 15, and 30 time intervals, and then feed them to the CNN decoder to obtain the intermediate reconstruction results. 
As shown in Fig.~\ref{fig:DiffStep}, the reconstruction suffers from noise and loss of details if only with events of the first 5 time intervals. When more events are fed to the network, \eg, the first 15 or 30 time intervals, more signal information in events is utilized and the leakage mechanism in spiking neurons better functions to alleviate noise, leading to visual results with smoother texture and less noise.

}

{\color{\colored}

\subsection{Limitations and Future Works}\label{sec:exp-limitations}

The principle of SAI is to see through occlusions from multi-view observations where the targets should be partially observed, and the same is true for our E-SAI. For the areas occluded from all viewpoints, \eg, horizontal camera motions with horizontal occlusions shown in Fig.~\ref{fig:Failure-horizontal}, the proposed E-SAI cannot achieve successful reconstructions due to the lack of signal information. This issue may be addressed by developing E-SAI methods with a 4D event field using more flexible camera motions, \eg, both horizontal and vertical motions, to acquire more effective observations of the occluded scenes. 
\par 
The proposed E-SAI system can handle a large set of natural occluded scenes where the brightness of occlusion boundaries is almost uniform with limited variations. For some special cases where the boundaries of occlusions have significant varying intensities and provide different offsets, our E-SAI system may fail to reconstruct occluded targets or give reversed intensities. As shown in Fig.~\ref{fig:Failure-occ}, the reconstruction results are not consistent with the target scenes under the occlusion of horizontal stripes, where intensities vary in occlusion edges and offer different brightness offsets for different regions. Chessboard not only disturbs the reconstruction via varying brightness offsets but also emits a massive number of noise events $\mathcal{E}_{\theta}^{OO}$, resulting in failure reconstruction. This problem could be tackled by combining frames and events to form a multi-modal SAI method, such as \cite{liao2022synthetic}, where the occlusion intensities contained in frames could provide reconstruction guidance for events.
\par 
Moreover, our method is tailored to restore the occluded targets with small depth variation as discussed in Sec.~\ref{sec:autoRefocus}. For multi-depth scenes, multiple refocusing procedures might be required to align events triggered at different depths.
To fulfill all-in-focus reconstruction, several techniques could be incorporated such as image matting \cite{7583662} and depth estimation \cite{yangContinuouslyTrackingSeethrough2011}, and we leave it for future work.

\begin{figure*}[t!]
	\centering
	\begin{subfigure}{0.468\textwidth}
	\begin{subfigure}[b]{0.32\textwidth}
			\includegraphics[width=\textwidth, trim={0cm 0 0cm 0}]{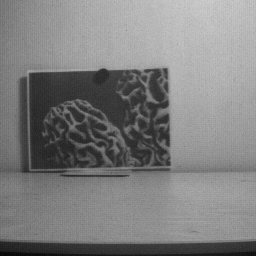}
			\vspace{-1em}\\
			\includegraphics[width=\textwidth, trim={0cm 0 0cm 0}]{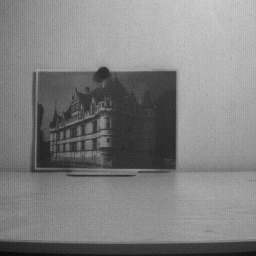}
			\vspace{-1.5em}
			\subcaption*{\scriptsize Reference}
		\end{subfigure}
		\begin{subfigure}[b]{0.32\textwidth}
			\includegraphics[width=\textwidth, trim={0cm 0 0cm 0}]{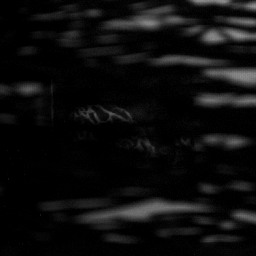}
			\vspace{-1em}\\
			\includegraphics[width=\textwidth, trim={0cm 0 0cm 0}]{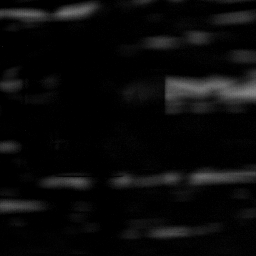}
			\vspace{-1.5em}
			\subcaption*{\scriptsize Occluded view}
		\end{subfigure}
		\begin{subfigure}[b]{0.32\textwidth}
			\includegraphics[width=\textwidth, trim={0cm 0 0cm 0}]{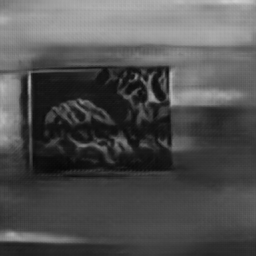}
			\vspace{-1em}\\
			\includegraphics[width=\textwidth, trim={0cm 0 0cm 0}]{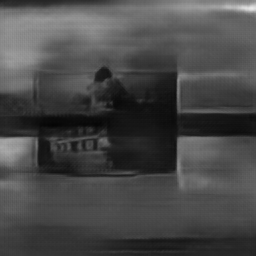}
			\vspace{-1.5em}
			\subcaption*{\scriptsize Reconstruction}
		\end{subfigure}
		\caption{Horizontal wooden fence}
		\label{fig:Failure-horizontal}
	\end{subfigure}
	\begin{subfigure}{0.468\textwidth}
	\begin{subfigure}[b]{0.32\textwidth}
			\includegraphics[width=\textwidth, trim={0cm 0 0cm 0}]{fig2/figure-texturedOcc/gt-007.png}
			\vspace{-1em}\\
			\includegraphics[width=\textwidth, trim={0cm 0 0cm 0}]{fig2/figure-texturedOcc/gt-004.png}
			\vspace{-1.5em}
			\subcaption*{\scriptsize Reference}
		\end{subfigure}
		\begin{subfigure}[b]{0.32\textwidth}
			\includegraphics[width=\textwidth, trim={0cm 0 0cm 1cm}]{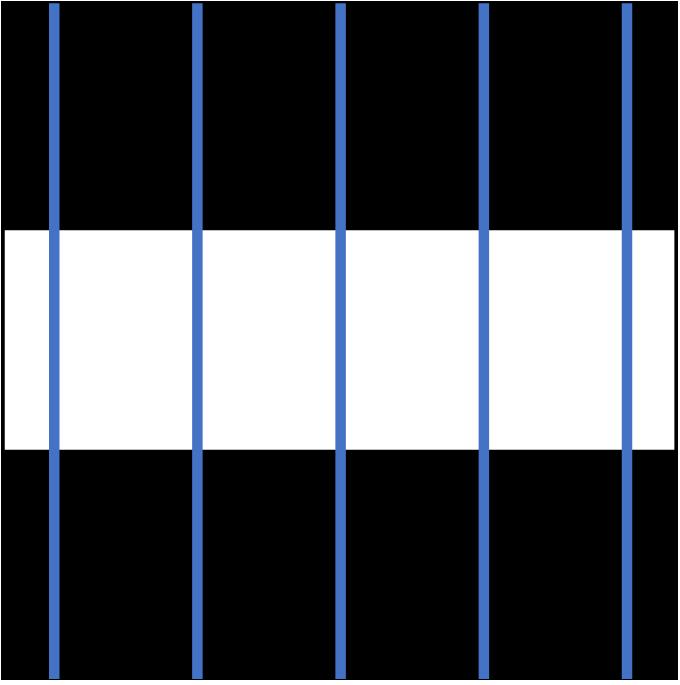}
			\vspace{-1em}\\
			\includegraphics[width=\textwidth, trim={0cm 0 0cm 0.8mm}]{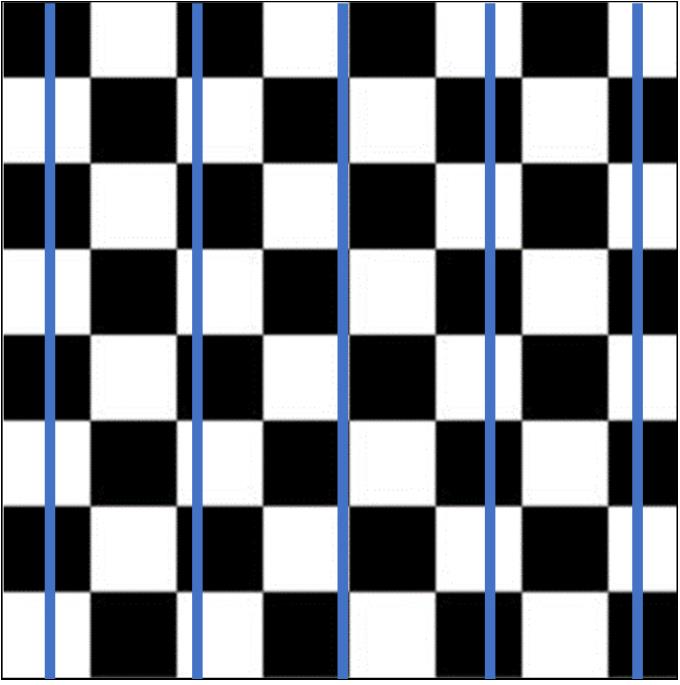}
			\vspace{-1.5em}
			\subcaption*{\scriptsize Occlusion}
		\end{subfigure}
		\begin{subfigure}[b]{0.32\textwidth}
			\includegraphics[width=\textwidth, trim={0cm 0 0cm 0}]{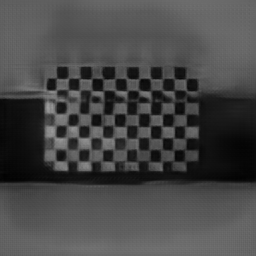}
			\vspace{-1em}\\
			\includegraphics[width=\textwidth, trim={0cm 0 0cm 0}]{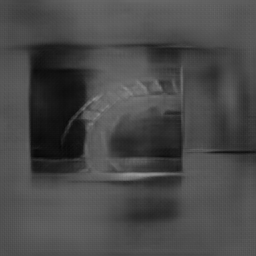}
			\vspace{-1.5em}
			\subcaption*{\scriptsize Reconstruction}
	\end{subfigure}
	    \caption{Horizontal stripes and chessboard}
		\label{fig:Failure-occ}
	\end{subfigure}
	\vspace{-.5em}
	\caption{
	{\color{\colored}
	Examples of failure cases. (a) Reconstruction results under horizontal dense wooden fence. (b) Reconstruction results under the occlusions of horizontal stripes and chessboard, where the blue lines indicate the cropped slits where the event camera can see the occluded scene.} 
	}
	\label{fig:Failure}
\end{figure*}

}

\section{Conclusion}
In this work, we propose a novel SAI method based on event cameras (E-SAI). Benefiting from the  extremely low latency and the high dynamic range of event cameras, E-SAI can achieve the seeing-through effects under very dense occlusions and extreme lighting conditions. Specifically, a Refocus-Net and a hybrid SNN-CNN network are proposed to effectively reconstruct the visual images of occluded scenes from collected events. Benefiting from the combination of SNN and CNN, the spatio-temporal information of events is well utilized, and the reconstruction quality of occluded targets is guaranteed.
We also construct a new SAI dataset containing both events and frames for a variety of indoor and outdoor scenes.
Quantitative and qualitative results over the SAI dataset show the effectiveness of our proposed E-SAI method and validate its superiority to the frame-based counterpart, \ie, F-SAI, under dense occlusions and extreme lighting scenes.

\section*{\Large Appendix}
\section{Network Architecture Details}
\begin{figure}[htp]
	\centering
	\includegraphics[width=.5\textwidth]{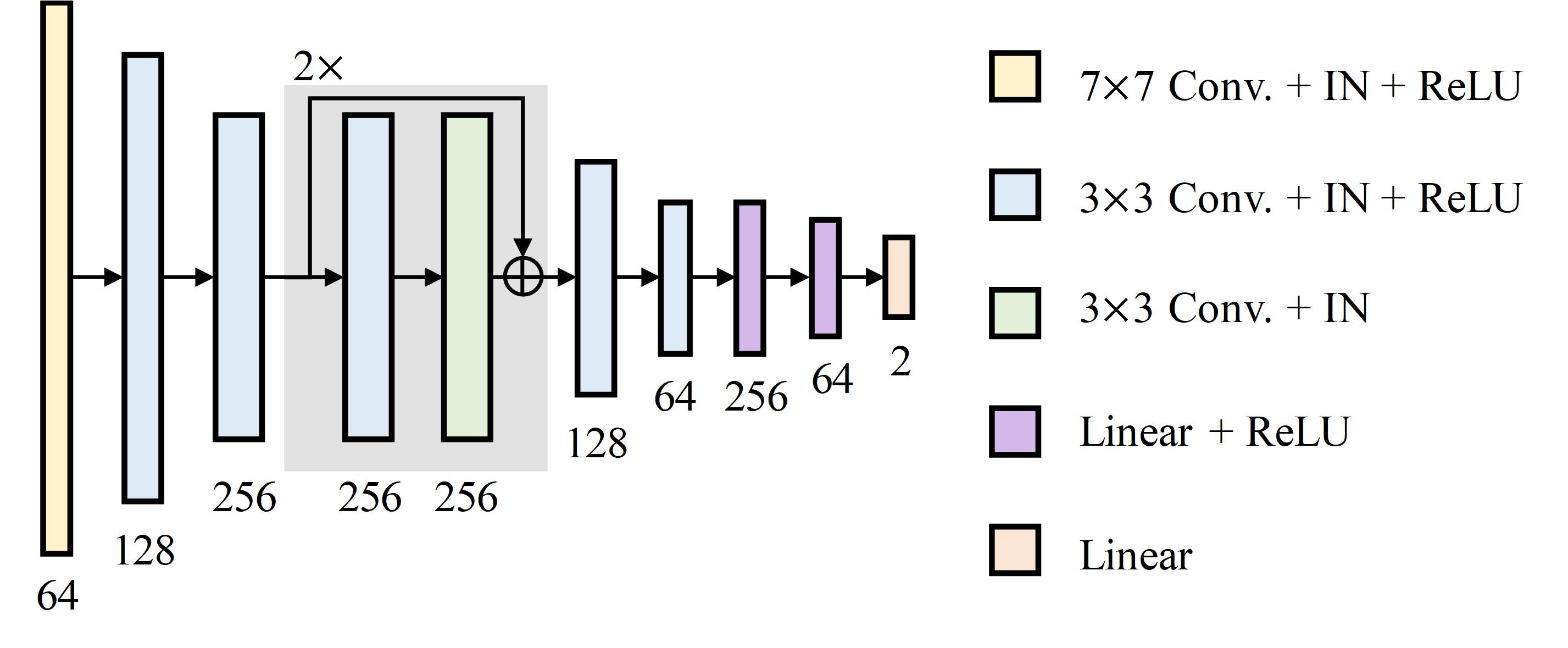}
    \caption{Architecture of our Refocus-Net. Conv. indicates convolutional layer, IN denotes instance normalization, and ReLU represents rectified linear unit.\vspace{-1em}}
    \label{fig:RefocusNet}
\end{figure}

Tab. \ref{tab:network} shows the detailed architecture of our Refocus-Net and hybrid reconstruction network. Refocus-Net receives a 2$N$-channel tensor stacked from unfocused event frames and outputs two scalars respectively corresponding to horizontal and vertical warping parameters of $\boldsymbol{\psi}$, as depicted in Fig.~\ref{fig:RefocusNet}. In the encoders of our hybrid and CNN reconstruction networks, skip connections are added between the outputs of 1$^{st}$ and 2$^{nd}$ convolution layers for a fair comparison. 
\begin{table}[t!]
\renewcommand{\arraystretch}{1.3}
\centering
\caption{
{\color{\colored}
Detailed network architecture. The input sizes of Refocus-Net and hybrid reconstruction network (Recon-Net) are (260,346) and (256,256), respectively. (S-)Conv indicates (spiking) convolutional layer; deConv denotes transposed convolutional layer; Res represents residual block implemented as \cite{zhu2017unpaired}; FC indicates fully connected layer; IN/BN denotes instance/batch normalization; ReLU represents rectified linear unit; Tanh indicates hyperbolic tangent function. Note that $2\times$ ($15\times$) Res means 2 (15) cascaded residual blocks.}
}
\begin{tabular}{c|ccc}
\hline
Network                                                                      & Layer          & Kernel & Shape         \\ \hline
\multirow{9}{*}{Refocus-Net}                                                 & Conv-IN-ReLU   & (7,7)  & (64,254,340)  \\
                                                                             & Conv-IN-ReLU   & (3,3)  & (128,126,169) \\
                                                                             & Conv-IN-ReLU   & (3,3)  & (256,62,84)   \\
                                                                             & 2x Res-IN      & (3,3)  & (256,62,84)   \\
                                                                             & Conv-IN-ReLU   & (3,3)  & (128,30,41)   \\
                                                                             & Conv-IN-ReLU   & (3,3)  & (64,14,20)    \\
                                                                             & FC-ReLU        & -      & 256           \\
                                                                             & FC-ReLU        & -      & 64            \\
                                                                             & FC             & -      & 2             \\ \hline
\multirow{3}{*}{\begin{tabular}[c]{@{}c@{}}Recon-Net\\ Encoder\end{tabular}} & S-Conv         & (1,1)  & (8,256,256)   \\
                                                                             & S-Conv         & (3,3)  & (16,256,256)  \\
                                                                             & S-Conv         & (7,7)  & (32,256,256)  \\ \hline
\multirow{7}{*}{\begin{tabular}[c]{@{}c@{}}Recon-Net\\ Decoder\end{tabular}} & Conv-BN-ReLU   & (7,7)  & (64,256,256)  \\
                                                                             & Conv-BN-ReLU   & (3,3)  & (128,128,128) \\
                                                                             & Conv-BN-ReLU   & (3,3)  & (256,64,64)   \\
                                                                             & 15x Res-BN     & (3,3)  & (256,64,64)   \\
                                                                             & deConv-BN-ReLU & (3,3)  & (128,128,128) \\
                                                                             & deConv-BN-ReLU & (3,3)  & (64,256,256)  \\
                                                                             & Conv-Tanh      & (7,7)  & (1,256,256)   \\ \hline
\end{tabular}
\label{tab:network}
\end{table}

\begin{table*}[th!]
    \centering
    \renewcommand{\arraystretch}{1.3}
    \caption{
    {\color{\seccolored} 
    Comparisons between pure CNN and hybrid SNN-CNN for Refocus-Net. The metrics PSNR, SSIM, and LPIPS are measured by concatenating the Refocus-Nets with the same hybrid reconstruction network, and the results are the average over all test sequences of our SAI dataset.
    }
    }
   \begin{tabular}{c|cccc|cccc}
\hline
\multirow{2}{*}{\textbf{Refocus-Net}} & \multicolumn{4}{c|}{\textbf{Indoor}}                                                                           & \multicolumn{4}{c}{\textbf{Outdoor}}                                                                          \\ \cline{2-9} 
                                  & \multicolumn{1}{l|}{APSE $\downarrow$} & \multicolumn{1}{l|}{PSNR $\uparrow$} & \multicolumn{1}{l|}{SSIM $\uparrow$} & \multicolumn{1}{l|}{LPIPS $\downarrow$} & \multicolumn{1}{l|}{APSE $\downarrow$} & \multicolumn{1}{l|}{PSNR $\uparrow$} & \multicolumn{1}{l|}{SSIM $\uparrow$} & \multicolumn{1}{l}{LPIPS $\downarrow$} \\ \hline
                                  
Hybrid SNN-CNN                    & \textcolor{\seccolored}{0.982}                    & 29.46                     & 0.8068                    & 0.0559                     & \textcolor{\seccolored}{1.166}                    & 20.21                     & 0.6924                    & 0.1078    
      \\
Pure CNN (Ours)                       & \textbf{\textcolor{\seccolored}{0.722}}            & \textbf{29.55}            & \textbf{0.8086}           & \textbf{0.0546}            & \textbf{\textcolor{\seccolored}{1.047}}            & \textbf{20.39}            & \textbf{0.6961}           & \textbf{0.1013}     
\\ \hline
\end{tabular}
    \label{tab:HybridRefocus}
    \end{table*}

\begin{figure*}[th!]
	\centering
	\begin{subfigure}[b]{0.16\linewidth}
		\includegraphics[width=\linewidth]{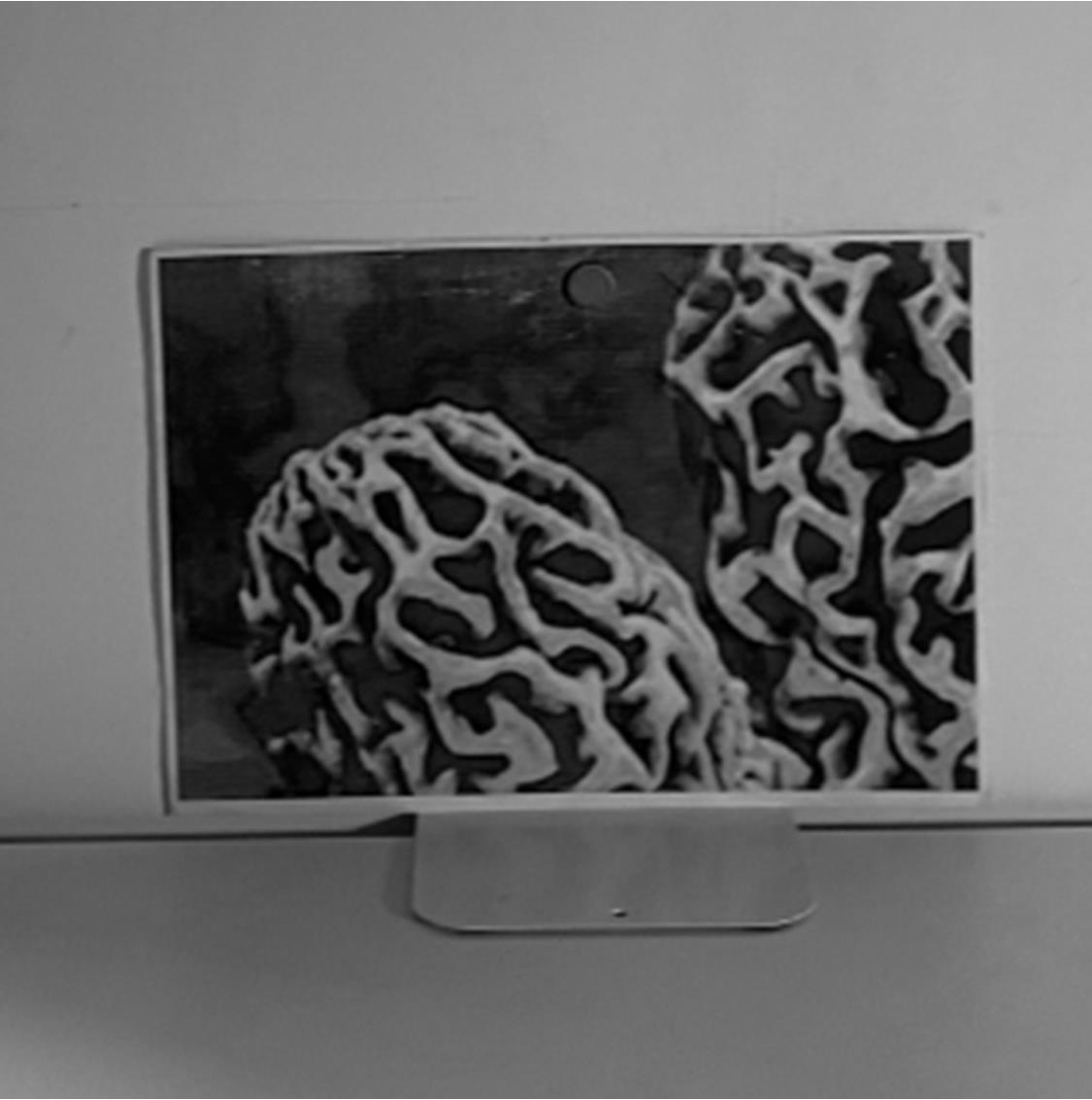}
		\vspace{-.8em}\\
		\includegraphics[width=\linewidth, trim={0cm 0 0cm 0}]{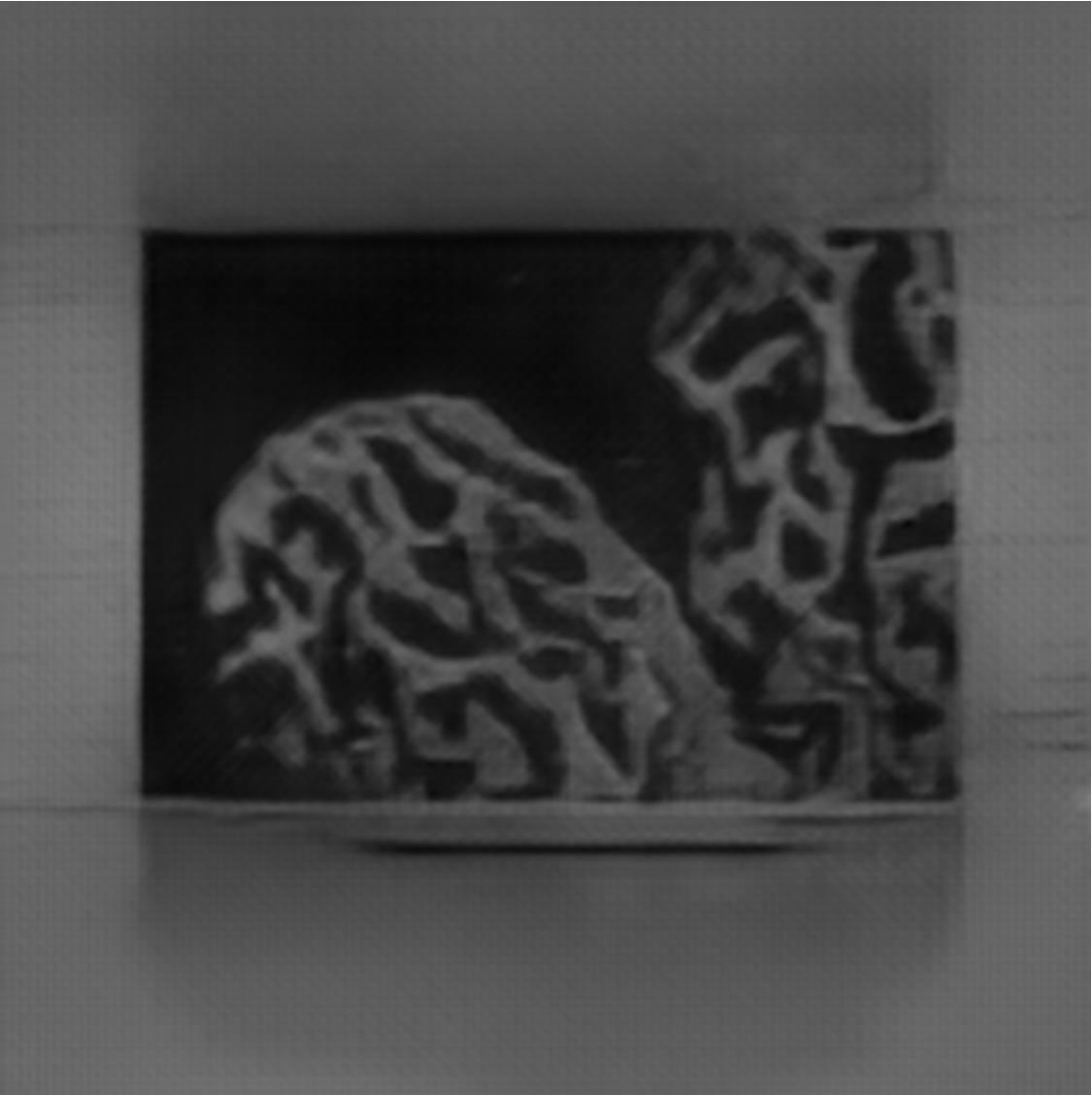}
		\vspace{-1em}
	\end{subfigure}
	\begin{subfigure}[b]{0.16\linewidth}
		\includegraphics[width=\linewidth]{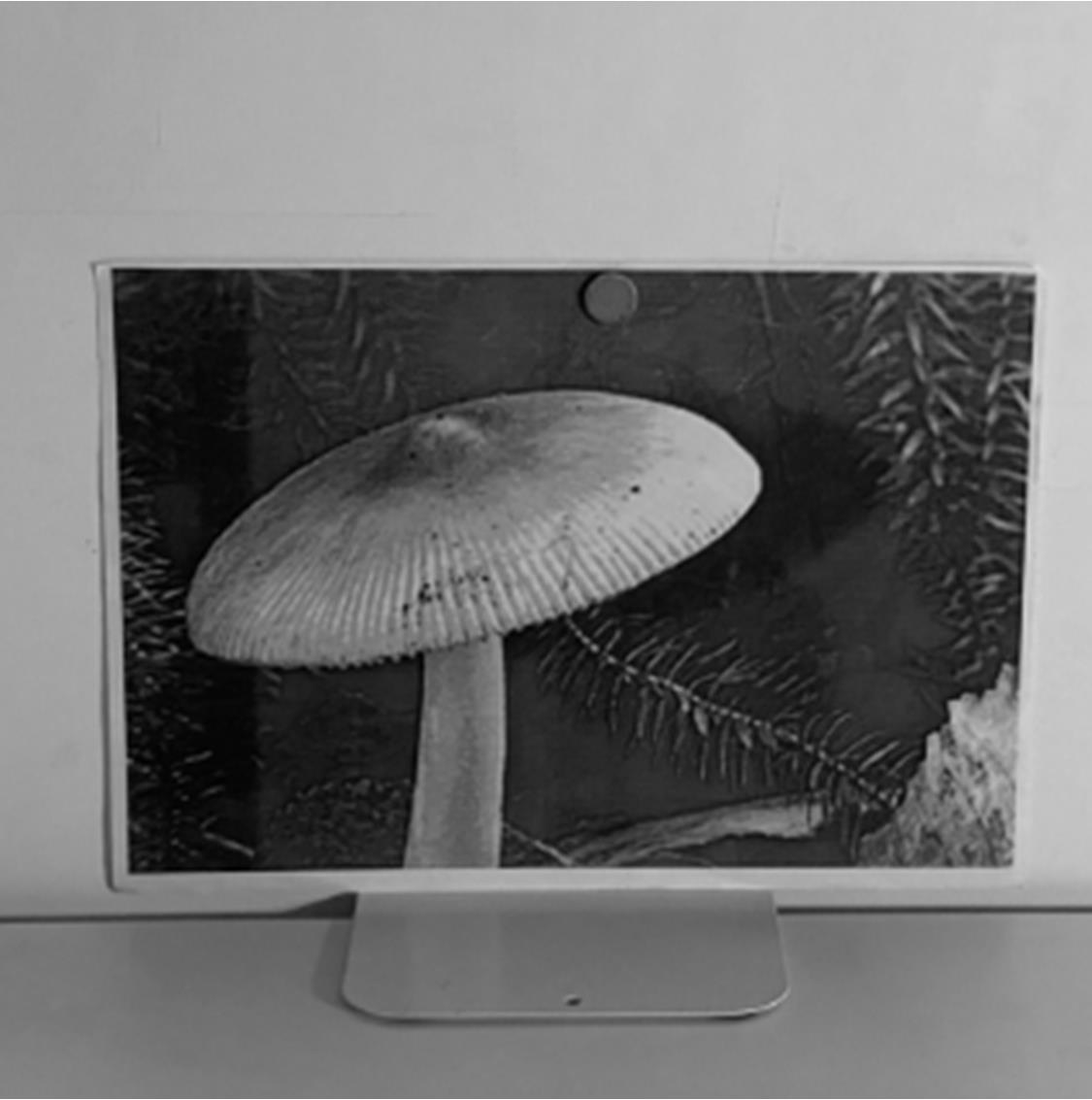}
		\vspace{-.8em}\\
		\includegraphics[width=\linewidth, trim={0cm 0 0cm 0}]{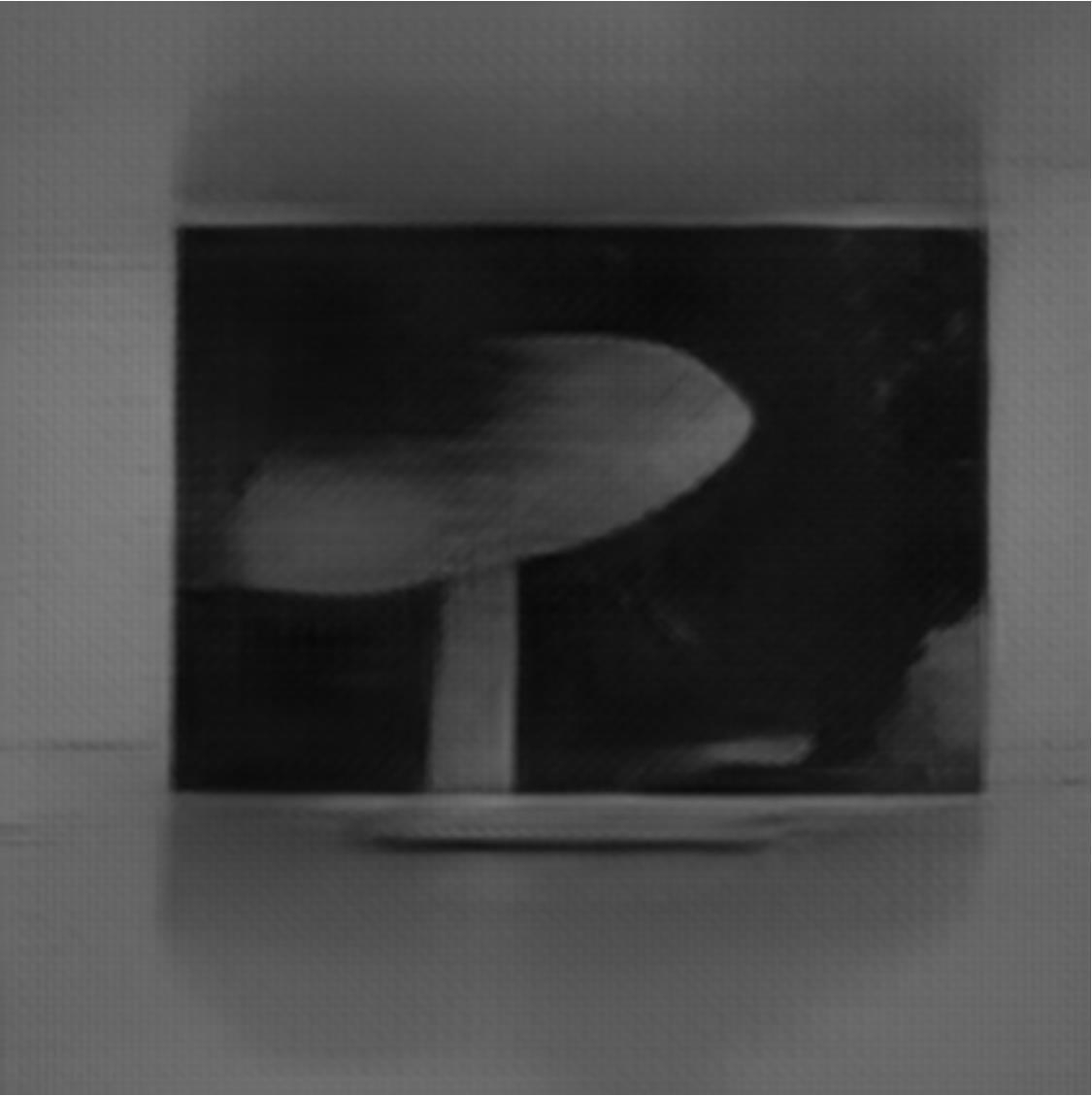}
		\vspace{-1em}
	\end{subfigure}
	\begin{subfigure}[b]{0.16\linewidth}
		\includegraphics[width=\linewidth]{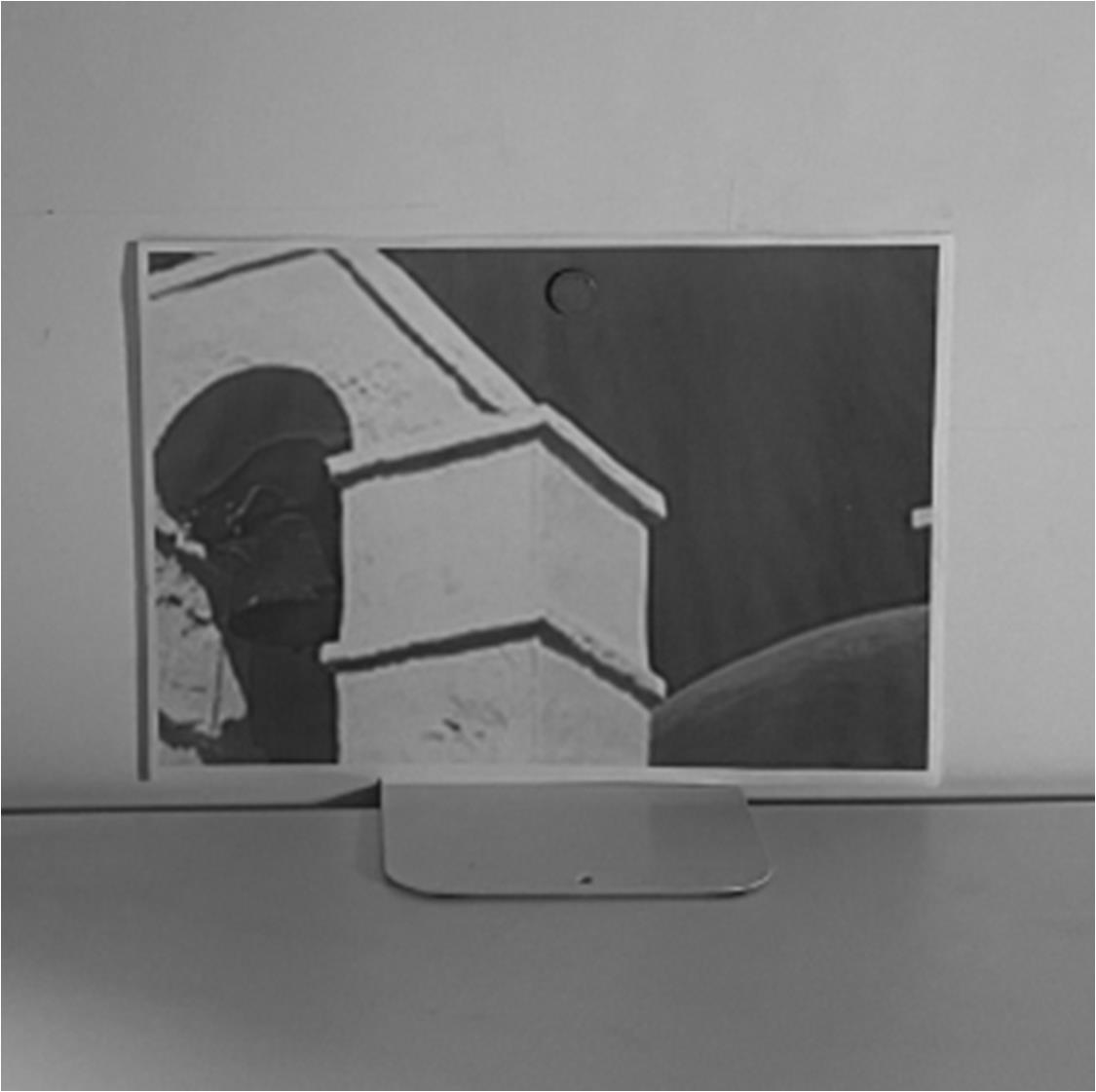}
		\vspace{-.8em}\\
		\includegraphics[width=\linewidth, trim={0cm 0 0cm 0}]{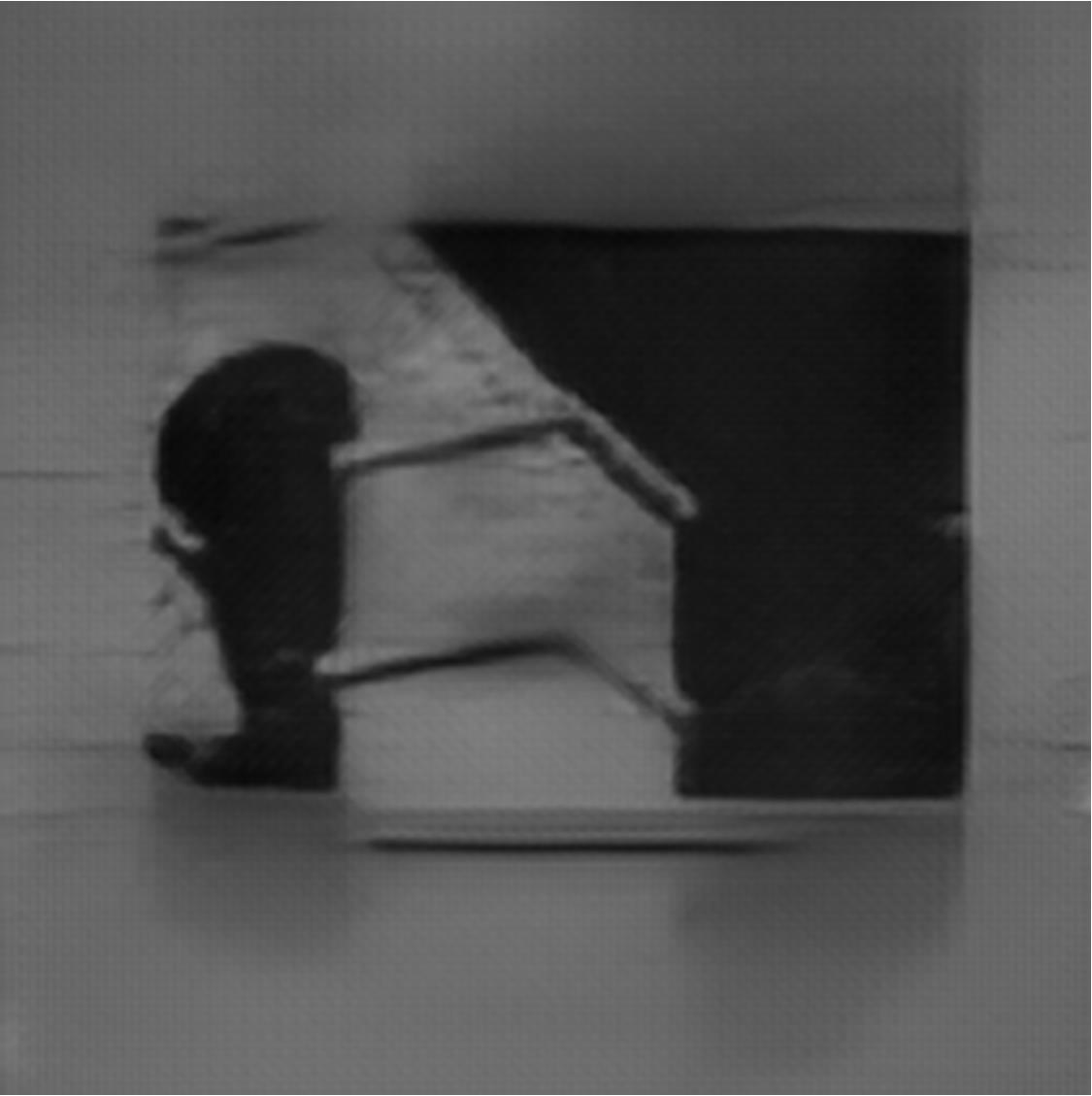}
		\vspace{-1em}
	\end{subfigure}
	\begin{subfigure}[b]{0.16\linewidth}
		\includegraphics[width=\linewidth]{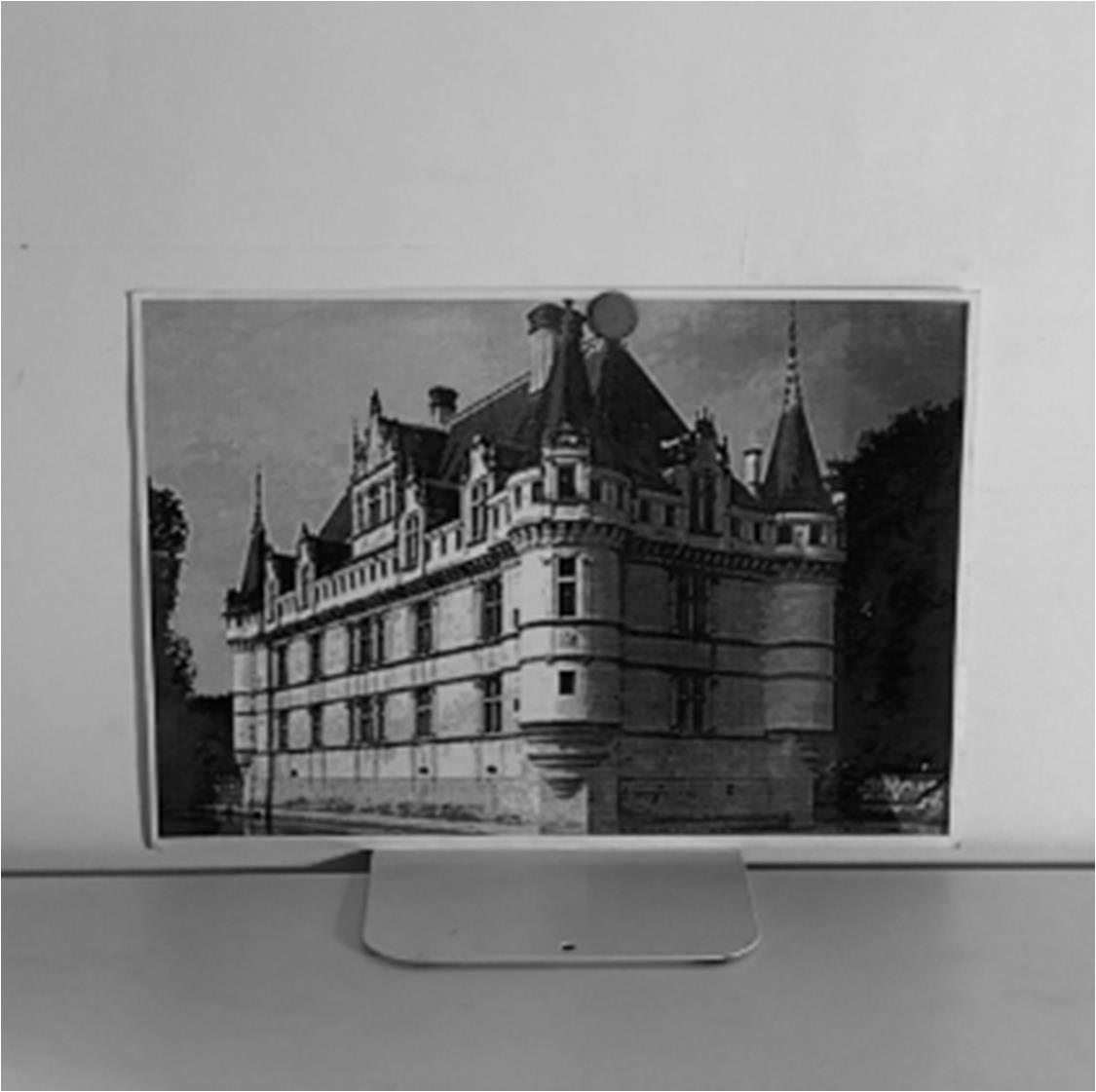}
		\vspace{-.8em}\\
		\includegraphics[width=\linewidth, trim={0cm 0 0cm 0}]{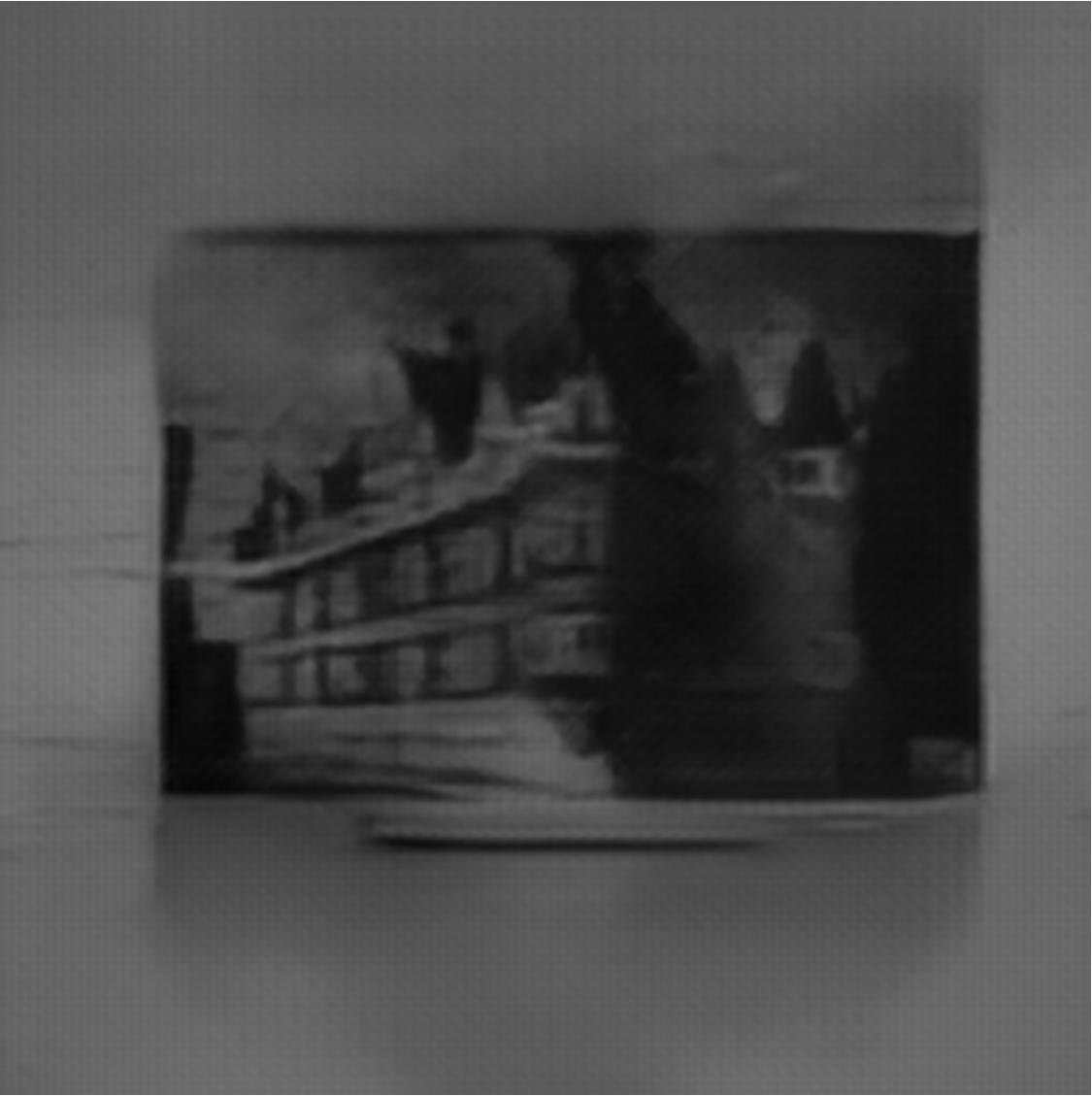}
		\vspace{-1em}
	\end{subfigure}
	\begin{subfigure}[b]{0.16\linewidth}
		\includegraphics[width=\linewidth]{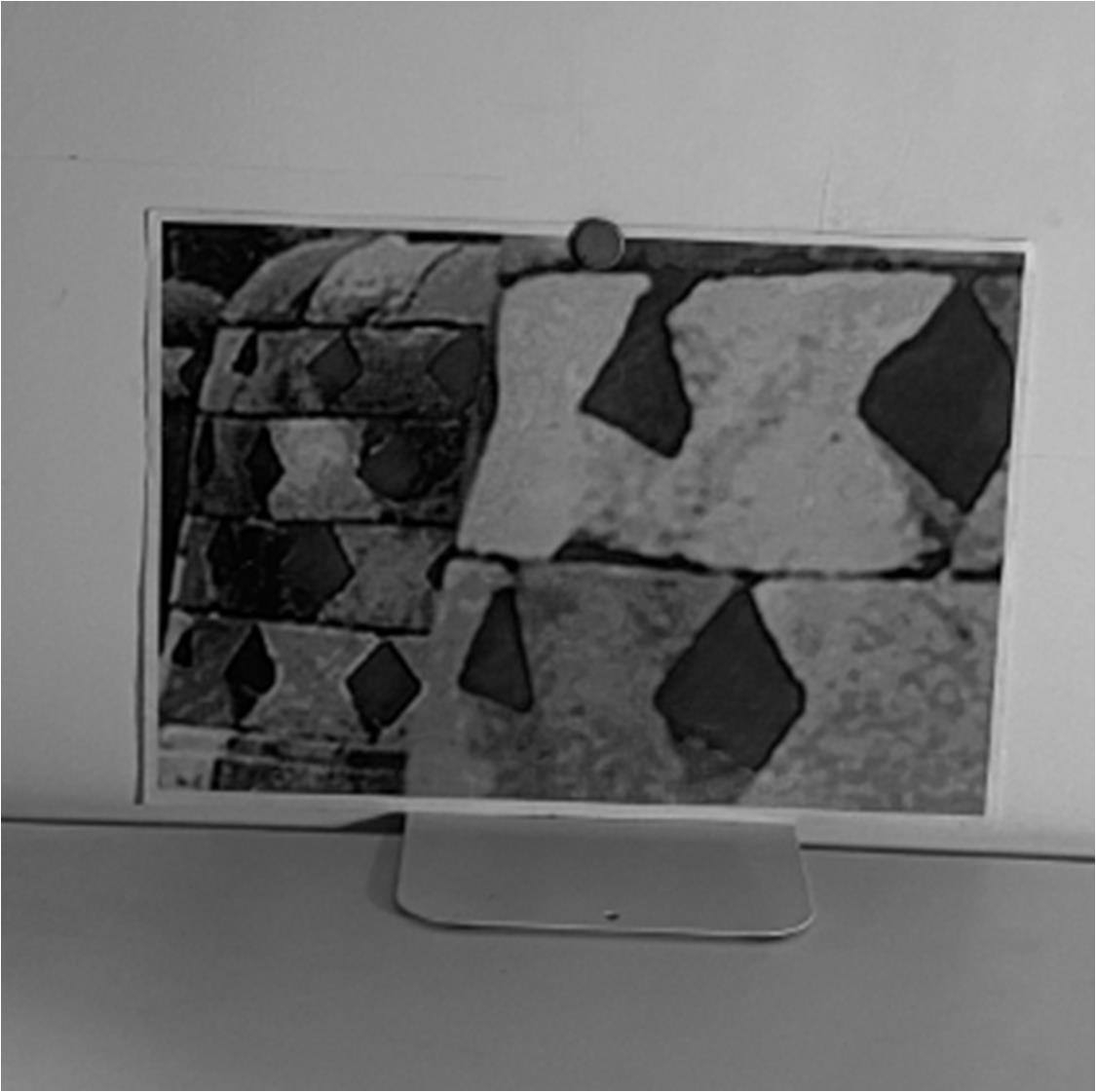}
		\vspace{-.8em}\\
		\includegraphics[width=\linewidth, trim={0cm 0 0cm 0}]{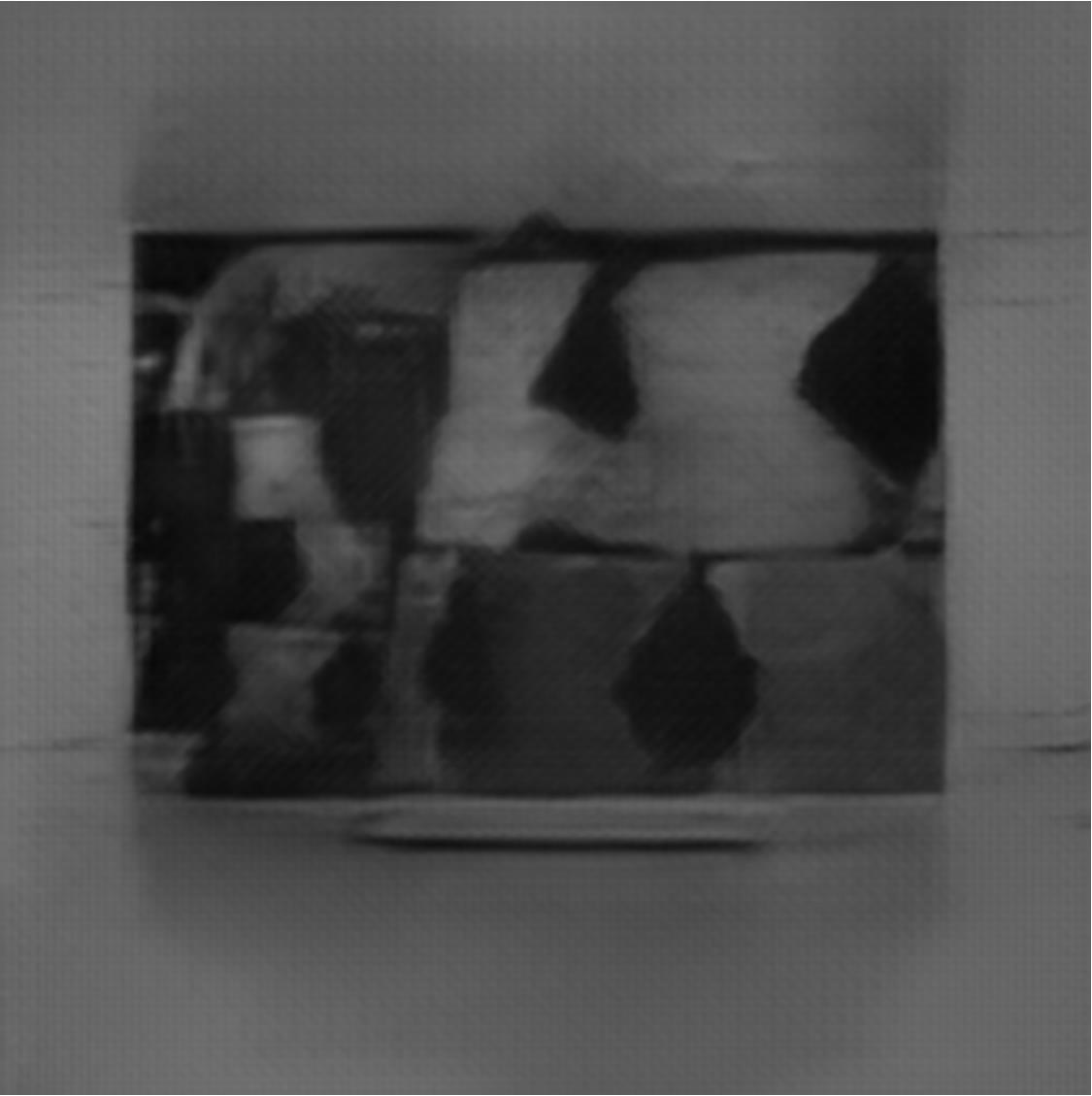}
		\vspace{-1em}
	\end{subfigure}
	\begin{subfigure}[b]{0.16\linewidth}
		\includegraphics[width=\linewidth]{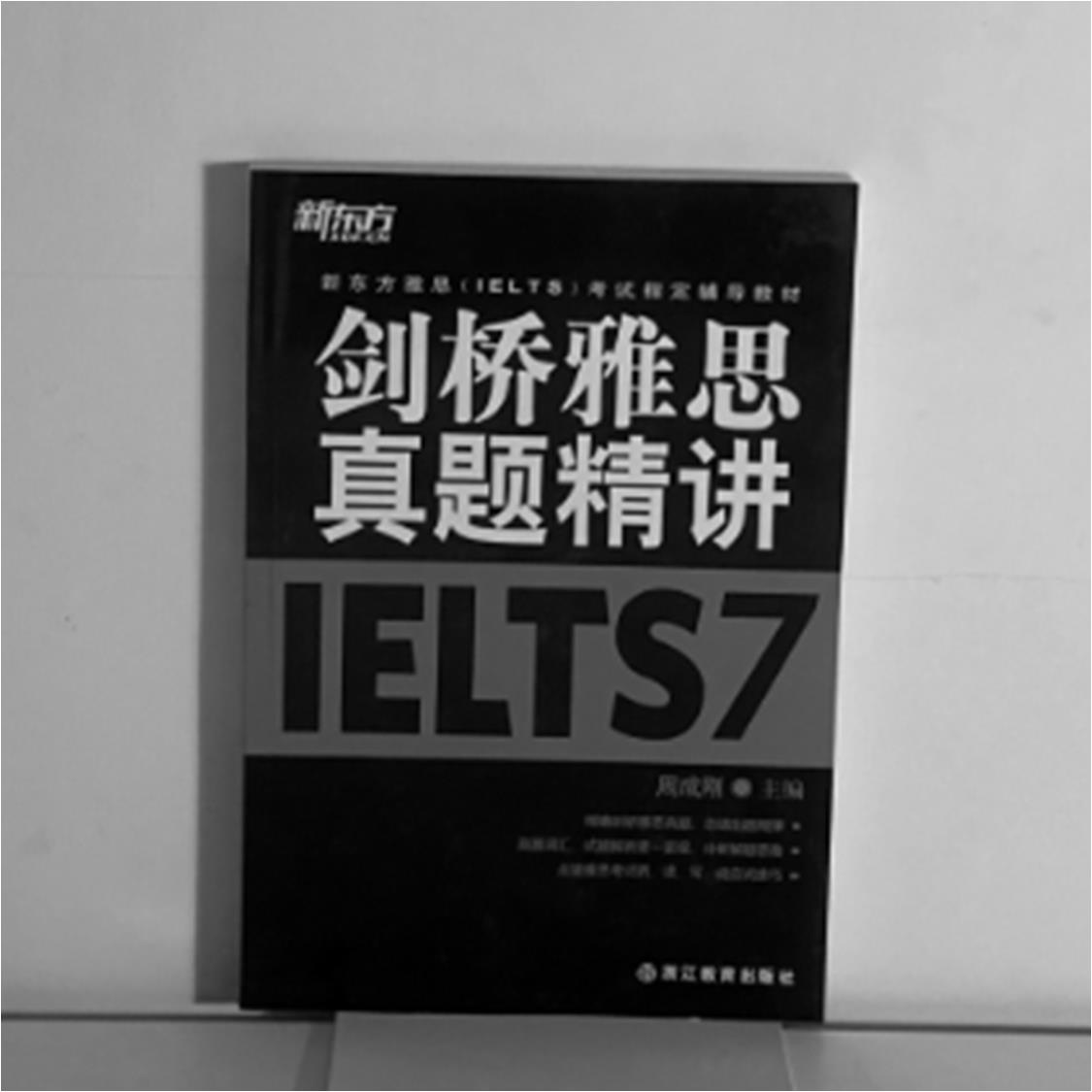}
		\vspace{-.8em}\\
		\includegraphics[width=\linewidth, trim={0cm 0 0cm 0}]{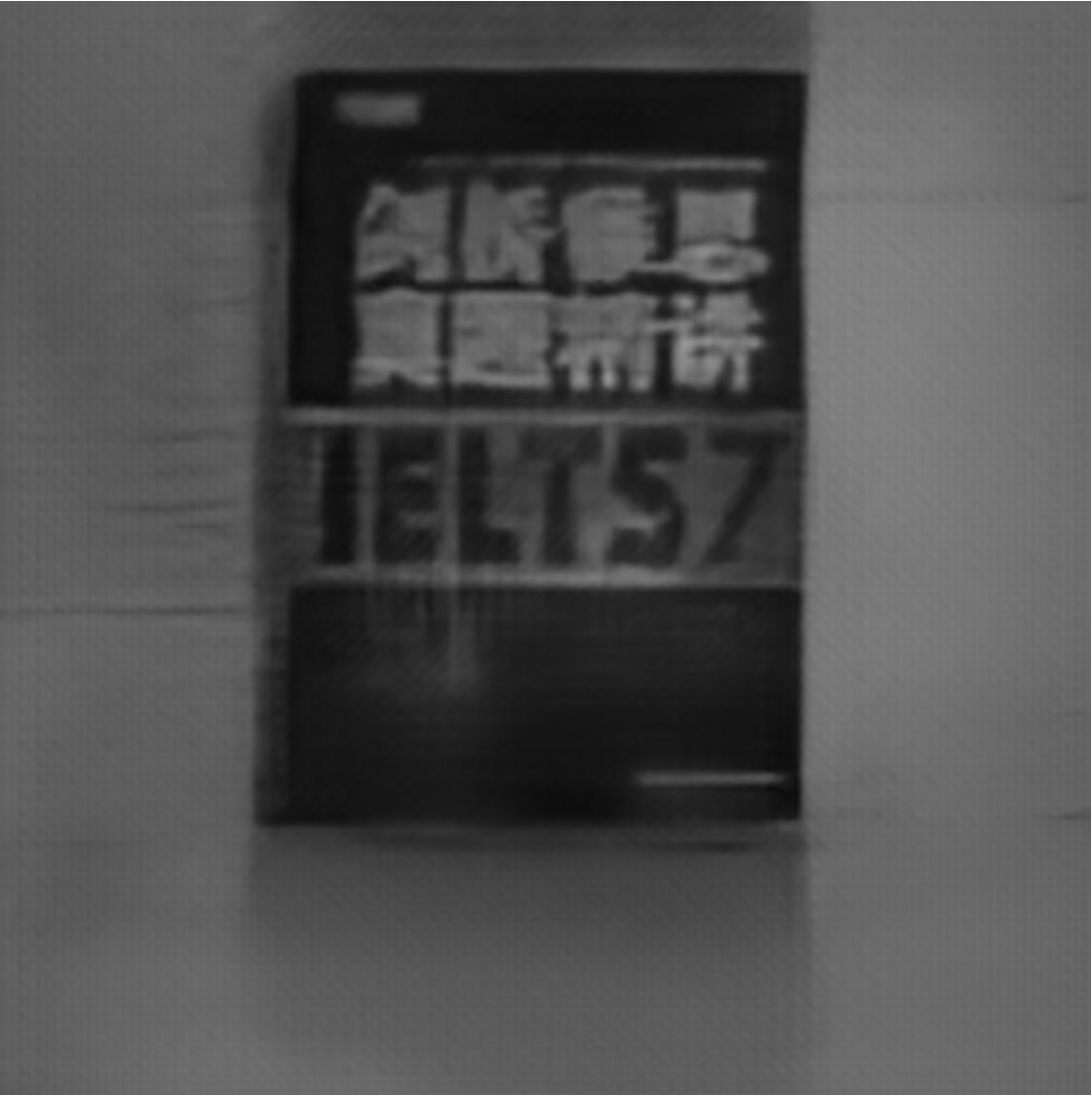}
		\vspace{-1em}
	\end{subfigure}
	\caption{Reconstruction results of our E-SAI method using events from iniVation DVXplorer camera. The reference images in the first row are captured by an iPhone 11 Pro without performing spatial matching.}
    \label{fig:DVXplorer}
\end{figure*}
\begin{figure*}[th!]
	\centering
	\begin{subfigure}[b]{0.135\linewidth}
		\includegraphics[width=\linewidth, trim={0cm 0 0cm 0}]{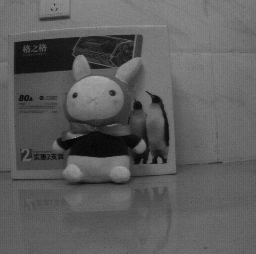}
		\vspace{-.8em}\\
		\includegraphics[width=\linewidth, trim={0cm 0 0cm 0}]{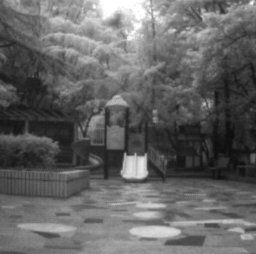}
		\vspace{-1.2em}
		\subcaption*{\scriptsize Reference}
	\end{subfigure}
	\begin{subfigure}[b]{0.135\linewidth}
		\includegraphics[width=\linewidth]{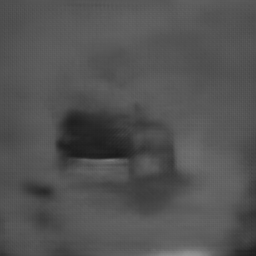}
		\vspace{-.8em}\\
		\includegraphics[width=\linewidth, trim={0cm 0 0cm 0}]{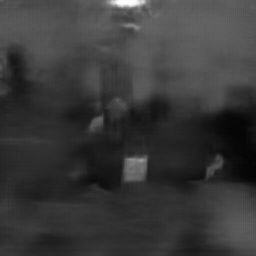}
		\vspace{-1.2em}
		\subcaption*{\scriptsize 0.1 s}
	\end{subfigure}
	\begin{subfigure}[b]{0.135\linewidth}
		\includegraphics[width=\linewidth, trim={0cm 0 0cm 0}]{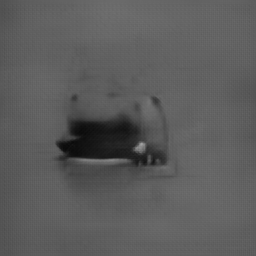}
		\vspace{-.8em}\\
		\includegraphics[width=\linewidth, trim={0cm 0 0cm 0}]{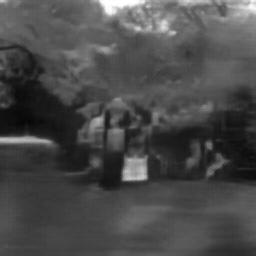}
		\vspace{-1.2em}
		\subcaption*{\scriptsize 0.2 s}
	\end{subfigure}
	\begin{subfigure}[b]{0.135\linewidth}
		\includegraphics[width=\linewidth, trim={0cm 0 0cm 0}]{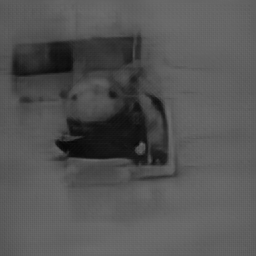}
		\vspace{-.8em}\\
		\includegraphics[width=\linewidth, trim={0cm 0 0cm 0}]{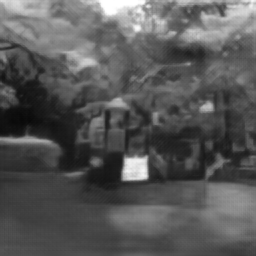}
		\vspace{-1.2em}
		\subcaption*{\scriptsize 0.3 s}
	\end{subfigure}
	\begin{subfigure}[b]{0.135\linewidth}
		\includegraphics[width=\linewidth, trim={0cm 0 0cm 0}]{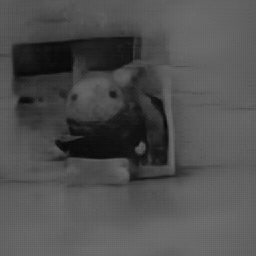}
		\vspace{-.8em}\\
		\includegraphics[width=\linewidth, trim={0cm 0 0cm 0}]{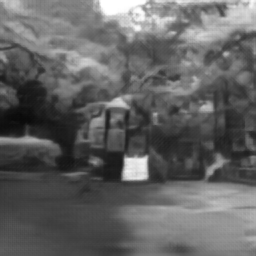}
		\vspace{-1.2em}
		\subcaption*{\scriptsize 0.4 s}
	\end{subfigure}
	\begin{subfigure}[b]{0.135\linewidth}
		\includegraphics[width=\linewidth, trim={0cm 0 0cm 0}]{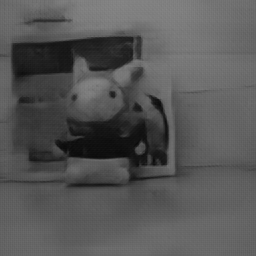}
		\vspace{-.8em}\\
		\includegraphics[width=\linewidth, trim={0cm 0 0cm 0}]{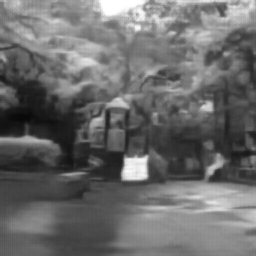}
		\vspace{-1.2em}
		\subcaption*{\scriptsize 0.5 s}
	\end{subfigure}
	\begin{subfigure}[b]{0.135\linewidth}
		\includegraphics[width=\linewidth, trim={0cm 0 0cm 0}]{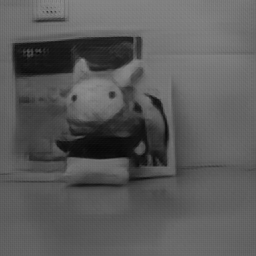}
		\vspace{-.8em}\\
		\includegraphics[width=\linewidth, trim={0cm 0 0cm 0}]{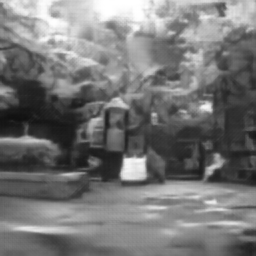}
		\vspace{-1.2em}
		\subcaption*{\scriptsize 0.7 s}
	\end{subfigure}
	\caption{
	{\color{\seccolored} Examples of reconstruction results with different durations of input event sequences. } } 
	\label{fig:DiffTime}
\end{figure*}
\section{Additional Experimental Results}
\subsection{Ablation Studies of the Refocus-Net}
We conduct experiments to verify the pure CNN architecture of our Refocus-Net. To achieve this end, we design a hybrid counterpart of our Refocus-Net by replacing the first three convolutional layers with spiking layers. As illustrated in Tab.~\ref{tab:HybridRefocus}, the hybrid Refocus-Net produces inferior results on both indoor and outdoor datasets. The SNN encoder in our reconstruction module is effective since it can filter out the noise events scattered by the refocusing procedure. However, the inputs of our Refocus-Net are unfocused events, where both signal and noise events are spatially scattered due to multi-view measurements. Thus the unaligned signal events will also be treated as noise and filtered by the SNN encoder, indicating that the hybrid SNN-CNN architecture is unsuitable for Refocus-Net.

\subsection{Generalization Performance}
To test the generalization performance of our E-SAI method, we conduct experiments with events collected by an iniVation DVXplorer camera, which differs from our DAVIS346 camera in many aspects, including spatial resolution, temporal resolution, max throughput, and contrast sensitivity. In addition, unlike DAVIS346 which simultaneously outputs frames and events, DVXplorer only outputs events, and thus we capture the occlusion-free image with an iPhone 11 Pro for reference. As shown in Fig.~\ref{fig:DVXplorer}, our method successfully restores the occluded scenes, meaning that our model generalizes well to other event cameras. 

\subsection{Performance with Different Numbers of Events} 
To investigate how much event data is sufficient for the successful reconstruction of our E-SAI method, we divide event streams into different sequences with the duration varying from 0.1 to 0.7 seconds, where each sequence is centered at the timestamp of the reference viewpoint. As shown in Tab.~\ref{tab:DiffTime}, the reconstruction results are not satisfied under short event sequences, \eg, 0.1-0.3 seconds, since the collected events only contain partial information about the occluded scenes. When the duration increases, more details can be restored using adequate signal information in the acquired events, and our method can reconstruct satisfactory results with event sequences lasting more than 0.5 seconds, as depicted in Fig.~\ref{fig:DiffTime}.

\section{Details of SAI Dataset}
For clarity, we further split the indoor SAI dataset into three categories (object, portrait, and picture) according to the type of occluded targets, and tabulate the dataset overview in Tab.~\ref{tab:dataset-overview}.
In our SAI dataset, each pair of data contains a 0.7-second event stream recorded by a DAVIS346 camera under different scenes occluded by the dense wooden fence, 30 occluded APS frames captured simultaneously with the event streams, an occlusion-free APS frame as ground truth image, and other information including target depth and camera movement speed.
The detailed data format is presented in Tab.~\ref{tab:dataset-format}.  For each pair of data, the detailed information including the number of events and the target depth is summarized in Tabs.~\ref{tab:dataset-object-train}-\ref{tab:dataset-outdoor-test}.

\begin{table}[!htb]
    \centering
    \renewcommand{\arraystretch}{1.3}
    \caption{
    {\color{\seccolored}
    Quantitative results with different durations of event sequences over all test sequences of our SAI dataset.
    }
    }
    \vspace{-1em}
  \begin{tabular}{c|c|c}
\hline
\multirow{2}{*}{\textbf{Duration}} & \multicolumn{1}{c|}{\textbf{Indoor}}                                    & \multicolumn{1}{c}{\textbf{Outdoor}}                                    \\ \cline{2-3} 
                                        & PSNR / SSIM / LPIPS          & PSNR / SSIM / LPIPS            
                                        \\ \hline
0.1 s                                    & 16.34                     / 0.5411                    / 0.3052          & 11.14                     / 0.2800                    / 0.3327          \\
0.2 s                                    & 19.63                     / 0.6705                    / 0.2416          & 13.74                     / 0.4345                    / 0.2396          \\
0.3 s                                    & 21.17                     / 0.7202                    / 0.1683          & 15.45                     / 0.5317                    / 0.1862          \\
0.4 s                                    & 23.01                     / 0.7578                    / 0.1046          & 16.82                     / 0.5902                    / 0.1555          \\
0.5 s                                    & 27.79                     / 0.8039                    / 0.0549          & 17.87                     / 0.6337                    / 0.1330          \\
0.6 s                                    & 29.92                     / 0.8212                    / 0.0432          & 18.66                     / 0.6658                    / 0.1176          \\
0.7 s                                    & \textbf{30.71}            / \textbf{0.8311}           / \textbf{0.0374} & \textbf{20.39}            / \textbf{0.7037}           / \textbf{0.0981} \\ \hline
\end{tabular}
\label{tab:DiffTime}
\end{table}
\begin{table}[ht]
\small
\renewcommand{\arraystretch}{1.1}
\centering
\caption{
{\color{\seccolored}
Overview of the proposed SAI dataset.
}}
\begin{tabular}{c|ccc|c|c}
\hline
\multirow{2}{*}{\textbf{Dataset}} & \multicolumn{3}{c|}{\textbf{Indoor}} & \multirow{2}{*}{\textbf{Outdoor}} & \multirow{2}{*}{\textbf{Sum}} \\ \cline{2-4}
                         & \textbf{Object} & \textbf{Portrait} & \textbf{Picture} &                          &                          \\ \hline
Train                    & 180    & 178      & 80      & 89                       & 527                      \\
Test                     & 20     & 20       & 10      & 11                       & 61                       \\ \hline
Total                    & 200    & 198      & 90      & 100                      & 588                      \\ \hline
\end{tabular}
\label{tab:dataset-overview}
\end{table}
\begin{table}[t!]
\small
\centering
\renewcommand{\arraystretch}{1.3}
\caption{
{\color{\seccolored}
Format and description of data in our SAI dataset.}
}
\begin{tabular}{l|p{4cm}}
\hline
\textbf{Attribute}  & \textbf{Description}                                                    \\ \hline
v                              & camera movement speed, equal to 17.7 cm/s                                                   \\
fx                             & camera intrinsic parameter                                            \\
size                          & frame size, equal to (260,346)                                                           \\
depth                          & distance between target plane and camera plane                                                           \\
events                        & event pixel positions, timestamps, and polarities \\
occ\_aps            & 30 occluded APS frames             \\
occ\_aps\_ts                  & timestamps of the 30 occluded APS frames                                   \\
occ\_free\_aps         & occlusion-free APS frame                                                \\
occ\_free\_aps\_ts            & timestamp of the occlusion-free APS frame                               \\ \hline
\end{tabular}
\label{tab:dataset-format}
\end{table}

\begin{table*}[ht]
\small
\centering
\renewcommand{\arraystretch}{1.2}
\caption{
{\color{\seccolored}
Details of the training set of our indoor object SAI dataset. \#Event indicates the number of events, Frame T represents the duration of the occluded APS sequences, and Depth is the target depth. }}
\begin{tabular}{l|ccccccccccccccc} 
\hline \hline
\textbf{ID}          & 0    & 1    & 2    & 3    & 4    & 5    & 6    & 7    & 8    & 9   & 10   & 11   & 12   & 13   & 14   \\ \hline
\textbf{\#Event (K)} & 1655 & 1648 & 1720 & 1691 & 1712 & 1714 & 1706 & 1716 & 1709 & 1717 & 1708 & 1704 & 1699 & 1694 & 1713 \\
\textbf{Frame T (s)} & 0.82 & 0.82 & 0.82 & 0.82 & 0.82 & 0.82 & 0.82 & 0.82 & 0.82 & 0.82 & 0.82 & 0.82 & 0.82 & 0.82 & 0.82  \\
\textbf{Depth (m)}   & 0.6  & 0.6  & 0.6  & 0.6  & 0.6  & 0.6  & 0.6  & 0.6  & 0.6  & 0.6  & 0.6  & 0.6  & 0.6  & 0.6  & 0.6  \\
\hline \hline
\textbf{ID}  & 15   & 16   & 17   & 18   & 19   & 20   & 21   & 22   & 23   & 24   & 25   & 26   & 27   & 28   & 29 \\ \hline
\textbf{\#Event (K)} & 1716 & 1705 & 1704 & 1682 & 1686 & 1694 & 1691 & 1680 & 1672 & 1688 & 1687 & 1702 & 1704 & 1674 & 1673 \\
\textbf{Frame T (s)} & 0.82 & 0.82 & 0.82 & 0.82 & 0.82 & 0.82 & 0.82 & 0.82 & 0.82 & 0.82 & 0.82 & 0.82 & 0.82 & 0.82 & 0.82  \\
\textbf{Depth (m)}   & 0.6  & 0.6  & 0.6  & 0.6  & 0.6  & 0.6  & 0.6  & 0.6  & 0.6  & 0.6  & 0.6  & 0.6  & 0.6  & 0.6  & 0.6  \\
\hline \hline
\textbf{ID}  & 30   & 31   & 32   & 33   & 34   & 35   & 36   & 37   & 38   & 39   & 40   & 41   & 42   & 43   & 44 \\ \hline
\textbf{\#Event (K)} & 1668 & 1671 & 1628 & 1639 & 1690 & 1686 & 1690 & 1663 & 1680 & 1666 & 1659 & 1657 & 1675 & 1684 & 1692 \\
\textbf{Frame T (s)} & 0.82 & 0.82 & 0.82 & 0.82 & 0.82 & 0.82 & 0.82 & 0.82 & 0.82 & 0.82 & 0.82 & 0.82 & 0.82 & 0.82 & 0.82  \\
\textbf{Depth (m)}   & 0.6  & 0.6  & 0.6  & 0.6  & 0.6  & 0.6  & 0.6  & 0.6  & 0.6  & 0.6  & 0.6  & 0.6  & 0.6  & 0.6  & 0.6  \\
\hline \hline
\textbf{ID}  & 45   & 46   & 47   & 48   & 49   & 50   & 51   & 52   & 53   & 54   & 55   & 56   & 57   & 58   & 59   \\ \hline
\textbf{\#Event (K)} & 1685 & 1678 & 1693 & 1666 & 1639 & 1671 & 1616 & 1651 & 1568 & 1648 & 1640 & 1661 & 1640 & 1597 & 1627 \\
\textbf{Frame T (s)} & 0.82 & 0.82 & 0.82 & 0.82 & 0.82 & 0.82 & 0.82 & 0.82 & 0.82 & 0.82 & 0.82 & 0.82 & 0.82 & 0.82 & 0.82  \\
\textbf{Depth (m)}   & 0.6  & 0.6  & 0.6  & 0.6  & 0.6  & 0.6  & 0.6  & 0.6  & 0.6  & 0.6  & 0.6  & 0.6  & 0.6  & 0.6  & 0.6  \\
\hline \hline
\textbf{ID}  & 60   & 61   & 62   & 63   & 64   & 65   & 66   & 67   & 68   & 69   & 70   & 71   & 72   & 73   & 74   \\ \hline
\textbf{\#Event (K)} & 1642 & 1617 & 1651 & 1642 & 1680 & 1685 & 1690 & 1695 & 1659 & 1654 & 1680 & 1694 & 1652 & 1621 & 1664 \\
\textbf{Frame T (s)} & 0.82 & 0.82 & 0.82 & 0.82 & 0.82 & 0.82 & 0.82 & 0.82 & 0.82 & 0.82 & 0.82 & 0.82 & 0.82 & 0.82 & 0.82  \\
\textbf{Depth (m)}   & 0.6  & 0.6  & 0.6  & 0.6  & 0.6  & 0.6  & 0.6  & 0.6  & 0.6  & 0.6  & 0.6  & 0.6  & 0.6  & 0.6  & 0.6  \\
\hline \hline
\textbf{ID}  & 75   & 76   & 77   & 78   & 79   & 80   & 81   & 82   & 83   & 84   & 85   & 86   & 87   & 88   & 89    \\ \hline
\textbf{\#Event (K)} & 1668 & 1652 & 1645 & 1665 & 1676 & 1674 & 1682 & 1661 & 1641 & 1655 & 1644 & 1662 & 1675 & 1675 & 1677 \\
\textbf{Frame T (s)} & 0.82 & 0.82 & 0.82 & 0.82 & 0.82 & 0.82 & 0.82 & 0.82 & 0.82 & 0.82 & 0.82 & 0.82 & 0.82 & 0.82 & 0.82  \\
\textbf{Depth (m)}   & 0.6  & 0.6  & 0.6  & 0.6  & 0.6  & 0.6  & 0.6  & 0.6  & 0.6  & 0.6  & 0.6  & 0.6  & 0.6  & 0.6  & 0.6  \\
\hline \hline
\textbf{ID}  & 90   & 91   & 92   & 93   & 94   & 95   & 96   & 97   & 98   & 99   & 100  & 101  & 102  & 103  & 104    \\ \hline
\textbf{\#Event (K)} & 1649 & 1646 & 1653 & 1654 & 1655 & 1659 & 1598 & 1371 & 1647 & 1351 & 1628 & 1590 & 1623 & 1583 & 1655 \\
\textbf{Frame T (s)} & 0.82 & 0.82 & 0.82 & 0.82 & 0.82 & 0.82 & 0.82 & 0.82 & 0.82 & 0.82 & 0.82 & 0.82 & 0.82 & 0.82 & 0.82  \\
\textbf{Depth (m)}   & 0.6  & 0.6  & 0.6  & 0.6  & 0.6  & 0.6  & 0.6  & 0.6  & 0.6  & 0.6  & 0.6  & 0.6  & 0.6  & 0.6  & 0.6  \\
\hline \hline
\textbf{ID}  & 105  & 106  & 107  & 108  & 109  & 110  & 111  & 112  & 113  & 114  & 115  & 116  & 117  & 118  & 119 \\ \hline
\textbf{\#Event (K)} & 1643 & 1686 & 1700 & 1643 & 1638 & 1595 & 1672 & 1656 & 1650 & 1670 & 1679 & 1674 & 1691 & 1685 & 1684 \\
\textbf{Frame T (s)} & 0.82 & 0.82 & 0.82 & 0.82 & 0.82 & 0.82 & 0.82 & 0.82 & 0.82 & 0.82 & 0.82 & 0.82 & 0.82 & 0.82 & 0.82  \\
\textbf{Depth (m)}   & 0.6  & 0.6  & 0.6  & 0.6  & 0.6  & 0.6  & 0.6  & 0.6  & 0.6  & 0.6  & 0.6  & 0.6  & 0.6  & 0.6  & 0.6  \\
\hline \hline
\textbf{ID}  & 120  & 121  & 122  & 123  & 124  & 125  & 126  & 127  & 128  & 129  & 130  & 131  & 132  & 133  & 134 \\ \hline
\textbf{\#Event (K)} & 1674 & 1678 & 1670 & 1675 & 1672 & 1656 & 1674 & 1660 & 1860 & 1900 & 1883 & 1903 & 1849 & 1841 & 1851 \\
\textbf{Frame T (s)} & 0.82 & 0.82 & 0.82 & 0.82 & 0.82 & 0.82 & 0.82 & 0.82 & 0.82 & 0.82 & 0.82 & 0.82 & 0.82 & 0.82 & 0.82  \\
\textbf{Depth (m)}   & 0.6  & 0.6  & 0.6  & 0.6  & 0.6  & 0.6  & 0.6  & 0.6  & 0.6  & 0.6  & 0.6  & 0.6  & 0.6  & 0.6  & 0.6  \\
\hline \hline
\textbf{ID}  & 135  & 136  & 137  & 138  & 139  & 140  & 141  & 142  & 143  & 144  & 145  & 146  & 147  & 148  & 149 \\ \hline
\textbf{\#Event (K)} & 1837 & 1763 & 1748 & 1751 & 1744 & 1737 & 1724 & 1629 & 1607 & 1703 & 1623 & 1667 & 1537 & 1663 & 1636 \\
\textbf{Frame T (s)} & 0.82 & 0.82 & 0.82 & 0.82 & 0.82 & 0.82 & 0.82 & 0.82 & 0.82 & 0.82 & 0.82 & 0.82 & 0.82 & 0.82 & 0.82  \\
\textbf{Depth (m)}   & 0.6  & 0.6  & 0.6  & 0.6  & 0.6  & 0.6  & 0.6  & 0.6  & 0.6  & 0.6  & 0.6  & 0.6  & 0.6  & 0.6  & 0.6  \\
\hline \hline
\textbf{ID}  & 150  & 151  & 152  & 153  & 154  & 155  & 156  & 157  & 158  & 159  & 160  & 161  & 162  & 163  & 164 \\ \hline
\textbf{\#Event (K)} & 1676 & 1637 & 1681 & 1665 & 1696 & 1511 & 1651 & 1543 & 1790 & 1812 & 1826 & 1839 & 1757 & 1502 & 1813 \\
\textbf{Frame T (s)} & 0.82 & 0.82 & 0.82 & 0.82 & 0.82 & 0.82 & 0.82 & 0.82 & 0.82 & 0.82 & 0.82 & 0.82 & 0.82 & 0.82 & 0.82  \\
\textbf{Depth (m)}   & 0.6  & 0.6  & 0.6  & 0.6  & 0.6  & 0.6  & 0.6  & 0.6  & 0.6  & 0.6  & 0.6  & 0.6  & 0.6  & 0.6  & 0.6  \\
\hline \hline
\textbf{ID}  & 165  & 166  & 167  & 168  & 169  & 170  & 171  & 172  & 173  & 174  & 175  & 176  & 177  & 178  & 179 \\ \hline
\textbf{\#Event (K)}& 1752 & 1810 & 1787 & 1784 & 1764 & 1741 & 1721 & 1768 & 1734 & 1737 & 1779 & 1741 & 1585 & 1762 & 1752 \\
\textbf{Frame T (s)} & 0.82 & 0.82 & 0.82 & 0.82 & 0.82 & 0.82 & 0.82 & 0.82 & 0.82 & 0.82 & 0.82 & 0.82 & 0.82 & 0.82 & 0.82  \\
\textbf{Depth (m)}   & 0.6  & 0.6  & 0.6  & 0.6  & 0.6  & 0.6  & 0.6  & 0.6  & 0.6  & 0.6  & 0.6  & 0.6  & 0.6  & 0.6  & 0.6  \\
\hline \hline
\end{tabular}
\label{tab:dataset-object-train}
\end{table*}

\begin{table*}[ht]
\small
\centering
\renewcommand{\arraystretch}{1.2}
\caption{
{\color{\seccolored}
Details of the test set of our indoor object SAI dataset. \#Event indicates the number of events, Frame T represents the duration of the occluded APS sequences, and Depth is the target depth.}}
\begin{tabular}{l|ccccccccccccccc}
\hline \hline
\textbf{ID}          &0    & 1    & 2    & 3    & 4    & 5    & 6    & 7    & 8    & 9    & 10   & 11   & 12   & 13   & 14   \\ \hline
\textbf{\#Event (K)} &1592 & 1576 & 1683 & 1686 & 1675 & 1676 & 1652 & 1609 & 1655 & 1627 & 1645 & 1681 & 1640 & 1613 & 1666 \\
\textbf{Frame T (s)} &0.82 & 0.82 & 0.82 & 0.82 & 0.82 & 0.82 & 0.82 & 0.82 & 0.82 & 0.82 & 0.83 & 0.83 & 0.82 & 0.82 & 0.82 \\
\textbf{Depth (m)}   &0.6  & 0.6  & 0.6  & 0.6  & 0.6  & 0.6  & 0.6  & 0.6  & 0.6  & 0.6  & 0.6  & 0.6  & 0.6  & 0.6  & 0.6 \\
\hline \hline
\textbf{ID}   & 15   & 16   & 17   & 18   & 19 \\ \hline
\textbf{\#Event (K)}  & 1658 & 1714 & 1649 & 1717 & 1767  \\
\textbf{Frame T (s)} & 0.82 & 0.82 & 0.82 & 0.82 & 0.82 \\
\textbf{Depth (m)}    & 0.6  & 0.6  & 0.6  & 0.6  & 0.6    \\
\hline \hline
\end{tabular}
\label{tab:dataset-object-test}
\end{table*}


\begin{table*}[ht]
\small
\centering
\renewcommand{\arraystretch}{1.2}
\caption{
{\color{\seccolored}
Details of the training set of our indoor picture SAI dataset. \#Event indicates the number of events, Frame T represents the duration of the occluded APS sequences, and Depth is the target depth. }}
\begin{tabular}{l|ccccccccccccccc}
\hline \hline
\textbf{ID}          &0    & 1    & 2    & 3    & 4    & 5    & 6    & 7    & 8    & 9    & 10   & 11   & 12   & 13   & 14    \\ \hline
\textbf{\#Event (K)} &1585 & 1606 & 1656 & 1647 & 1659 & 1643 & 1589 & 1595 & 1616 & 1620 & 1641 & 1648 & 1560 & 1597 & 1601 \\
\textbf{Frame T (s)} &0.82 & 0.82 & 0.82 & 0.82 & 0.82 & 0.82 & 0.82 & 0.82 & 0.82 & 0.82 & 0.82 & 0.82 & 0.82 & 0.82 & 0.82 \\
\textbf{Depth (m)}   &0.65 & 0.65 & 0.65 & 0.65 & 0.65 & 0.65 & 0.65 & 0.65 & 0.65 & 0.65 & 0.65 & 0.65 & 0.65 & 0.65 & 0.65 \\
\hline \hline
\textbf{ID}   & 15   & 16   & 17   & 18   & 19   & 20   & 21   & 22   & 23   & 24   & 25   & 26   & 27   & 28   & 29 \\ \hline
\textbf{\#Event (K)}  & 1615 & 1608 & 1623 & 1603 & 1603 & 1657 & 1631 & 1643 & 1609 & 1598 & 1590 & 1608 & 1605 & 1596 & 1590  \\
\textbf{Frame T (s)} & 0.82 & 0.82 & 0.82 & 0.82 & 0.82 & 0.82 & 0.82 & 0.82 & 0.82 & 0.82 & 0.82 & 0.82 & 0.82 & 0.82 & 0.82 \\
\textbf{Depth (m)}    & 0.65 & 0.65 & 0.65 & 0.65 & 0.65 & 0.65 & 0.65 & 0.65 & 0.65 & 0.65 & 0.65 & 0.65 & 0.65 & 0.65 & 0.65   \\
\hline \hline
\textbf{ID}   & 30   & 31   & 32   & 33   & 34   & 35   & 36   & 37   & 38   & 39   & 40   & 41   & 42   & 43   & 44 \\ \hline
\textbf{\#Event (K)}  & 1602 & 1589 & 1588 & 1583 & 1311 & 1288 & 1596 & 1582 & 1582 & 1574 & 1594 & 1559 & 1566 & 1452 & 1402  \\
\textbf{Frame T (s)} & 0.82 & 0.82 & 0.82 & 0.82 & 0.82 & 0.82 & 0.82 & 0.82 & 0.82 & 0.82 & 0.82 & 0.82 & 0.82 & 0.82 & 0.82 \\
\textbf{Depth (m)}    & 0.65 & 0.65 & 0.65 & 0.65 & 0.65 & 0.65 & 0.65 & 0.65 & 0.65 & 0.65 & 0.65 & 0.65 & 0.65 & 0.65 & 0.65   \\
\hline \hline
\textbf{ID}   & 45   & 46   & 47   & 48   & 49   & 50   & 51   & 52   & 53   & 54   & 55   & 56   & 57   & 58   & 59  \\ \hline
\textbf{\#Event (K)}  & 1422 & 1635 & 1638 & 1607 & 1606 & 1628 & 1621 & 1572 & 1596 & 1598 & 1587 & 1579 & 1599 & 1565 & 1584  \\
\textbf{Frame T (s)} & 0.82 & 0.82 & 0.82 & 0.82 & 0.82 & 0.82 & 0.82 & 0.82 & 0.82 & 0.82 & 0.82 & 0.82 & 0.82 & 0.82 & 0.82 \\
\textbf{Depth (m)}    & 0.65 & 0.65 & 0.65 & 0.65 & 0.65 & 0.65 & 0.65 & 0.65 & 0.65 & 0.65 & 0.65 & 0.65 & 0.65 & 0.65 & 0.65  \\
\hline \hline
\textbf{ID}   & 60   & 61   & 62   & 63   & 64   & 65   & 66   & 67   & 68   & 69   & 70   & 71   & 72   & 73   & 74   \\ \hline
\textbf{\#Event (K)}  & 1570 & 1584 & 1600 & 1606 & 1595 & 1601 & 1606 & 1594 & 1594 & 1593 & 1584 & 1588 & 1571 & 1581 & 1563  \\
\textbf{Frame T (s)} & 0.82 & 0.82 & 0.82 & 0.82 & 0.82 & 0.82 & 0.82 & 0.82 & 0.82 & 0.82 & 0.82 & 0.82 & 0.82 & 0.82 & 0.82 \\
\textbf{Depth (m)}    &  0.65 & 0.65 & 0.65 & 0.65 & 0.65 & 0.65 & 0.65 & 0.65 & 0.65 & 0.65 & 0.65 & 0.65 & 0.65 & 0.65 & 0.65  \\
\hline \hline
\textbf{ID}   & 75   & 76   & 77   & 78   & 79  \\ \hline
\textbf{\#Event (K)}  & 1575 & 1583 & 1547 & 1554 & 1544  \\
\textbf{Frame T (s)} & 0.82 & 0.82 & 0.82 & 0.82 & 0.82 \\
\textbf{Depth (m)}    & 0.65 & 0.65 & 0.65 & 0.65 & 0.65 \\
\hline \hline
\end{tabular}
\label{tab:dataset-picture-train}
\end{table*}

\begin{table*}[ht]
\small
\centering
\renewcommand{\arraystretch}{1.2}
\caption{
{\color{\seccolored}
Details of the test set of our indoor picture SAI dataset. \#Event indicates the number of events, Frame T represents the duration of the occluded APS sequences, and Depth is the target depth.}}
\begin{tabular}{l|ccccccccccccccc}
\hline \hline
\textbf{ID}          &0    & 1    & 2    & 3    & 4    & 5    & 6    & 7    & 8    & 9   \\ \hline
\textbf{\#Event (K)} &1574 & 1600 & 1645 & 1642 & 1584 & 1595 & 1572 & 1587 & 1595 & 1599 \\
\textbf{Frame T (s)} &0.82 & 0.82 & 0.82 & 0.82 & 0.82 & 0.82 & 0.82 & 0.82 & 0.82 & 0.82 \\
\textbf{Depth (m)}   &0.65 & 0.65 & 0.65 & 0.65 & 0.65 & 0.65 & 0.65 & 0.65 & 0.65 & 0.65 \\
\hline \hline
\end{tabular}
\label{tab:dataset-picture-test}
\end{table*}


\begin{table*}[ht]
\small
\centering
\renewcommand{\arraystretch}{1.2}
\caption{
{\color{\seccolored}
Details of the training set of our indoor portrait SAI dataset. \#Event indicates the number of events, Frame T represents the duration of the occluded APS sequences, and Depth is the target depth.}}
\begin{tabular}{l|ccccccccccccccc}
\hline \hline
\textbf{ID}          &0    & 1    & 2    & 3    & 4    & 5    & 6    & 7    & 8    & 9    & 10   & 11   & 12   & 13   & 14   \\ \hline
\textbf{\#Event (K)} &1573 & 1522 & 1632 & 1607 & 1568 & 1645 & 1615 & 1674 & 1623 & 1690 & 1622 & 1615 & 1622 & 1673 & 1628 \\
\textbf{Frame T (s)} &0.82 & 0.82 & 0.82 & 0.82 & 0.82 & 0.82 & 0.82 & 0.82 & 0.82 & 0.82 & 0.82 & 0.82 & 0.82 & 0.82 & 0.82 \\
\textbf{Depth (m)}   &0.8  & 0.8  & 0.8  & 0.8  & 0.8  & 0.8  & 0.8  & 0.8  & 0.8  & 0.8  & 0.8  & 0.8  & 0.8  & 0.8  & 0.8\\
\hline \hline
\textbf{ID}          & 15   & 16   & 17   & 18   & 19   & 20   & 21   & 22   & 23   & 24   & 25   & 26   & 27   & 28   & 29   \\ \hline
\textbf{\#Event (K)} &1685 & 1627 & 1680 & 1660 & 1718 & 1663 & 1718 & 1609 & 1567 & 1619 & 1678 & 1610 & 1636 & 1623 & 1654 \\
\textbf{Frame T (s)} & 0.82 & 0.82 & 0.82 & 0.82 & 0.82 & 0.82 & 0.82 & 0.82 & 0.82 & 0.82 & 0.82 & 0.82 & 0.82 & 0.82 & 0.82 \\
\textbf{Depth (m)}   &0.8  & 0.8  & 0.8  & 0.8  & 0.8  & 0.8  & 0.8  & 0.8  & 0.8  & 0.8  & 0.8  & 0.8  & 0.8  & 0.8  & 0.8\\
\hline \hline
\textbf{ID}          & 30   & 31   & 32   & 33   & 34   & 35   & 36   & 37   & 38   & 39   & 40   & 41   & 42   & 43   & 44   \\ \hline
\textbf{\#Event (K)} & 1639 & 1626 & 1638 & 1689 & 1621 & 1635 & 1645 & 1643 & 1639 & 1692 & 1314 & 1345 & 1635 & 1673 & 1534 \\
\textbf{Frame T (s)} & 0.82 & 0.82 & 0.82 & 0.82 & 0.82 & 0.82 & 0.82 & 0.82 & 0.82 & 0.82 & 0.82 & 0.82 & 0.82 & 0.82 & 0.82 \\
\textbf{Depth (m)}   &0.8  & 0.8  & 0.8  & 0.8  & 0.8  & 0.8  & 0.8  & 0.8  & 0.8  & 0.8  & 0.8  & 0.8  & 0.8  & 0.8  & 0.8\\
\hline \hline
\textbf{ID}          & 45   & 46   & 47   & 48   & 49   & 50   & 51   & 52   & 53   & 54   & 55   & 56   & 57   & 58   & 59    \\ \hline
\textbf{\#Event (K)} & 1557 & 1479 & 1533 & 1542 & 1607 & 1530 & 1548 & 1541 & 1566 & 1582 & 1645 & 1584 & 1647 & 1605 & 1642 \\
\textbf{Frame T (s)} & 0.82 & 0.82 & 0.82 & 0.82 & 0.82 & 0.82 & 0.82 & 0.82 & 0.82 & 0.82 & 0.82 & 0.82 & 0.82 & 0.82 & 0.82 \\
\textbf{Depth (m)}   &0.8  & 0.8  & 0.8  & 0.8  & 0.8  & 0.8  & 0.8  & 0.8  & 0.8  & 0.8  & 0.8  & 0.8  & 0.8  & 0.8  & 0.8\\
\hline \hline
\textbf{ID}          & 60   & 61   & 62   & 63   & 64   & 65   & 66   & 67   & 68   & 69   & 70   & 71   & 72   & 73   & 74    \\ \hline
\textbf{\#Event (K)} & 1600 & 1655 & 1522 & 1588 & 1611 & 1663 & 1612 & 1650 & 1611 & 1659 & 1498 & 1576 & 1495 & 1549 & 1489 \\
\textbf{Frame T (s)} & 0.82 & 0.82 & 0.82 & 0.82 & 0.82 & 0.82 & 0.82 & 0.82 & 0.82 & 0.82 & 0.82 & 0.82 & 0.82 & 0.82 & 0.82 \\
\textbf{Depth (m)}   &0.8  & 0.8  & 0.8  & 0.8  & 0.8  & 0.8  & 0.8  & 0.8  & 0.8  & 0.8  & 0.8  & 0.8  & 0.8  & 0.8  & 0.8\\
\hline \hline
\textbf{ID}          & 75   & 76   & 77   & 78   & 79   & 80   & 81   & 82   & 83   & 84   & 85   & 86   & 87   & 88   & 89   \\ \hline
\textbf{\#Event (K)} & 1527 & 1492 & 1584 & 1552 & 1619 & 1568 & 1627 & 1587 & 1658 & 1511 & 1592 & 1523 & 1598 & 1524 & 1463 \\
\textbf{Frame T (s)} & 0.82 & 0.82 & 0.82 & 0.82 & 0.82 & 0.82 & 0.82 & 0.82 & 0.82 & 0.82 & 0.82 & 0.82 & 0.82 & 0.82 & 0.82 \\
\textbf{Depth (m)}   &0.8  & 0.8  & 0.8  & 0.8  & 0.8  & 0.8  & 0.8  & 0.8  & 0.8  & 0.8  & 0.8  & 0.8  & 0.8  & 0.8  & 0.8\\
\hline \hline
\textbf{ID}          & 90   & 91   & 92   & 93   & 94   & 95   & 96   & 97   & 98   & 99   & 100  & 101  & 102  & 103  & 104  \\ \hline
\textbf{\#Event (K)} & 1591 & 1579 & 1598 & 1657 & 1594 & 1651 & 1583 & 1649 & 1605 & 1648 & 1369 & 1462 & 1605 & 1648 & 1547\\
\textbf{Frame T (s)} & 0.82 & 0.82 & 0.82 & 0.82 & 0.82 & 0.82 & 0.82 & 0.82 & 0.82 & 0.82 & 0.82 & 0.82 & 0.82 & 0.82 & 0.82 \\
\textbf{Depth (m)}   &0.8  & 0.8  & 0.8  & 0.8  & 0.8  & 0.8  & 0.8  & 0.8  & 0.8  & 0.8  & 0.8  & 0.8  & 0.8  & 0.8  & 0.8\\
\hline \hline
\textbf{ID}          & 105  & 106  & 107  & 108  & 109  & 110  & 111  & 112  & 113  & 114  & 115  & 116  & 117  & 118  & 119 \\ \hline
\textbf{\#Event (K)} & 1464 & 1572 & 1584 & 1537 & 1512 & 1553 & 1555 & 1530 & 1291 & 1540 & 1589 & 1585 & 1586 & 1570 & 1578 \\
\textbf{Frame T (s)} & 0.82 & 0.82 & 0.82 & 0.82 & 0.82 & 0.82 & 0.82 & 0.82 & 0.82 & 0.82 & 0.82 & 0.82 & 0.82 & 0.82 & 0.82 \\
\textbf{Depth (m)}   &0.8  & 0.8  & 0.8  & 0.8  & 0.8  & 0.8  & 0.8  & 0.8  & 0.8  & 0.8  & 0.8  & 0.8  & 0.8  & 0.8  & 0.8\\
\hline \hline
\textbf{ID}          & 120  & 121  & 122  & 123  & 124  & 125  & 126  & 127  & 128  & 129  & 130  & 131  & 132  & 133  & 134 \\ \hline
\textbf{\#Event (K)} & 1608 & 1661 & 1592 & 1590 & 1591 & 1580 & 1600 & 1581 & 1599 & 1582 & 1506 & 1446 & 1459 & 1503 & 1546\\
\textbf{Frame T (s)} & 0.82 & 0.82 & 0.82 & 0.82 & 0.82 & 0.82 & 0.82 & 0.82 & 0.82 & 0.82 & 0.82 & 0.82 & 0.82 & 0.82 & 0.82 \\
\textbf{Depth (m)}   &0.8  & 0.8  & 0.8  & 0.8  & 0.8  & 0.8  & 0.8  & 0.8  & 0.8  & 0.8  & 0.8  & 0.8  & 0.8  & 0.8  & 0.8\\
\hline \hline
\textbf{ID}          & 135  & 136  & 137  & 138  & 139  & 140  & 141  & 142  & 143  & 144  & 145  & 146  & 147  & 148  & 149 \\ \hline
\textbf{\#Event (K)} & 1556 & 1543 & 1544 & 1568 & 1562 & 1484 & 1338 & 1530 & 1539 & 1615 & 1477 & 1607 & 1596 & 1651 & 1602\\
\textbf{Frame T (s)} & 0.82 & 0.82 & 0.82 & 0.82 & 0.82 & 0.82 & 0.82 & 0.82 & 0.82 & 0.82 & 0.82 & 0.82 & 0.82 & 0.82 & 0.82 \\
\textbf{Depth (m)}   &0.8  & 0.8  & 0.8  & 0.8  & 0.8  & 0.8  & 0.8  & 0.8  & 0.8  & 0.8  & 0.8  & 0.8  & 0.8  & 0.8  & 0.8\\
\hline \hline
\textbf{ID}          & 150  & 151  & 152  & 153  & 154  & 155  & 156  & 157  & 158  & 159  & 160  & 161  & 162  & 163  & 164 \\ \hline
\textbf{\#Event (K)} & 1654 & 1613 & 1583 & 1596 & 1592 & 1603 & 1549 & 1557 & 1541 & 1561 & 1604 & 1613 & 1524 & 1419 & 1590\\
\textbf{Frame T (s)} & 0.82 & 0.82 & 0.82 & 0.82 & 0.82 & 0.82 & 0.82 & 0.82 & 0.82 & 0.82 & 0.82 & 0.82 & 0.82 & 0.82 & 0.82 \\
\textbf{Depth (m)}   &0.8  & 0.8  & 0.8  & 0.8  & 0.8  & 0.8  & 0.8  & 0.8  & 0.8  & 0.8  & 0.8  & 0.8  & 0.8  & 0.8  & 0.8\\
\hline \hline
\textbf{ID}          &165  & 166  & 167  & 168  & 169  & 170  & 171  & 172  & 173  & 174  & 175  & 176  & 177 \\ \hline
\textbf{\#Event (K)} & 1583 & 1595 & 1602 & 1593 & 1607 & 1600 & 1546 & 1596 & 1594 & 1589 & 1596 & 1601 & 1602\\
\textbf{Frame T (s)} & 0.82 & 0.82 & 0.82 & 0.82 & 0.82 & 0.82 & 0.82 & 0.82 & 0.82 & 0.82 & 0.82 & 0.82 & 0.82  \\
\textbf{Depth (m)}   &0.8  & 0.8  & 0.8  & 0.8  & 0.8  & 0.8  & 0.8  & 0.8  & 0.8  & 0.8  & 0.8  & 0.8  & 0.8 \\
\hline \hline
\end{tabular}
\label{tab:dataset-portrait-train}
\end{table*}

\begin{table*}[ht]
\small
\centering
\renewcommand{\arraystretch}{1.2}
\caption{
{\color{\seccolored}
Details of the test set of our indoor portrait SAI dataset. \#Event indicates the number of events, Frame T represents the duration of the occluded APS sequences, and Depth is the target depth.}}
\begin{tabular}{l|ccccccccccccccc}
\hline \hline
\textbf{ID}          &0    & 1    & 2    & 3    & 4    & 5    & 6    & 7    & 8    & 9    & 10   & 11   & 12   & 13   & 14   \\ \hline
\textbf{\#Event (K)} &2001 & 2063 & 1629 & 1672 & 1645 & 1697 & 1601 & 1655 & 1512 & 1551 & 1523 & 1582 & 1570 & 1625 & 1598 \\
\textbf{Frame T (s)} &0.73 & 0.73 & 0.82 & 0.82 & 0.82 & 0.82 & 0.82 & 0.82 & 0.82 & 0.82 & 0.82 & 0.82 & 0.82 & 0.82 & 0.82 \\
\textbf{Depth (m)}   &0.8  & 0.8  & 0.8  & 0.8  & 0.8  & 0.8  & 0.8  & 0.8  & 0.8  & 0.8  & 0.8  & 0.8  & 0.8  & 0.8  & 0.8  \\
\hline \hline
\textbf{ID}   & 15   & 16   & 17   & 18   & 19 \\ \hline
\textbf{\#Event (K)}  & 1501 & 1558 & 1547 & 1597 & 1619  \\
\textbf{Frame T (s)} & 0.82 & 0.82 & 0.82 & 0.82 & 0.82 \\
\textbf{Depth (m)}    & 0.8  & 0.8  & 0.8  & 0.8  & 0.8    \\
\hline \hline
\end{tabular}
\label{tab:dataset-portrait-test}
\end{table*}


\begin{table*}[ht]
\small
\centering
\renewcommand{\arraystretch}{1.2}
\caption{
{\color{\seccolored}
Details of the training set of our outdoor SAI dataset. \#Event indicates the number of events, Frame T represents the duration of the occluded APS sequences, and Depth is the target depth. The max target depth is set to 50 m.}}
\begin{tabular}{l|ccccccccccccccc}
\hline \hline
\textbf{ID}          &0    & 1    & 2    & 3    & 4    & 5    & 6    & 7    & 8    & 9    & 10   & 11   & 12   & 13   & 14  \\ \hline
\textbf{\#Event (K)} &1763 & 1235 & 1518 & 1785 & 1957 & 1705 & 2013 & 1878 & 933  & 992  & 1431 & 1395 & 1838 & 1610 & 1848 \\
\textbf{Frame T (s)} &0.73 & 0.73 & 0.73 & 0.73 & 0.73 & 0.73 & 0.73 & 0.73 & 0.73 & 0.72 & 0.72 & 0.73 & 0.73 & 0.73 & 0.73 \\
\textbf{Depth (m)}   &9.3  & 2.9  & 4.3  & 12.3 & 3.6  & 15.5 & 26.5 & 16.1 & 50   & 50   & 50   & 7    & 50   & 30.4 & 11.7 \\
\hline \hline
\textbf{ID}   & 15   & 16   & 17   & 18   & 19   & 20   & 21   & 22   & 23   & 24   & 25   & 26   & 27   & 28   & 29\\ \hline
\textbf{\#Event (K)}  & 1866 & 1534 & 1565 & 1385 & 1670 & 930  & 1677 & 1819 & 788  & 1730 & 260  & 1525 & 1728 & 1209 & 1705 \\
\textbf{Frame T (s)} & 0.72 & 0.73 & 0.73 & 0.73 & 0.73 & 0.73 & 0.73 & 0.73 & 0.73 & 0.73 & 0.73 & 0.72 & 0.73 & 0.73 & 0.73 \\
\textbf{Depth (m)}    & 3.9  & 4.3  & 50   & 27.9 & 28.5 & 6.9  & 8    & 6    & 5.3  & 12   & 14.4 & 8.9  & 50   & 10.2 & 5     \\
\hline \hline
\textbf{ID}   & 30   & 31   & 32    & 33   & 34   & 35    & 36   & 37   & 38   & 39   & 40   & 41   & 42   & 43   & 44 \\ \hline
\textbf{\#Event (K)}  & 1928 & 1783 & 1519  & 1448 & 1765 & 1564  & 1452 & 1797 & 1827 & 1307 & 1745 & 1684 & 1764 & 1915 & 1794 \\
\textbf{Frame T (s)} & 0.72 & 0.72 & 1.11  & 1.12 & 1.11 & 1.11  & 1.12 & 1.11 & 1.11 & 1.11 & 1.11 & 1.11 & 1.11 & 1.11 & 1.12 \\
\textbf{Depth (m)}    & 7.6  & 10   & 19.74 & 50   & 50   & 19.74 & 50   & 50   & 12.1 & 12.1 & 5    & 5    & 1.8  & 1.8  & 9       \\
\hline \hline
\textbf{ID}   & 45   & 46   & 47   & 48   & 49   & 50   & 51   & 52   & 53   & 54   & 55   & 56   & 57   & 58   & 59 \\ \hline
\textbf{\#Event (K)}  & 1771 & 1784 & 1459 & 2160 & 2083 & 500  & 894  & 493  & 1763 & 1679 & 1904 & 1825 & 1732 & 1683 & 1802 \\
\textbf{Frame T (s)} & 1.11 & 0.82 & 0.82 & 0.82 & 1.12 & 1.11 & 1.11 & 1.11 & 1.11 & 1.11 & 0.97 & 0.97 & 0.97 & 0.97 & 0.97 \\
\textbf{Depth (m)}    & 9    & 6.2  & 4.6  & 4.6  & 2.5  & 2.5  & 1.4  & 1.4  & 2.6  & 2.6  & 7.8  & 9.7  & 5.4  & 4.8  & 7.4   \\
\hline \hline
\textbf{ID}   & 60   & 61   & 62   & 63   & 64   & 65   & 66   & 67   & 68   & 69   & 70   & 71   & 72   & 73   & 74 \\ \hline
\textbf{\#Event (K)}  &1501 & 1774 & 1725 & 1797 & 1715 & 1670 & 1848 & 1911 & 1819 & 1615 & 1784 & 2023 & 1784 & 1881 & 1666 \\
\textbf{Frame T (s)} &  0.97 & 0.97 & 0.97 & 0.97 & 0.97 & 0.97 & 0.97 & 0.97 & 0.97 & 0.97 & 0.97 & 0.97 & 0.97 & 0.97 & 0.97 \\
\textbf{Depth (m)}    & 3.8  & 2.5  & 1.5  & 19.2 & 10.2 & 20   & 20   & 20   & 20   & 20   & 20   & 6.3  & 4    & 5.7  & 1.3  \\
\hline \hline
\textbf{ID}   & 75   & 76   & 77   & 78   & 79   & 80   & 81   & 82   & 83   & 84   & 85   & 86   & 87   & 88  \\ \hline
\textbf{\#Event (K)}  &2082 & 1738 & 1871 & 1737 & 2050 & 2062 & 1607 & 1977 & 1716 & 2090 & 1995 & 1650 & 1763 & 2054 \\
\textbf{Frame T (s)} &  0.97 & 1.16 & 0.97 & 0.97 & 0.97 & 0.97 & 0.97 & 0.97 & 0.97 & 0.97 & 0.97 & 0.97 & 0.98 & 0.97 \\
\textbf{Depth (m)}    & 1.8  & 1.8  & 2.5  & 1.35 & 3.7  & 2.6  & 2.6  & 2.05 & 2.05 & 2.4  & 1.9  & 1.05 & 6.5  & 1.85  \\
\hline \hline
\end{tabular}
\label{tab:dataset-outdoor-train}
\end{table*}

\begin{table*}[ht]
\small
\centering
\renewcommand{\arraystretch}{1.2}
\caption{
{\color{\seccolored}
Details of the test set of our outdoor SAI dataset. \#Event indicates the number of events, Frame T represents the duration of the occluded APS sequences, and Depth is the target depth.}}
\begin{tabular}{l|ccccccccccccccc}
\hline \hline
\textbf{ID}          &0    & 1    & 2    & 3    & 4    & 5    & 6    & 7    & 8    & 9    & 10    \\ \hline
\textbf{\#Event (K)} &1869 & 1847 & 1751 & 1743 & 1790 & 1685 & 1864 & 1722 & 1461 & 1797 & 1789 \\
\textbf{Frame T (s)} &0.72 & 0.72 & 1.11 & 1.11 & 1.11 & 1.11 & 1.11 & 0.97 & 0.97 & 0.97 & 0.97 \\
\textbf{Depth (m)}   &9.3  & 9.3  & 5    & 3    & 7.1  & 17.3 & 14.9 & 20   & 1.7  & 6.5  & 6.5  \\
\hline \hline
\end{tabular}
\label{tab:dataset-outdoor-test}
\end{table*}

\ifCLASSOPTIONcaptionsoff
  \newpage
\fi


\bibliographystyle{IEEEtran}
\bibliography{library}

\begin{thebibliography}{10}
\providecommand{\url}[1]{#1}
\csname url@samestyle\endcsname
\providecommand{\newblock}{\relax}
\providecommand{\bibinfo}[2]{#2}
\providecommand{\BIBentrySTDinterwordspacing}{\spaceskip=0pt\relax}
\providecommand{\BIBentryALTinterwordstretchfactor}{4}
\providecommand{\BIBentryALTinterwordspacing}{\spaceskip=\fontdimen2\font plus
\BIBentryALTinterwordstretchfactor\fontdimen3\font minus
  \fontdimen4\font\relax}
\providecommand{\BIBforeignlanguage}[2]{{%
\expandafter\ifx\csname l@#1\endcsname\relax
\typeout{** WARNING: IEEEtran.bst: No hyphenation pattern has been}%
\typeout{** loaded for the language `#1'. Using the pattern for}%
\typeout{** the default language instead.}%
\else
\language=\csname l@#1\endcsname
\fi
#2}}
\providecommand{\BIBdecl}{\relax}
\BIBdecl

\bibitem{vaishUsingPlaneParallax2004}
V.~Vaish, B.~Wilburn, N.~Joshi, and M.~Levoy, ``\BIBforeignlanguage{en}{Using
  plane + parallax for calibrating dense camera arrays},'' in
  \emph{\BIBforeignlanguage{en}{IEEE Conf. Comput. Vis. Pattern Recog.}},
  vol.~1, 2004, pp. 2--9.

\bibitem{vaishReconstructingOccludedSurfaces2006}
V.~Vaish, M.~Levoy, R.~Szeliski, C.~L. Zitnick, and S.~B. Kang,
  ``\BIBforeignlanguage{en}{Reconstructing occluded surfaces using synthetic
  apertures: Stereo, focus and robust measures},'' in
  \emph{\BIBforeignlanguage{en}{IEEE Conf. Comput. Vis. Pattern Recog.}},
  vol.~2, 2006, pp. 2331--2338.

\bibitem{peiSyntheticApertureImaging2013}
Z.~Pei, Y.~Zhang, X.~Chen, and Y.-H. Yang, ``\BIBforeignlanguage{en}{Synthetic
  aperture imaging using pixel labeling via energy minimization},''
  \emph{\BIBforeignlanguage{en}{Pattern Recognition}}, vol.~46, no.~1, pp.
  174--187, 2013.

\bibitem{xiaoSeeingForegroundOcclusion2017}
Z.~Xiao, L.~Si, and G.~Zhou, ``\BIBforeignlanguage{en}{Seeing beyond foreground
  occlusion: A joint framework for {SAP}-based scene depth and appearance
  reconstruction},'' \emph{\BIBforeignlanguage{en}{IEEE J. Selected Topics in
  Signal Processing}}, vol.~11, no.~7, pp. 979--991, 2017.

\bibitem{wangDeOccNetLearningSee2019}
Y.~Wang, T.~Wu, J.~Yang, L.~Wang, W.~An, and Y.~Guo,
  ``\BIBforeignlanguage{en}{{DeOccNet}: Learning to see through foreground
  occlusions in light fields},'' in \emph{\BIBforeignlanguage{en}{IEEE Conf.
  Wint. Applic. Comput. Vis.}}, 2020, pp. 118--127.

\bibitem{lichtsteiner128Times1282008}
P.~Lichtsteiner, C.~Posch, and T.~Delbruck, ``\BIBforeignlanguage{en}{A
  128$\times$128 120 {dB} 15 $\mu$s latency asynchronous temporal contrast
  vision sensor},'' \emph{\BIBforeignlanguage{en}{IEEE J. Solid-State
  Circuits}}, vol.~43, no.~2, pp. 566--576, 2008.

\bibitem{9138762}
G.~Gallego, T.~Delbruck, G.~M. Orchard, C.~Bartolozzi, B.~Taba, A.~Censi,
  S.~Leutenegger, A.~Davison, J.~Conradt, K.~Daniilidis, and D.~Scaramuzza,
  ``Event-based vision: {A} survey,'' \emph{IEEE Trans. Pattern Anal. Mach.
  Intell.}, pp. 1--1, 2020.

\bibitem{cohen2019}
\BIBentryALTinterwordspacing
G.~Cohen, ``Active sensing and its application to neuromorphic space imaging
  [talk],'' ICONS 2019. [Online]. Available: \url{https://youtu.be/mnfQwngwW78}
\BIBentrySTDinterwordspacing

\bibitem{maass1998pulsed}
W.~Maass and C.~Bishop, \emph{Pulsed neural networks}.\hskip 1em plus 0.5em
  minus 0.4em\relax MIT Press, 1998.

\bibitem{maassNetworksSpikingNeurons1997}
W.~Maass, ``\BIBforeignlanguage{English}{Networks of spiking neurons: The third
  generation of neural network models},''
  \emph{\BIBforeignlanguage{English}{Neural Networks}}, vol.~10, no.~9, pp.
  1659--1671, 1997.

\bibitem{lee2020spike}
C.~Lee, A.~K. Kosta, A.~Z. Zhu, K.~Chaney, K.~Daniilidis, and K.~Roy,
  ``Spike-flownet: event-based optical flow estimation with energy-efficient
  hybrid neural networks,'' in \emph{Eur. Conf. Comput. Vis.}\hskip 1em plus
  0.5em minus 0.4em\relax Springer, 2020, pp. 366--382.

\bibitem{pandaScalableEfficientAccurate2020}
P.~Panda, S.~A. Aketi, and K.~Roy, ``\BIBforeignlanguage{en}{Toward scalable,
  efficient, and accurate deep spiking neural networks with backward residual
  connections, stochastic softmax, and hybridization},''
  \emph{\BIBforeignlanguage{en}{Frontiers in Neuroscience}}, vol.~14, p. 653,
  2020.

\bibitem{zhang2021event}
X.~Zhang, W.~Liao, L.~Yu, W.~Yang, and G.-S. Xia, ``Event-based synthetic
  aperture imaging with a hybrid network,'' in \emph{IEEE Conf. Comput. Vis.
  Pattern Recog.}, 2021.

\bibitem{gehrig2020eklt}
D.~Gehrig, H.~Rebecq, G.~Gallego, and D.~Scaramuzza, ``{EKLT}: Asynchronous
  photometric feature tracking using events and frames,'' \emph{Int. J. Comput.
  Vis.}, vol. 128, no.~3, pp. 601--618, 2020.

\bibitem{hagenaars2021self}
J.~Hagenaars, F.~Paredes-Vall{\'e}s, and G.~De~Croon, ``Self-supervised
  learning of event-based optical flow with spiking neural networks,''
  \emph{Adv. Neural Inform. Process. Syst.}, vol.~34, pp. 7167--7179, 2021.

\bibitem{vidal2018ultimate}
A.~R. Vidal, H.~Rebecq, T.~Horstschaefer, and D.~Scaramuzza, ``Ultimate {SLAM}?
  {Combining} events, images, and {IMU} for robust visual {SLAM} in {HDR} and
  high-speed scenarios,'' \emph{IEEE Robotics and Automation Letters}, vol.~3,
  no.~2, pp. 994--1001, 2018.

\bibitem{mostafavi2021learning}
M.~Mostafavi, L.~Wang, and K.-J. Yoon, ``Learning to reconstruct {HDR} images
  from events, with applications to depth and flow prediction,'' \emph{Int. J.
  Comput. Vis.}, vol. 129, no.~4, pp. 900--920, 2021.

\bibitem{rebecqEventsToVideoBringingModern2019}
H.~Rebecq, R.~Ranftl, V.~Koltun, and D.~Scaramuzza,
  ``\BIBforeignlanguage{en}{Events-to-video: Bringing modern computer vision to
  event cameras},'' in \emph{\BIBforeignlanguage{en}{IEEE Conf. Comput. Vis.
  Pattern Recog.}}, 2019, pp. 3852--3861.

\bibitem{rebecq2019high}
------, ``High speed and high dynamic range video with an event camera,''
  \emph{IEEE Trans. Pattern Anal. Mach. Intell.}, vol.~43, no.~6, pp.
  1964--1980, 2021.

\bibitem{wang2020event}
B.~Wang, J.~He, L.~Yu, G.-S. Xia, and W.~Yang, ``Event enhanced high-quality
  image recovery,'' in \emph{Eur. Conf. Comput. Vis.}, 2020.

\bibitem{9252186}
L.~Pan, R.~Hartley, C.~Scheerlinck, M.~Liu, X.~Yu, and Y.~Dai, ``High frame
  rate video reconstruction based on an event camera,'' \emph{IEEE Trans.
  Pattern Anal. Mach. Intell.}, 2020.

\bibitem{cook2011}
M.~Cook, L.~Gugelmann, F.~Jug, C.~Krautz, and A.~Steger, ``Interacting maps for
  fast visual interpretation,'' in \emph{The 2011 International Joint
  Conference on Neural Networks}, 2011, pp. 770--776.

\bibitem{kim2014}
H.~Kim, A.~Handa, R.~Benosman, S.-H. Ieng, and A.~Davison, ``Simultaneous
  mosaicing and tracking with an event camera,'' in \emph{Brit. Mach. Vis.
  Conf.}, 2014.

\bibitem{kim2016}
H.~Kim, S.~Leutenegger, and A.~J. Davison, ``Real-time {3D} reconstruction and
  {6-DoF} tracking with an event camera,'' in \emph{European Conference on
  Computer Vision}.\hskip 1em plus 0.5em minus 0.4em\relax Springer, 2016, pp.
  349--364.

\bibitem{bardow2016}
P.~Bardow, A.~J. Davison, and S.~Leutenegger, ``Simultaneous optical flow and
  intensity estimation from an event camera,'' in \emph{IEEE Conf. Comput. Vis.
  Pattern Recog.}, 2016, pp. 884--892.

\bibitem{scheerlinckContinuousTimeIntensityEstimation2019}
C.~Scheerlinck, N.~Barnes, and R.~Mahony, ``Continuous-time intensity
  estimation using event cameras,'' in \emph{ACCV}.\hskip 1em plus 0.5em minus
  0.4em\relax Springer, 2018, pp. 308--324.

\bibitem{munda2018}
G.~Munda, C.~Reinbacher, and T.~Pock, ``Real-time intensity-image
  reconstruction for event cameras using manifold regularisation,'' \emph{Int.
  J. Comput. Vis.}, vol. 126, no.~12, pp. 1381--1393, 2018.

\bibitem{wang2019}
L.~Wang, S.~M.~M. I, Y.-S. Ho, and K.-J. Yoon, ``Event-based high dynamic range
  image and very high frame rate video generation using conditional generative
  adversarial networks,'' in \emph{IEEE Conf. Comput. Vis. Pattern Recog.},
  2019, pp. 10\,081--10\,090.

\bibitem{zou2021}
Y.~Zou, Y.~Zheng, T.~Takatani, and Y.~Fu, ``Learning to reconstruct high speed
  and high dynamic range videos from events,'' in \emph{IEEE Conf. Comput. Vis.
  Pattern Recog.}, 2021, pp. 2024--2033.

\bibitem{4409032}
N.~Joshi, S.~Avidan, W.~Matusik, and D.~J. Kriegman, ``Synthetic aperture
  tracking: Tracking through occlusions,'' in \emph{Int. Conf. Comput. Vis.},
  2007, pp. 1--8.

\bibitem{yangContinuouslyTrackingSeethrough2011}
T.~Yang, Y.~Zhang, X.~Tong, X.~Zhang, and R.~Yu,
  ``\BIBforeignlanguage{en}{Continuously tracking and see-through occlusion
  based on a new hybrid synthetic aperture imaging model},'' in
  \emph{\BIBforeignlanguage{en}{IEEE Conf. Comput. Vis. Pattern Recog.}}, 2011,
  pp. 3409--3416.

\bibitem{7583662}
Z.~Pei, X.~Chen, and Y.-H. Yang, ``All-in-focus synthetic aperture imaging
  using image matting,'' \emph{IEEE Trans. Circuit Syst. Video Technol.},
  vol.~28, no.~2, pp. 288--301, 2018.

\bibitem{zhang2017synthetic}
X.~Zhang, Y.~Zhang, T.~Yang, and Y.-H. Yang, ``Synthetic aperture photography
  using a moving camera-imu system,'' \emph{Pattern Recognition}, vol.~62, pp.
  175--188, 2017.

\bibitem{s19030607}
Z.~Pei, Y.~Li, M.~Ma, J.~Li, C.~Leng, X.~Zhang, and Y.~Zhang, ``Occluded-object
  {3D} reconstruction using camera array synthetic aperture imaging,''
  \emph{Sensors}, vol.~19, no.~3, p. 607, 2019.

\bibitem{10.1145/1186822.1073259}
B.~Wilburn, N.~Joshi, V.~Vaish, E.-V. Talvala, E.~Antunez, A.~Barth, A.~Adams,
  M.~Horowitz, and M.~Levoy, ``High performance imaging using large camera
  arrays,'' in \emph{{ACM} {SIGGRAPH}}, 2005, pp. 765--776.

\bibitem{PEI20121637}
Z.~Pei, Y.~Zhang, T.~Yang, X.~Zhang, and Y.-H. Yang, ``A novel multi-object
  detection method in complex scene using synthetic aperture imaging,''
  \emph{Pattern Recognition}, vol.~45, no.~4, pp. 1637--1658, 2012.

\bibitem{wu2017light}
G.~Wu, B.~Masia, A.~Jarabo, Y.~Zhang, L.~Wang, Q.~Dai, T.~Chai, and Y.~Liu,
  ``Light field image processing: An overview,'' \emph{{IEEE J. Selected Topics
  in Signal Processing}}, vol.~11, no.~7, pp. 926--954, 2017.

\bibitem{hartley2003multiple}
R.~Hartley and A.~Zisserman, \emph{Multiple view geometry in computer
  vision}.\hskip 1em plus 0.5em minus 0.4em\relax Cambridge university press,
  2003.

\bibitem{rebecq2018emvs}
H.~Rebecq, G.~Gallego, E.~Mueggler, and D.~Scaramuzza, ``{EMVS: Event-based
  multi-view stereo—3D reconstruction with an event camera in real-time},''
  \emph{Int. J. Comput. Vis.}, vol. 126, no.~12, pp. 1394--1414, 2018.

\bibitem{Gallego_2018_CVPR}
G.~Gallego, H.~Rebecq, and D.~Scaramuzza, ``A unifying contrast maximization
  framework for event cameras, with applications to motion, depth, and optical
  flow estimation,'' in \emph{IEEE Conf. Comput. Vis. Pattern Recog.}, 2018,
  pp. 3867--3876.

\bibitem{gallego2019focus}
G.~Gallego, M.~Gehrig, and D.~Scaramuzza, ``Focus is all you need: Loss
  functions for event-based vision,'' in \emph{IEEE Conf. Comput. Vis. Pattern
  Recog.}, 2019, pp. 12\,280--12\,289.

\bibitem{stoffregen2019event}
T.~Stoffregen and L.~Kleeman, ``Event cameras, contrast maximization and reward
  functions: An analysis,'' in \emph{IEEE Conf. Comput. Vis. Pattern Recog.},
  2019, pp. 12\,300--12\,308.

\bibitem{nunes2020entropy}
U.~M. Nunes and Y.~Demiris, ``Entropy minimisation framework for event-based
  vision model estimation,'' in \emph{Eur. Conf. Comput. Vis.}\hskip 1em plus
  0.5em minus 0.4em\relax Springer, 2020, pp. 161--176.

\bibitem{xu2020a}
J.~Xu, M.~Jiang, L.~Yu, W.~Yang, and W.~Wang, ``Robust motion compensation for
  event cameras with smooth constraint,'' \emph{IEEE Transactions on
  Computational Imaging}, vol.~6, pp. 604--614, 2020.

\bibitem{jaderberg2015spatial}
M.~{Jaderberg}, K.~{Simonyan}, A.~{Zisserman}, and K.~{Kavukcuoglu}, ``{Spatial
  transformer networks},'' in \emph{Adv. Neural Inform. Process. Syst.},
  vol.~28, 2015, pp. 2017--2025.

\bibitem{lin2017inverse}
C.-H. {Lin} and S.~{Lucey}, ``{Inverse compositional spatial transformer
  networks},'' in \emph{IEEE Conf. Comput. Vis. Pattern Recog.}, 2017, pp.
  2252--2260.

\bibitem{wuDirectTrainingSpiking2019}
Y.~Wu, L.~Deng, G.~Li, J.~Zhu, Y.~Xie, and L.~Shi,
  ``\BIBforeignlanguage{en}{Direct training for spiking neural networks:
  Faster, larger, better},'' in \emph{\BIBforeignlanguage{en}{AAAI Conf. Artif.
  Intell.}}, vol.~33, 2019, pp. 1311--1318.

\bibitem{10.3389/fnins.2018.00331}
Y.~Wu, L.~Deng, G.~Li, J.~Zhu, and L.~Shi, ``Spatio-temporal backpropagation
  for training high-performance spiking neural networks,'' \emph{Frontiers in
  Neuroscience}, vol.~12, p. 331, 2018.

\bibitem{zhu2017unpaired}
J.-Y. Zhu, T.~Park, P.~Isola, and A.~A. Efros, ``Unpaired image-to-image
  translation using cycle-consistent adversarial networks,'' in \emph{Int.
  Conf. Comput. Vis.}, 2017, pp. 2223--2232.

\bibitem{johnson2016perceptual}
J.~Johnson, A.~Alahi, and L.~Fei-Fei, ``Perceptual losses for real-time style
  transfer and super-resolution,'' in \emph{Eur. Conf. Comput. Vis.}, 2016, pp.
  694--711.

\bibitem{mahendran2015understanding}
A.~Mahendran and A.~Vedaldi, ``Understanding deep image representations by
  inverting them,'' in \emph{IEEE Conf. Comput. Vis. Pattern Recog.}, 2015, pp.
  5188--5196.

\bibitem{wang2003multiscale}
Z.~Wang, E.~P. Simoncelli, and A.~C. Bovik, ``Multiscale structural similarity
  for image quality assessment,'' in \emph{IEEE Asilomar Conf. Sign. Syst.
  Comput.}, vol.~2, 2003, pp. 1398--1402.

\bibitem{he2015delving}
K.~He, X.~Zhang, S.~Ren, and J.~Sun, ``Delving deep into rectifiers: Surpassing
  human-level performance on imagenet classification,'' in \emph{Proceedings of
  the IEEE international conference on computer vision}, 2015, pp. 1026--1034.

\bibitem{kingma2014adam}
D.~P. Kingma and J.~Ba, ``Adam: {A} method for stochastic optimization,''
  \emph{arXiv preprint arXiv:1412.6980}, 2014.

\bibitem{loshchilov2016sgdr}
I.~{Loshchilov} and F.~{Hutter}, ``{SGDR: Stochastic gradient descent with warm
  restarts},'' in \emph{Int. Conf. Learn. Represent.}, 2017.

\bibitem{simonyan2014very}
K.~Simonyan and A.~Zisserman, ``Very deep convolutional networks for
  large-scale image recognition,'' \emph{arXiv preprint arXiv:1409.1556}, 2014.

\bibitem{russakovsky2015imagenet}
O.~Russakovsky, J.~Deng, H.~Su, J.~Krause, S.~Satheesh, S.~Ma, Z.~Huang,
  A.~Karpathy, A.~Khosla, M.~Bernstein, and {others}, ``Imagenet large scale
  visual recognition challenge,'' \emph{Int. J. Comput. Vis.}, vol. 115, no.~3,
  pp. 211--252, 2015.

\bibitem{zhang2018perceptual}
R.~Zhang, P.~Isola, A.~A. Efros, E.~Shechtman, and O.~Wang, ``The unreasonable
  effectiveness of deep features as a perceptual metric,'' in \emph{IEEE Conf.
  Comput. Vis. Pattern Recog.}, 2018, pp. 586--595.

\bibitem{liao2022synthetic}
W.~Liao, X.~Zhang, L.~Yu, S.~Lin, W.~Yang, and N.~Qiao, ``Synthetic aperture
  imaging with events and frames,'' in \emph{IEEE Conf. Comput. Vis. Pattern
  Recog.}, 2022, pp. 17\,735--17\,744.

\end{thebibliography}
%

%


\vskip -2.5\baselineskip plus -1fil

\begin{IEEEbiography}[{\includegraphics[width=1in,height=1.25in,clip, keepaspectratio, trim={0 50 0 90}]{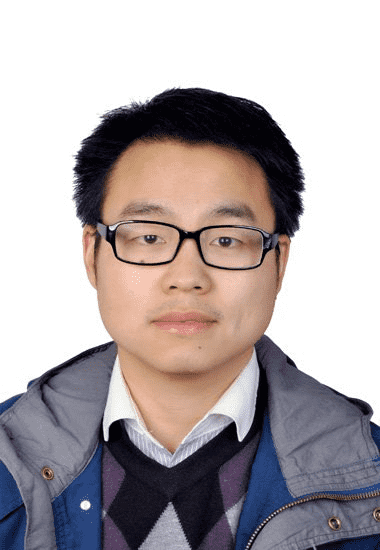}}]{Lei Yu}
received his B.S. and Ph.D. degrees in signal processing from Wuhan University, Wuhan, China, in 2006 and 2012, respectively. From 2013 to 2014, he has been a Postdoc Researcher with the VisAGeS Group at the Institut National de Recherche en Informatique et en Automatique (INRIA) for one and half years. He is currently working as an associate professor at the School of Electronics and Information, Wuhan University, Wuhan, China. From 2016 to 2017, he has also been a Visiting Professor at Duke University for one year. He has been working as a guest professor in the École Nationale Supérieure de l'Électronique et de ses Applications (ENSEA), Cergy, France, for one month in 2018. His research interests include event-based vision, neuromorphic computation, and signal processing.
\end{IEEEbiography}

\vskip -2.5\baselineskip plus -1fil

\begin{IEEEbiography}[{\includegraphics[width=1in,height=1.25in,clip,keepaspectratio, trim={20 40 20 40}]{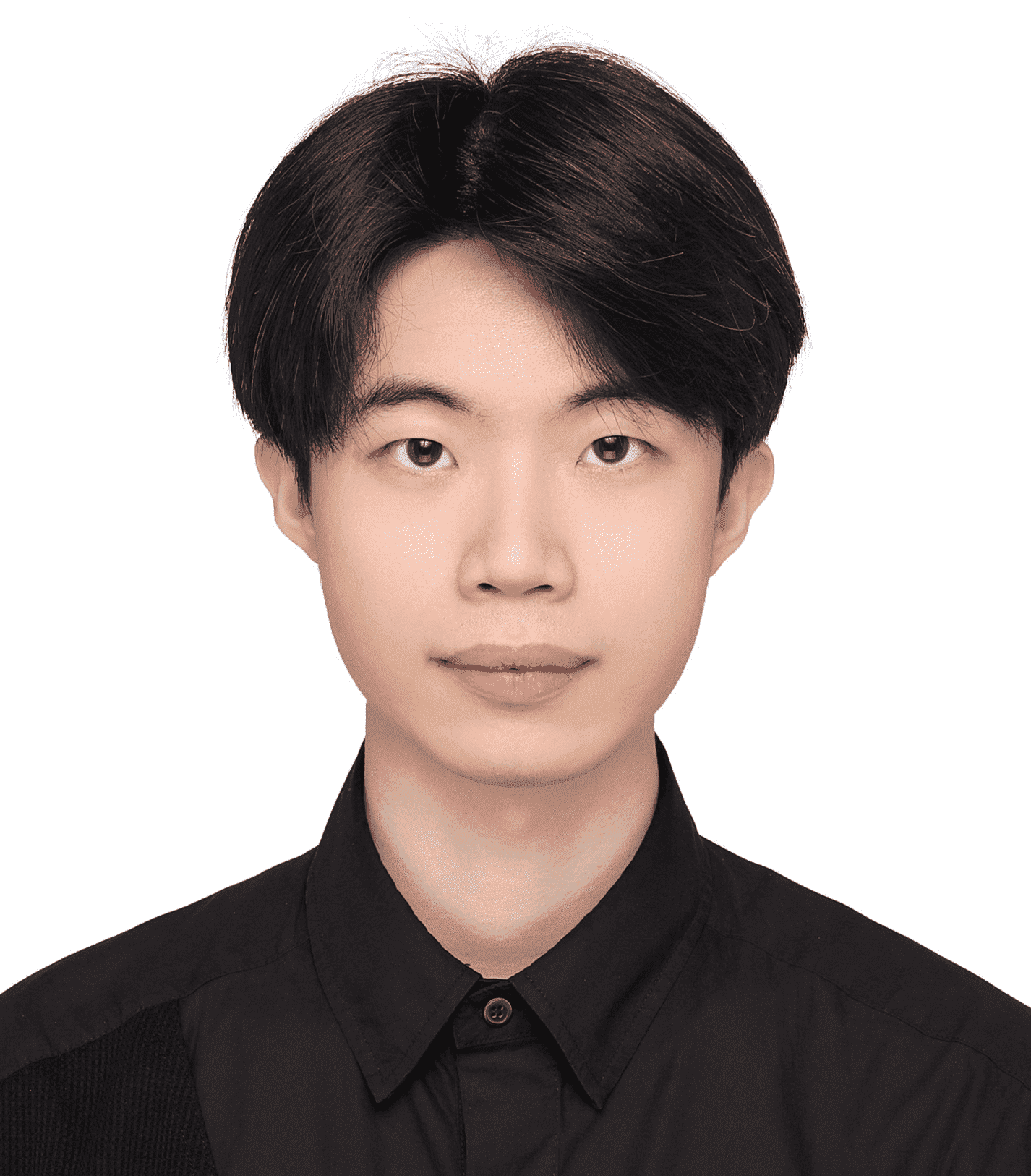}}]{Xiang Zhang}
received his B.E. degree in communication engineering from Wuhan University, Wuhan, China, in 2020. He is currently working toward an M.S. degree in information and communication engineering with the electronic information school, Wuhan University, Wuhan, China. His research interests include computer vision and neuromorphic computation.
\end{IEEEbiography}

\vskip -2.5\baselineskip plus -1fil

\begin{IEEEbiography}[{\includegraphics[width=1in,height=1.25in,clip,keepaspectratio, trim={0 40 0 5}]{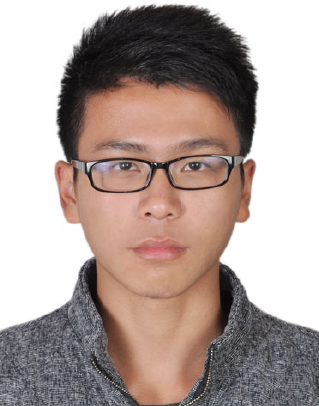}}]{Wei Liao}
received his B.E. degree in communication engineering from Wuhan University, Wuhan, China, in 2020. He is currently working toward an M.S. degree in information and communication engineering with the electronic information school, Wuhan University, Wuhan, China.  His research interests include image processing and signal processing.
\end{IEEEbiography}

\vskip -2.5\baselineskip plus -1fil

\begin{IEEEbiography}[{\includegraphics[width=1in,height=1.25in,clip,keepaspectratio, trim={0 25 0 10}]{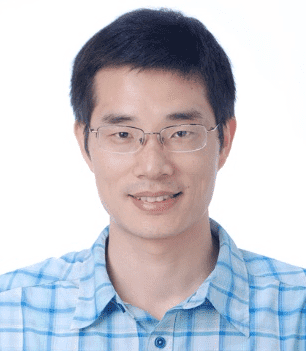}}]{Wen Yang} received the B.S. degree in electronic apparatus and surveying technology, and the M.S. degree in computer application technology and the Ph.D. degree in communication and information system from Wuhan University, Wuhan, China, in 1998, 2001, and 2004, respectively. From 2008 to 2009, he worked as a Visiting Scholar with the Apprentissage et Interfaces (AI) Team, Laboratoire Jean Kuntzmann, Grenoble, France. From 2010 to 2013, he worked as a Post-Doctoral Researcher with the State Key Laboratory of Information Engineering, Surveying, Mapping and Remote Sensing, Wuhan University. Since then, he has been a Full Professor with the School of Electronic Information, Wuhan University. He is also a guest professor of the Future Lab AI4EO in Technical University of Munich. His research interests include object detection and recognition, multisensor information fusion, and remote sensing image processing.
\end{IEEEbiography}

\vskip -2.5\baselineskip plus -1fil

\begin{IEEEbiography}[{\includegraphics[width=1in,height=1.25in,clip,keepaspectratio]{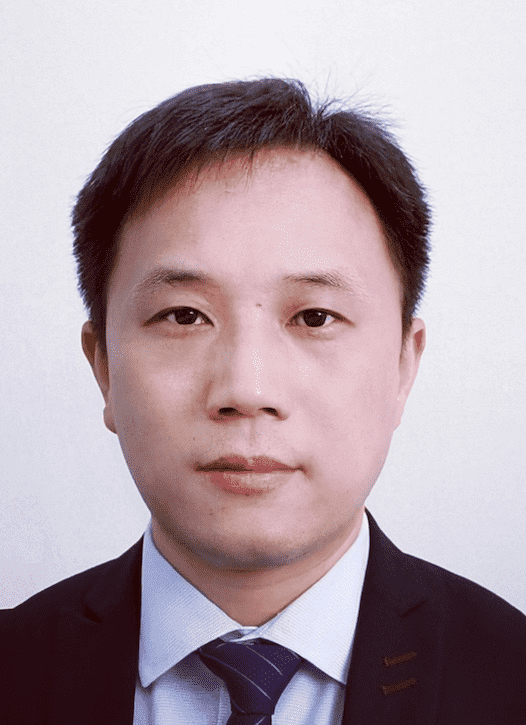}}]{Gui-Song Xia}
received his Ph.D. degree in image processing and computer vision from CNRS LTCI, T{\'e}l{\'e}com ParisTech, Paris, France, in 2011. From 2011 to 2012, he has been a Post-Doctoral Researcher with the Centre de Recherche en Math{\'e}matiques de la Decision, CNRS, Paris-Dauphine University, Paris, for one and a half years.
He is currently working as a full professor in computer vision and photogrammetry at Wuhan University. He has also been working as Visiting Scholar at DMA, {\'E}cole Normale Sup{\'e}rieure (ENS-Paris) for two months in 2018. He is also a guest professor of the Future Lab AI4EO in Technical University of Munich (TUM). His current research interests include mathematical modeling of images and videos, structure from motion, perceptual grouping, and remote sensing image understanding. He serves on the Editorial Boards of several journals, including {\em ISPRS Journal of Photogrammetry and Remote Sensing, Pattern Recognition, Signal Processing: Image Communications, EURASIP Journal on Image \& Video Processing, Journal of Remote Sensing, and Frontiers in Computer Science: Computer Vision}.
\end{IEEEbiography}







\end{document}